\documentclass{article} 
\usepackage{iclr2027_conference,times}

\usepackage{amsmath,amsfonts,bm}

\usepackage{svg}

\def\eqref#1{equation~\ref{#1}}

\def\1{\bm{1}}

\DeclareMathAlphabet{\mathsfit}{\encodingdefault}{\sfdefault}{m}{sl}
\SetMathAlphabet{\mathsfit}{bold}{\encodingdefault}{\sfdefault}{bx}{n}

\usepackage[figuresleft]{rotating}
\usepackage{graphicx}
\usepackage{subcaption}
\usepackage{titletoc}
\usepackage[table]{xcolor}
\definecolor{bestblue}{RGB}{220,235,247}

\usepackage{hyperref}
\usepackage{url}
\usepackage{booktabs}
\usepackage{multirow}
\usepackage{graphicx}
\usepackage[table]{xcolor}
\usepackage{tikz}
\usepackage{algorithm}
\usepackage{enumitem}
\usepackage{wrapfig}
\usepackage{booktabs}
\usepackage{multirow}
\usepackage{graphicx}
\newtheorem{definition}{Definition}
\usepackage{caption}
\usepackage{algpseudocode}
\newcommand{\blackcircle}[1]{%
    \tikz[baseline=(char.base)]{
        \node[shape=circle,fill=black,text=white,
              inner sep=1.2pt,font=\bfseries\small] (char) {#1};
    }%
}

\title{Unlocking the Specialization Capability of PPG Foundation Models for Wearable Signals through Structured Pruning and Adaptive Fine-Tuning}
\author{%
\makebox[\textwidth][l]{\hspace*{3pt}\textbf{Huaming Du\textsuperscript{1,2}\thanks{Equal contribution.} \ , Zhixian Xie\textsuperscript{3}\footnotemark[1] \ , Gaoyang Zhen\textsuperscript{3}\footnotemark[1] \ , Gang Kou\textsuperscript{1}, Guisong Liu\textsuperscript{1},}}\\
\makebox[\textwidth][l]{\hspace*{3pt}\textbf{Yanchao Tan\textsuperscript{3}, Wendan Tao\textsuperscript{4}\thanks{Corresponding authors.} \ , Yuhao Xu\textsuperscript{2}\footnotemark[2] \ , Carl Yang\textsuperscript{5}}}\\
\makebox[\textwidth][l]{\hspace*{3pt}\textsuperscript{1}Southwestern University of Finance and Economics\quad
\textsuperscript{2}Jizhi Technology Co., Ltd.}\\
\makebox[\textwidth][l]{\hspace*{3pt}\textsuperscript{3}Fuzhou University\quad
\textsuperscript{4}West China Hospital\quad
\textsuperscript{5}Emory University}\\
}

\iclrfinalcopy 

\begin{document}
\maketitle
\lhead{Preprint}

\begin{abstract}
Photoplethysmography (PPG) plays a central role in wearable health monitoring and clinical decision support. However, existing approaches to universal PPG representation learning are predominantly trained on high-quality large-scale datasets obtained from oximeters in hospital settings and often struggle to adapt to sparse and low-quality real-world datasets, thereby limiting their generalization to clinically relevant tasks on prevalent PPG data such as collected from wearable devices. Surprisingly, even after full fine-tuning, PPG foundation models cannot consistently outperform general-purpose time-series models. Therefore, a key question is how to effectively adapt PPG foundation models to target prediction tasks. Through empirical studies on various PPG foundation models, we find that pretrained models often exhibit inherent sparsity and computational redundancy, suggesting that PPG foundation models have learned to activate task-relevant network substructures for different prediction tasks. To preserve this valuable prior knowledge, we propose a structured pruning-and-fine-tuning framework, \textbf{SPAFT}, which regularizes the adaptive fine-tuning process by constraining it to a more relevant and compact parameter space based on parameter importance and structural redundancy. Extensive experiments across four PPG foundation models, five benchmark datasets, and ten downstream tasks demonstrate that SPAFT achieves win rates of 82.8\% and 87.5\% on regression and classification tasks, respectively. This “prune-then-finetune” paradigm often enables PPG foundation models to achieve state-of-the-art performance, laying a promising foundation for the real-world clinical deployment of PPG foundation models and the development of personalized PPG foundation models.

\end{abstract}
\section{Introduction}
Photoplethysmography (PPG) in wearable devices has become a widely used sensing modality due to its ability to assess physiological states non-invasively without requiring firm attachment to the skin \citep{charlton20232023}. By measuring blood volume changes through optical sensors, PPG enables the non-invasive assessment of various physiological indicators, including heart rate \citep{demirel2026continuous}, atrial fibrillation \citep{pereira2020photoplethysmography}, blood pressure \citep{wang2023pulsedb}, blood glucose \citep{saha2025pulse}, and respiratory rate \citep{he2026cap,kazemi2025respiration}.
In recent years, the growing availability of large-scale PPG data has driven the development of Physiological Foundation Models. Compared with training separate models from scratch for each clinical task, PPG foundation models (FMs) provide a promising solution for leveraging knowledge acquired through large-scale pretraining while reducing the dependence of downstream tasks on labeled data.

Despite the rapid development of PPG FMs, existing studies \citep{zhang2026ppgpt,chen2026bridging} have primarily focused on improving the representation capabilities of FMs through larger-scale pretraining data, more effective pretraining objectives, and cross-modal physiological information, while research on how to effectively adapt pre-trained models to different downstream datasets and tasks remains very limited.
Furthermore, learning high-quality general-purpose PPG representations through pretraining does not necessarily mean that these representations can directly meet the requirements of all downstream clinical tasks. PPG downstream applications exhibit substantial task heterogeneity, with different prediction targets relying on distinct physiological information embedded in the waveforms. For example, heart rate prediction primarily depends on the periodicity of pulse signals, whereas atrial fibrillation detection focuses more on rhythm irregularities. Moreover, clinically collected waveform data are often limited in scale, suboptimal in quality, and sparsely labeled \citep{xie2025state}. Therefore, \textit{how to effectively adapt pretrained PPG FMs to specific downstream tasks is a key challenge in transforming their general-purpose representations into task-specific representations.}

To address this challenge, we further investigate the internal representations of PPG FMs. Please note that we focus on convolutional neural network-based PPG FMs for two main reasons: 1) convolution-based PPG FMs achieve the best performance in most cases; and 2) most Transformer-based PPG FMs are not open-sourced. Our analysis reveals that pre-trained PPG FMs typically exhibit inherent sparsity, with a certain degree of redundancy in their parameters. For example, during downstream task execution, up to 50\% of filter outputs can be nearly negligible, while up to 80\% of intermediate filters remain inactive. These sparse patterns vary across datasets, suggesting that PPG FMs activate different subnetworks depending on the specific task. Although conventional fine-tuning preserves the original model architecture, indiscriminately updating all parameters may disrupt these task-relevant activation patterns. This not only increases the risk of overfitting but also fails to fully exploit the valuable task-specific substructures formed during pre-training.

In this paper, we propose \textbf{SPAFT}, a filter-based adaptive pruning-and-fine-tuning framework consisting of two stages: \textit{structural pruning} and \textit{adaptive fine-tuning}. During the structural compression stage, we first freeze the encoder and fit the task head to estimate the importance of individual units. We then construct an intra-layer convolutional parameter graph to quantify structural redundancy and accordingly allocate the pruning budget: the redundancy determines which layers should be pruned, while the importance determines which units should be removed within each layer. During the fine-tuning stage, we re-estimate the importance of individual units in the compressed model and adaptively allocate the trainable-unit budget for fine-tuning according to the relative concentration of highly scored units across different layers. Furthermore, to address individual heterogeneity, we explore \textit{personalized training} to further improve the predictive performance of PPG FMs. The key contributions of SPAFT are:
\begin{enumerate}[
    leftmargin=*,
    labelindent=0pt,
    labelsep=0.5em,
    itemsep=0.5em,
    topsep=0pt
]
    \item[\blackcircle{1}] \textbf{Identifying the specialization bottleneck of PPG FMs.} We conduct comprehensive and fair benchmark comparisons, showing that even after fine-tuning, PPG FMs still struggle to achieve strong generalization performance across different downstream tasks, highlighting PPG FM specialization as a critical unresolved challenge.
    \item[\blackcircle{2}] \textbf{Uncovering the inherent sparsity of PPG FMs.} Our empirical analysis reveals that pretrained PPG FMs activate only task-relevant subsets of parameters during downstream inference. We argue that these specialized model components represent valuable prior knowledge acquired during pretraining and should therefore be preserved when adapting PPG FMs to specific downstream PPG data.
    \item[\blackcircle{3}] \textbf{Pioneering fine-tuning with pruning for PPG FMs.} We propose a novel prune-then-finetune paradigm for post-training PPG FMs. To the best of our knowledge, this is the first work to improve predictive performance through pruning and personalized training of PPG FMs. Remarkably, we find that optimal performance can be achieved by fine‑tuning as little as approximately 1\% of the model parameters.
    \item[\blackcircle{4}] \textbf{Achieving simultaneous improvements in performance and efficiency.} We conduct extensive experiments across four PPG FMs, four public datasets, and a new dataset collected using smart rings in one of the largest hospitals in Asia. The results demonstrate that our method enables PPG FMs to outperform state-of-the-art baselines across diverse downstream task settings.
\end{enumerate}
\section{Related work}
\subsection{PPG Foundation Models}
In the field of PPG, existing foundation models primarily leverage the intrinsic characteristics of the signals, including inter-individual variability, waveform morphology, susceptibility to noise and artifacts, and temporal predictability \citep{ICLR2024_Large-scale,pillai2025papagei,ding2024siamquality}. PulsePPG \citep{saha2025pulse} further incorporates waveform morphology by designing contrastive learning objectives based on physiological indices or motif-level similarity, enabling the extraction of physiologically meaningful features. In contrast, GPT-PPG \citep{chen2025gpt} and PPGPT \citep{zhang2026ppgpt} explore generative pretraining approaches inspired by autoregressive language models to learn representations from PPG time series. Meanwhile, another line of research on foundation models for wearable sensing leverages multimodal data, including PPG or its derived signals \citep{2025-Beyond,narayanswamy2025scaling}, such as AnyPPG \citep{nie2026anyppg} and LeJEPA \citep{chen2026bridging}. However, these approaches are not specifically designed for PPG and therefore differ fundamentally in their modeling objectives from PPG-focused foundation models. A few studies \citep{xiang2026ultra,ni2025ppgdistill} have begun to explore the distillation and compression of PPG FMs, aiming to facilitate their deployment on edge devices. Despite the significant progress made by these studies, they have yet to explore the generalization capability of PPG FMs across downstream tasks, particularly on wearable data collected in real-world clinical settings, leaving a critical research gap that remains to be addressed.

\subsection{Model Pruning}
CNN filter pruning methods can be broadly divided into two categories: data-dependent filter pruning and data-independent filter pruning \citep{he2019filter,he2023structured}. Since the use of training data introduces additional computational overhead, data-independent methods are generally more efficient than data-dependent methods. \textbf{Data-Dependent Filter Pruning}. Some filter pruning methods \citep{zhuang2018discrimination,suau2018principal} require training data to determine which filters should be pruned.
\citep{he2020learning} designed a differentiable pruning criterion sampler, which is optimized using the validation loss of the pruned network obtained from the sampled criteria. \textbf{Data-Independent Filter Pruning}. Several data-independent filter pruning strategies \citep{Rethinking_2018,sui2021chip,park2024reprune} have also been explored. REPrune \citep{park2024reprune} first employs agglomerative clustering to identify similar kernels within each filter. It then selects filters that maximize the inclusion of representative kernels by optimizing the maximum cluster coverage problem. 
However, these techniques have not yet been applied to PPG FMs to investigate their generalizability.



\begin{figure*}[t]
\centering
\includegraphics[width=0.98\textwidth]{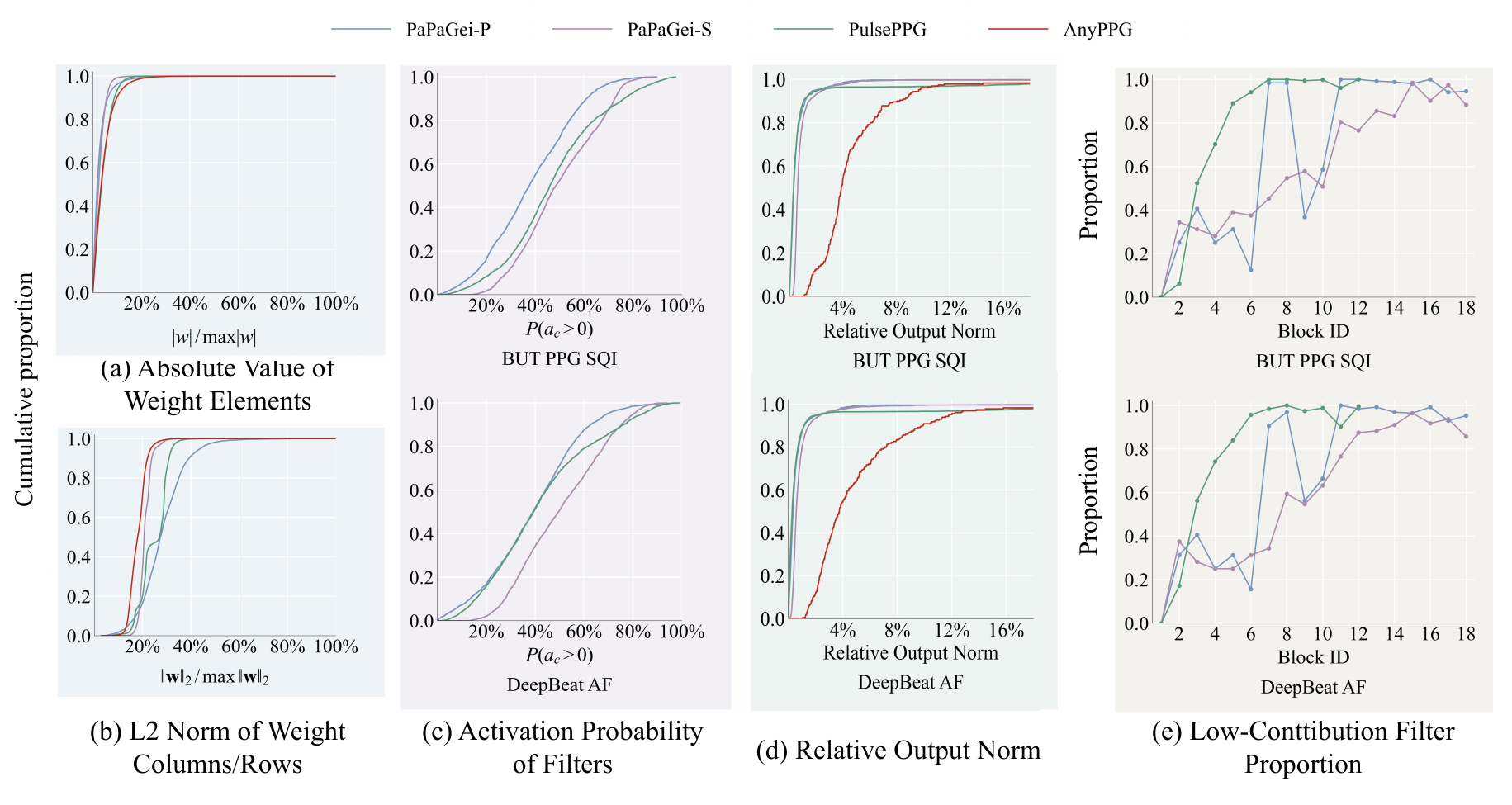}\par
\vspace{-3mm}
\caption{Empirical analysis. More results and analyses are provided in Appendix \ref{appendix:empirical}.}
\vspace{-4mm}
\label{fig:empirical}
\end{figure*}

\section{Preliminary}
\subsection{Empirical Analysis of PPG FMs}
To investigate the internal mechanisms of PPG FMs, we study them from the perspective of convolutional filters. We conduct experiments on the BUT PPG \citep{nemcova2021brno} and DeepBeat \citep{torres2020multi} datasets using four pretrained PPG FMs, including AnyPPG, PaPaGei-P~\citep{pillai2025papagei}, PaPaGei-S~\citep{pillai2025papagei}, and Pulse-PPG~\citep{saha2025pulse}.

\textbf{(1) Insignificant Weight Magnitudes and Outputs of filters.} We first examine at the parameter level whether PPG FMs contain a large proportion of low-magnitude parameters, providing initial evidence that the models may exhibit substantial redundancy. For each filter, we separately compute its parameter magnitude in the following two ways: (1) \textit{the normalized absolute values of the weight elements}; and (2) \textit{the normalized \(L_2\)-norm of the weight vector}.

As shown in Figure \ref{fig:empirical}(a) and (b), a substantial proportion of the parameter weights in the four PPG FMs have small magnitudes, exhibiting similarly skewed distributions.
Notably, parameter magnitude-based pruning is a task-agnostic approach and may lead to suboptimal performance on downstream tasks. Therefore, we use changes in the loss to compute importance scores.
We further measure the magnitude of each filter's output relative to the residual representation to quantify the extent to which that filter actually modifies the residual component. Specifically, we define it as follows.
\begin{definition}
Given the output of the \(i\)-th filter, \(o_i\in\mathbb{R}^d\), and the residual representation \(x\in\mathbb{R}^d\), the relative output norm of the \(i\)-th filter is defined over the downstream dataset as
$
\mathbb{E}_x\left(\|o_i\|_2/\|x\|_2\right).
$
\end{definition}
Figure \ref{fig:empirical}(d) illustrates the different distributions of relative output norms. Compared with AnyPPG, where all filters actively contribute at least 1.8\% relative norm, some filters in the other three PPG FMs produce outputs with nearly zero norms. Moreover, the degree of sparsity is affected by both downstream tasks and model architectures.

\textbf{(2) Sparse Activations in filters.} To further verify whether the low-contribution phenomenon arises from the model's selective use of filters, we measure and analyze the activation patterns of each filter.
\begin{definition} The activation probability of the \(i\)-th intermediate filter in PPG FMs is defined as the frequency with which its activation value $a_i$ is positive over the entire dataset, i.e., $P(a_i>0)$.
\end{definition}
Figure \ref{fig:empirical}(c) shows that PPG FMs exhibit a high degree of sparsity. For example, in the PaPaGei-P model, approximately 71\%–78\% of filters have an activation probability below 50\%.

\textbf{(3) Task-Specific Subnetwork.} We further investigate whether the structural sparsity of PPG FMs is a systematic characteristic that persists across different blocks\footnote{A block refers to a ResNet layer in the PPG FMs.}. We define blocks with an average relative output norm below \(1\%\) as unimportant blocks. Figure \ref{fig:empirical}(e) further demonstrates that significant sparsity patterns are widely present in PPG FMs. 

Based on the above analysis, most pre-trained PPG FMs exhibit inherent sparsity during inference, with a substantial proportion of parameters being either redundant or inactive. Moreover, they exhibit different sparsity patterns across different datasets. This motivates us to further explore more generalizable pruning-and-fine-tuning methods.

\begin{figure*}[t]
\centering
\includegraphics[width=0.88\textwidth]{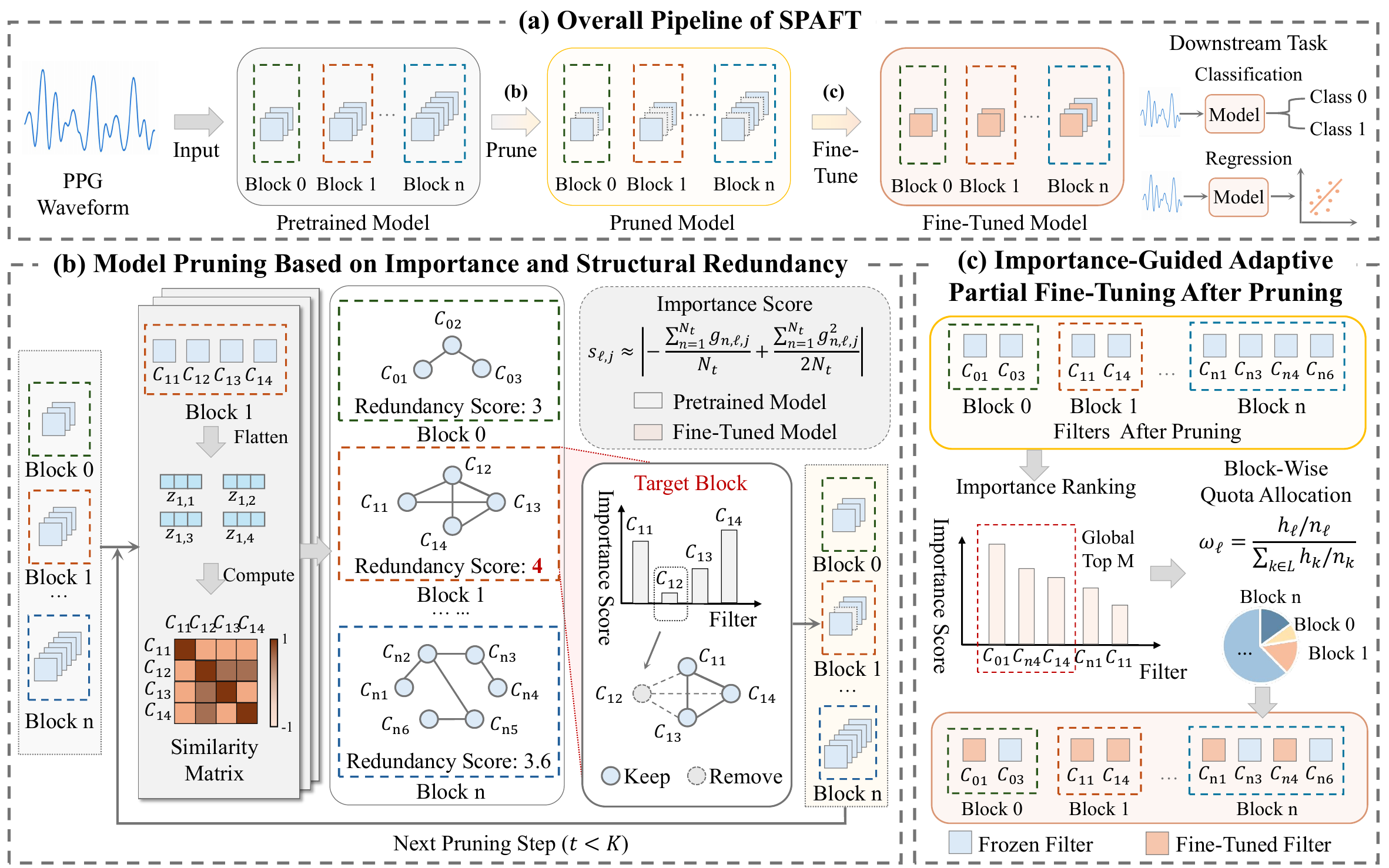}
\vspace{-6pt}
\caption{Architecture of Our Proposed SPAFT Framework.}
\label{fig:framework}
\vspace{-4mm}
\end{figure*}
\section{Methods}
Given a pretrained PPG encoder $E_{\boldsymbol\theta}$, a task head $h_{\boldsymbol\phi}$ and a downstream dataset $\mathcal D=\{(\mathbf x_n,y_n)\}_{n=1}^{N}$, where $y_n$ denotes either a continuous physiological measurement or a binary classification label, the complete prediction model is defined as
$
F_{\boldsymbol\theta,\boldsymbol\phi}(\mathbf x)
=h_{\boldsymbol\phi}\!\left(E_{\boldsymbol\theta}(\mathbf x)\right).
$
Given a global pruning ratio $\rho$, our goal is to remove computational units from the encoder that contribute less to the current task and exhibit strong structural redundancy, thereby obtaining a compact encoder $E_{\boldsymbol\theta^\star}$ and its task head $h_{\boldsymbol\phi^\star}$, which are then fine-tuned to improve downstream task performance.

The overall pipeline is illustrated in Figure \ref{fig:framework}. First, we perform a single forward pass over the unpruned model to estimate the task importance of all nodes\footnote{The nodes refer to filters in PPG FMs.}. We then construct an intra-layer structural graph based on the cosine relationships among convolutional kernel parameters and allocate pruning opportunities across layers according to their structural redundancy. Based on the computed structural redundancy and task importance, we perform node pruning planning and subsequently conduct adaptive fine-tuning on the compact model.
\subsection{Model Pruning Based on Importance and Structural Redundancy
}\label{sec:model_pruning}
\textbf{Task Importance Estimation Before Pruning.} We first freeze the encoder and then fit a linear regression head or a logistic classification head according to the task requirements. Next, we perform a one-time importance estimation on the unpruned model using the training set $\mathcal D_t$. The loss change caused by removing a specific node is
\begin{equation}
\Delta_{\ell,j} = \left| \mathcal L(\mathbf{m}-\mathbf e_{\ell,j}) - \mathcal L(\mathbf{m}) \right|, \quad
\mathcal L(\mathbf m)
=\mathcal L\!\left(F_{\boldsymbol\theta,\boldsymbol\phi}^{\mathbf m}(\mathbf x),y\right).
\end{equation}
where $\mathbf{m}$ denotes the gate vector for all nodes, and $\mathbf{e}_{\ell,j}$ denotes an indicator vector that takes the value 1 only at node $\left(\ell,j\right)$, corresponding to the $j$-th filter in the $\ell$-th block.
A smaller $\Delta_{\ell,j}$ indicates that the corresponding node has a weaker influence on the current task. However, explicitly computing this loss difference for each node requires a large number of repeated inference passes. To address this issue, we approximate the loss change caused by node removal at $\mathbf{m}_{\ell,j}=1$. Let the gradient of the $n$-th sample with respect to the node gate be
\begin{equation}
g_{n,\ell,j} = \left (\partial\mathcal L_n(\mathbf {m})/ {\partial \mathbf{m}_{\ell,j}} \right )|_{\mathbf{m}_{\ell,j}=1}.
\end{equation}
By applying a second-order Taylor expansion to \(\Delta_{\ell,j}\) and approximating the diagonal entries of the Hessian matrix using the squared per-sample first-order gradients, the node importance is given by
\begin{equation}\label{equation:s}
    s_{\ell,j}\approx
\left|
-\mathbf{e}_{\ell,j}^{\top}\nabla_{\mathbf{m}}\mathcal{L}
+\frac{1}{2}
\mathbf{e}_{\ell,j}^{\top}
\left(\nabla_{\mathbf{m}}^2\mathcal{L}\right)
\mathbf{e}_{\ell,j}
\right|\approx \left| -\frac{\sum_{n=1}^{N_t}g_{n,\ell,j}} {N_t} + \frac{\sum_{n=1}^{N_t}g_{n,\ell,j}^{2}}{2N_t}  \right|\ ,
\end{equation}

where the first term describes the first-order loss change induced by removing the gate, while the second term provides a second-order correction based on an empirical Fisher information approximation. $N_t$ denotes the number of samples in the training set. Note that, for efficiency, $s$ remains fixed throughout the subsequent pruning steps and is not recomputed.

\textbf{Structural Redundancy on the Cosine Graph.} Task importance characterizes the relationship between individual nodes and the downstream target, but it alone cannot determine how many nodes can be removed from different layers \citep{Convolutional2021}. To address this issue, we represent each block as an undirected graph and use graph structural redundancy to characterize the substitutability of nodes within the block.

Specifically, the structural feature $\mathbf z_{\ell,j}$ of a node is the flattened parameter vector of the corresponding output convolution kernel in the first convolutional layer. 
We normalize the node features and construct the current adjacency relations according to the cosine threshold $\tau$:
\begin{equation}
\overline{\mathbf z}_{\ell,j}
=\mathbf z_{\ell,j}/\|\mathbf z_{\ell,j}\|_2,\quad
A_{\ell,ij}^{(t)}
=\mathbb I\!\left[
\overline{\mathbf z}_{\ell,i}^{\mathsf T}\overline{\mathbf z}_{\ell,j}\geq\tau
\right],
\quad i,j\in V_\ell^{(t)},\ i\neq j\ ,
\end{equation}
where $A_{\ell,ii}^{(t)}=0$, $V_\ell^{(t)}$ denote the set of active nodes in block $\ell$ after $t$ global pruning steps. The cosine metric removes the influence of the overall magnitude of the parameter vectors, while the threshold $\tau$ controls edge formation. 

Based on the constructed cosine graph, we combine connected components and local coverage \citep{kolmogorov1959varepsilon} to measure structural redundancy. Let $k_\ell^{(t)}$ denote the number of connected components, and define a closed ball of radius $r$ using the graph distance $d_{G_\ell^{(t)}}$: $
\mathcal B_{\ell,r}^{(t)}(v)
=\left\{u\in V_\ell^{(t)}:
d_{G_\ell^{(t)}}(u,v)\leq r\right\}, r\in\{1,2\}.
$ For each radius, among the uncovered nodes, we select the node with the largest degree in the current induced graph and mark its closed ball as covered, repeating this procedure until the entire graph is covered. 
Let the resulting greedy cover number be $c_{\ell,t}^{(r)}$, and define $\widehat N_{\ell,1c}^{(t)}=(c_{\ell,t}^{(1)}+c_{\ell,t}^{(2)})/2$. The structural redundancy is defined as:
\begin{equation}\label{equation:r}
    R_\ell^{(t)}
=\frac{N_\ell^{(t)}}
{w_1k_\ell^{(t)}+w_2\widehat N_{\ell,1c}^{(t)}},
\qquad
N_\ell^{(t)}=|V_\ell^{(t)}|\ ,
\end{equation}
where ${w_1,w_2}$ are the probability weights.
 Redundancy is higher when a larger number of nodes can be represented by fewer structural families and local centers.


\textbf{Task-Structure Collaborative Node Pruning.} We use structural redundancy to allocate pruning opportunities across layers and frozen importance scores to select nodes within each layer. This hierarchical decision mechanism avoids directly comparing importance scores across layers and does not require predefined layer-wise pruning ratios.

To prevent any residual branch from degenerating into an empty branch, only blocks that retain at least two nodes are considered eligible for pruning. At step $t$, we first select the block $\ell_t$ with the highest current structural redundancy. Subsequently, among the currently active nodes in the selected block, we remove the actual node $j_t$ with the lowest task importance, thereby updating the active node sets. Then, only the induced graph of block \(\ell_t\) over \(V_{\ell_t}^{(t)}\) is updated, and the number of connected components, coverage estimate, and structural redundancy of that block are recomputed, while the graph states of all other blocks remain unchanged. Repeating the above steps for $K$ iterations yields the final node set for each block: $V_\ell^\star = V_\ell^{(K)}$.

\subsection{Importance-Guided Adaptive Fine-Tuning After Pruning}
Node removal changes the predictive function, and the importance of the remaining nodes may change accordingly. To allocate the limited update budget to paths \footnote{The path refers to the connection between two blocks in PPG FMs, including BatchNorm (BN) layers.} that remain highly task-sensitive after pruning, we re-estimate importance on the compact model and allocate fine-tuning quotas according to the concentration of high-scoring nodes within each block.

\textbf{Post-Pruning Importance Re-estimation.} Generate gating masks based on the final active set \(V_\ell^\star\), and slice the convolutional and normalization tensors to obtain the compact encoder parameters \(\theta_c\).
We reintroduce gates $\widetilde m_{\ell,j}$ initialized to $1$ on the remaining nodes, freeze the compact encoder, the original task head, and the BN states, and reuse the losses from Section \ref{sec:model_pruning}. Furthermore, we can compute
\begin{equation}\label{eq:restimation}
\widetilde g_{n,\ell,j}
=\left.
\frac{\partial\ell\!\left(
F_{\boldsymbol\theta_{\mathrm c},\quad \boldsymbol\phi}^{\widetilde{\mathbf m}}(\mathbf x_n),y_n
\right)}
{\partial\widetilde m_{\ell,j}}
\right|_{\widetilde{\mathbf m}=\mathbf1},\quad
\widetilde s_{\ell,j}
=\left|-\frac{\sum_{n=1}^{N_f}\widetilde{g}_{n,\ell,j}} {N_f} + \frac{\sum_{n=1}^{N_f}\widetilde{g}_{n,\ell,j}^{2}}{2N_f}  \right|\ ,
\quad j\in V_\ell^\star.
\end{equation}
The pre-pruning scores $s_{\ell,j}$ determine which nodes are removed, whereas the post-pruning scores $\widetilde s_{\ell,j}$ determine which nodes are selected for fine-tuning; they respectively characterize task sensitivity in the original and compact networks. 

\textbf{Layer-Width-Normalized Budget Allocation.} Since different blocks may have different widths after pruning, directly allocating the budget based on the number of high-scoring nodes would be influenced by block size. Therefore, we measure the contribution of each layer using the proportion of high-scoring nodes within that block.
The specific formulation is given as follows:
\begin{equation}
h_\ell=\left|{j:(\ell,j)\in\mathcal I_\mathrm{M}}\right|,
\qquad
d_\ell=\frac{h_\ell}{n_\ell},
\qquad
\omega_\ell=\frac{d_\ell}{\sum_{k\in L}d_k}\ ,
\end{equation}
where the set $\mathcal I_\mathrm{M}$ contains the $\mathrm{M}=\lceil\gamma N_{\mathrm c}\rceil$ highest-scoring remaining nodes, and $\gamma$ is the filter ratio of the PPG FMs to be fine-tuned, $N_{\mathrm c}$ is the total number of remaining nodes,  $\left \lceil \cdot \right \rceil$ denotes the ceiling operation, $L$ is the number of blocks in the PPG FMs. To preserve an adaptation path for every block, we first reserve one node for each block and then allocate the remaining $\mathrm{M}-L$ quotas according to $\omega_\ell$. The final quota satisfies
\begin{equation}\label{eq:quota}
   q_\ell=1+\mathrm{M}\omega_\ell,
\qquad
0\leq \mathrm{M}\omega_\ell\leq n_\ell-1,
\qquad
\sum_{\ell\in L}q_\ell=\mathrm{M}\ . 
\end{equation}


\textbf{Constrained Optimization Along Filter Paths.} For each selected node, the convolutional parameters along its dependency path are made trainable.
For classification tasks, the fine-tuning objective is formulated as
\begin{equation}\label{eq:fine-object}
\min_{\boldsymbol\theta,\boldsymbol\phi'}\quad
\frac1{N_f}\sum_{(\mathbf x_n,y_n)\in\mathcal D_f}
\ell_{\mathrm{fin}}\left(
F_{\boldsymbol\theta,\boldsymbol\phi'}(\mathbf x_n),y_n\right)\ .
\end{equation}

For regression tasks, the prediction residual is normalized by the target standard deviation \(\sigma_f\) of the fitting set, and the Smooth L1 loss \citep{girshick2015fast} is used:
\begin{equation}\label{eq:l1}
    e_n=\frac{\hat y_n-y_n}{\sigma_f},\quad
\ell_{\mathrm{rec}}(\hat y_n,y_n)
=\begin{cases}
e_n^2/2,&|e_n|<1,\\
|e_n|-1/2,&|e_n|\geq1.
\end{cases}
\end{equation}

\section{Experiments}
In this section, we conduct extensive experiments under both data-wise and subject-wise splitting settings to evaluate the effectiveness of our method for adapting PPG FMs to downstream tasks, as well as its effectiveness for personalized training. Additional experimental results are provided in Appendix \ref{appendix:add_experimental}, including subnetwork transfer analysis (Appendix \ref{appendix:transfer}), inference efficiency analysis (Appendix \ref{appendix:infer}), adaptivity analysis (Appendix \ref{appendix:adaptive}), and parameter analysis (Appendix \ref{appendix:parameter}).
\subsection{Experimental Settings}
\textbf{Datasets.} 
To evaluate the effectiveness of SPAFT, we conduct extensive experiments across multiple datasets and tasks. Our evaluation covers both clinical settings and portable or wearable device scenarios. For regression tasks, we use UCI-BP~\citep{kachuee2015cuff}, PPG-DaLiA~\citep{reiss2019ppg}, BUT PPG, and our newly collected Ring-PPG Dataset; for classification tasks, we use BUT PPG and DeepBeat. Further details are provided in Appendix \ref{appendix:data}.

\textbf{Baselines.} To comprehensively evaluate SPAFT, we compare it with fourteen representative baseline methods from five categories: (1) \textbf{conventional adaptation methods}: Linear Probe, Full-FT, LoRA, and QLoRA; (2) \textbf{pruning-based adaptation methods}: PFEC~\citep{li2017pruning}, Taylor-FO~\citep{molchanov2019importance}, FPGM~\citep{he2019filter}, SRR-GR~\citep{Convolutional2021}, and Less-is-More~\citep{zhao2026less}; (3) \textbf{Distillation-based Methods}:  PPG-DISTIL \citep{ni2025ppgdistill}; (4) \textbf{Time-Series Foundation Models}: Chronos \citep{ansari2024chronos} and MOMENT \citep{goswami2024moment}; and (5)\textbf{Transformer-based PPG FMs}: GPT-PPG \citep{chen2025gpt} and SIGMA-PPG \citep{guo2026sigmappg}. The details of the baselines are provided in Appendix~\ref{app:baseline_details}.

\textbf{Implementation Details.} We evaluate SPAFT using pretrained AnyPPG, PaPaGei-P, PaPaGei-S, and Pulse-PPG. The adaptive fine-tuning stage uses AdamW with an effective batch size of 1,024 for at most 10 epochs and early-stopping patience of three. Task-specific search spaces cover the cosine threshold $\tau$, global pruning ratio $\rho$, fine-tuning ratio $\gamma$, and learning rate.
Specific details are provided in Appendix~\ref{appendix:exper_detail}.
\subsection{Overall Performance}
Tables~\ref{tab:overall_performance_1} and~\ref{tab:overall_performance_2} report the regression and classification results, respectively, across diverse downstream PPG tasks. We have the following observations: \textbf{(1) Overall, SPAFT achieves the strongest aggregate performance across both regression and classification tasks.}
SPAFT achieves win rates of 82.8\% and 87.5\% on regression and classification tasks, respectively. Its superior performance demonstrates that SPAFT can effectively specialize general-purpose PPG representations to meet the requirements of diverse downstream tasks. \textbf{(2) Compared with conventional adaptation methods.}
SPAFT achieves better results in 60 out of 64 regression comparisons and 15 out of 16 classification comparisons. In particular, Full-FT also underperforms SPAFT in most settings, indicating that updating a larger parameter space does not necessarily lead to better task specialization. This observation is consistent with our motivation that indiscriminate model adaptation may disturb task-relevant substructures acquired during pretraining. In contrast, SPAFT provides a more targeted adaptation process for downstream task specialization.
 \textbf{(3) Compared with pruning-based adaptation methods.}
SPAFT outperforms the best-performing pruning baseline in 54 out of 64 regression comparisons and 13 out of 16 classification comparisons, indicating that its performance gains do not simply stem from introducing pruning before fine-tuning. Although Less-is-More also adopts a task-aware prune-then-finetune strategy, it directly fine-tunes the retained compact backbone after pruning. In contrast, SPAFT adapts only the selectively identified parameter paths after compression, thereby further constraining downstream optimization to task-relevant substructures.
\begin{table*}[t]
\centering
\caption{Model performance on regression tasks.
We mark the winner in \textbf{bold}, and the runner-up is indicated by an \underline{underline}. The complete results are provided in Table \ref{tab:Complete_overall_performance_1} of Appendix \ref{appendix:overall}.}

\label{tab:overall_performance_1}

\resizebox{\textwidth}{!}{
\begin{tabular}{
c|c|c|
*{9}{cc|}
cc
}
\toprule
\multirow{2}{*}{\raisebox{-0.5ex}{\textbf{Data}}}
& \multirow{2}{*}{\raisebox{-0.5ex}{\textbf{Tasks}}}
& \multirow{2}{*}{\raisebox{-0.5ex}{\textbf{FM}}}
& \multicolumn{2}{c|}{\textbf{Linear Probe}}
& \multicolumn{2}{c|}{\textbf{Full-FT}}
& \multicolumn{2}{c|}{\textbf{LoRA}}
& \multicolumn{2}{c|}{\textbf{QLoRA}}
& \multicolumn{2}{c|}{\textbf{PFEC}}
& \multicolumn{2}{c|}{\textbf{Taylor-FO}}
& \multicolumn{2}{c|}{\textbf{FPGM}}
& \multicolumn{2}{c|}{\textbf{SRR-GR}}
& \multicolumn{2}{c|}{\textbf{Less-is-More}}
& \multicolumn{2}{c}{\textbf{Ours}}
\\

& & &
MSE $\downarrow$ & MAE $\downarrow$
& MSE $\downarrow$ & MAE $\downarrow$
& MSE $\downarrow$ & MAE $\downarrow$
& MSE $\downarrow$ & MAE $\downarrow$
& MSE $\downarrow$ & MAE $\downarrow$
& MSE $\downarrow$ & MAE $\downarrow$
& MSE $\downarrow$ & MAE $\downarrow$
& MSE $\downarrow$ & MAE $\downarrow$
& MSE $\downarrow$ & MAE $\downarrow$
& MSE $\downarrow$ & MAE $\downarrow$
\\
\midrule

\multirow{8}{*}{UCI-BP}&\multirow{4}{*}{SBP}
& AnyPPG
& 379.531 & 15.388 & 221.626 & 10.804
& 327.334 & 13.870 & 392.230 & 15.488
& 460.250 & 16.821 & \underline{191.174} & \underline{9.871}
& 223.532 & 10.885 & 338.435 & 14.133
& 379.268 & 15.086 & \cellcolor{bestblue}\textbf{143.083} & \cellcolor{bestblue}\textbf{8.345}
\\

&& PaPaGei-P
& 430.231 & 16.542 & 371.494 & 14.974
& 379.325 & 15.183 & 471.292 & 17.481
& 482.716 & 17.735 & 371.448 & 14.924
& \underline{365.637} & \underline{14.757} & 482.458 & 17.775
& 434.871 & 16.591 & \cellcolor{bestblue}\textbf{245.788} & \cellcolor{bestblue}\textbf{11.438}
\\

&&PaPaGei-S
& 461.856 & 17.295 & 383.028 & 15.299
& 422.433 & 16.301 & 438.100 & 16.600
& 492.882 & 17.987 & 393.232 & 15.628
& \underline{380.783} & \underline{15.240} & 445.977 & 16.893
& 387.265 & 15.478 & \cellcolor{bestblue}\textbf{259.401} & \cellcolor{bestblue}\textbf{11.858}
\\

&&Pulse-PPG
& 417.647 & 16.263 & 256.021 & 11.825
& 356.951 & 14.652 & 354.624 & 14.660
& 335.480 & 14.175 & 257.260 & 11.845
& \underline{254.844} & \underline{11.788} & 312.325 & 13.553
& 277.007 & 12.581 & \cellcolor{bestblue}\textbf{183.581} & \cellcolor{bestblue}\textbf{9.661}
\\

\cmidrule{2-23}
&\multirow{4}{*}{DBP}
& AnyPPG
& 87.673 & 6.881 & 59.591 & 5.156
& 78.304 & 6.224 & 92.546 & 7.243
& 82.621 & 6.564 & \underline{53.857} & \underline{4.816}
& 78.587 & 6.081 & 76.814 & 6.256
& 80.273 & 6.475 & \cellcolor{bestblue}\textbf{41.276} & \cellcolor{bestblue}\textbf{4.111}
\\

&& PaPaGei-P
& 98.548 & 7.436 & 87.802 & 6.735
& 88.378 & 6.752 & 92.338 & 7.192
& 88.084 & 6.814 & 90.350 & 6.846
& \underline{87.061} & \underline{6.659} & 99.238 & 7.458
& 89.353 & 6.992 & \cellcolor{bestblue}\textbf{62.908} & \cellcolor{bestblue}\textbf{5.317}
\\

&&PaPaGei-S
& 104.376 & 7.697 & \underline{82.521} & \underline{6.446}
& 95.727 & 7.113 & 99.174 & 7.196
& 88.765 & 6.918 & 85.060 & 6.608
& 84.573 & 6.579 & 95.835 & 7.156
& 90.817 & 6.973 & \cellcolor{bestblue}\textbf{64.756} & \cellcolor{bestblue}\textbf{5.420}
\\

&&Pulse-PPG
& 95.357 & 7.283 & 66.772 & 5.575
& 82.859 & 6.460 & 82.715 & 6.559
& 96.364 & 7.074 & \underline{66.458} & \underline{5.550}
& 67.699 & 5.617 & 78.050 & 6.202
& 91.655 & 6.907 & \cellcolor{bestblue}\textbf{49.865} & \cellcolor{bestblue}\textbf{4.609}
\\

\midrule
\multirow{4}{*}{PPG-DaLiA}&\multirow{4}{*}{HR}
& AnyPPG
& 92.093 & 6.505 & 82.302 & 5.597
& 87.603 & 6.304 & 242.786 & 11.836
& 151.705 & 7.873 & \underline{75.095} & \underline{5.385}
& 164.685 & 7.387 & 142.293 & 7.181
& 145.387 & 7.923 & \cellcolor{bestblue}\textbf{71.870} & \cellcolor{bestblue}\textbf{5.066}
\\

&& PaPaGei-P
& 196.916 & 10.282 & 222.989 & 11.029
& 190.257 & 10.043 & 263.867 & 12.824
& 212.701 & 11.105 & 212.615 & 11.002
& 259.243 & 12.630 & 190.598 & 10.054
& \underline{174.190} & \underline{9.386} & \cellcolor{bestblue}\textbf{108.284} & \cellcolor{bestblue}\textbf{6.818}
\\

&&PaPaGei-S
& 228.658 & 11.563 & 258.853 & 12.017
& 217.661 & 11.195 & 277.062 & 13.108
& 229.755 & 11.384 & \underline{174.363} & \underline{9.736}
& 311.547 & 13.493 & 215.235 & 11.246
& 285.047 & 13.502 & \cellcolor{bestblue}\textbf{130.235} & \cellcolor{bestblue}\textbf{7.977}
\\

&& Pulse-PPG
& 149.143 & 8.996 & 144.507 & 8.707
& 145.238 & 8.869 & 199.532 & 10.704
& 221.702 & 11.086 & 130.593 & 8.291
& \underline{121.318} & \underline{7.922} & 242.504 & 11.332
& 253.971 & 12.197 & \cellcolor{bestblue}\textbf{106.274} & \cellcolor{bestblue}\textbf{7.191}
\\

\midrule
\multirow{16}{*}{Ring-PPG}&\multirow{4}{*}{HR} 
& AnyPPG
& 74.642 & \underline{6.167} & 80.774 & 7.376
& 77.026 & 6.446 & 62.443 & 6.479
& 75.636 & 7.004 & 76.961 & 6.427
& 92.972 & 7.778 & 64.864 & 6.469
& \underline{59.285} & 6.371 & \cellcolor{bestblue}\textbf{58.447} & \cellcolor{bestblue}\textbf{5.717}
\\

&& PaPaGei-P
& 78.552 & 6.645 & 91.423 & 7.289
& 80.245 & 6.703 & 74.180 & 7.021
& \cellcolor{bestblue}\textbf{58.648} & \cellcolor{bestblue}\textbf{6.147} & 76.612 & 6.611
& 154.472 & 10.378 & 60.984 & 6.424
& \underline{59.671} & \underline{6.283} & 70.638 & 6.296
\\

&& PaPaGei-S
& \underline{92.354} & 7.999 & 129.305 & 9.622
& 153.571 & 10.458 & 94.347 & 8.019
& 99.860 & 8.328 & 209.814 & 12.192
& 109.040 & 8.859 & 102.507 & 8.464
& 94.254 & \underline{7.893} & \cellcolor{bestblue}\textbf{77.574} & \cellcolor{bestblue}\textbf{7.278}
\\
&& Pulse-PPG
& 76.184 & 6.715 & \cellcolor{bestblue}\textbf{69.563} & \cellcolor{bestblue}\textbf{6.531}
& 81.410 & 7.009 & 80.308 & 7.259
& 93.583 & 7.684 & 77.627 & \underline{6.544}
& 164.395 & 10.448 & 97.916 & 7.799
& 86.114 & 7.228 & \underline{75.062} & 6.697
\\

\cmidrule{2-23}
&\multirow{4}{*}{SBP}
& AnyPPG
& 335.459 & 14.787 & 555.882 & 19.105
& 323.997 & 14.672 & 214.162 & 12.225
& 234.743 & 12.668 & 321.300 & 13.957
& 333.044 & 14.664 & 224.976 & 12.301
& \underline{199.295} & \underline{11.839} & \cellcolor{bestblue}\textbf{155.748} & \cellcolor{bestblue}\textbf{10.751}
\\

&& PaPaGei-P
& 389.351 & 16.079 & 391.395 & 16.431
& 358.480 & 15.566 & 212.676 & 12.178
& 219.713 & 12.468 & 388.049 & 16.048
& 239.318 & 13.021 & 223.084 & 12.487
& \underline{188.714} & \underline{11.984} & \cellcolor{bestblue}\textbf{162.895} & \cellcolor{bestblue}\textbf{10.843}
\\

&& PaPaGei-S
& 391.653 & 15.372 & 427.919 & 16.231
& 339.616 & 14.218 & \underline{189.743} & 11.814
& 204.748 & 12.191 & 580.246 & 19.623
& 230.095 & 12.836 & 197.943 & 11.873
& 191.154 & \underline{11.752} & \cellcolor{bestblue}\textbf{152.555} & \cellcolor{bestblue}\textbf{10.454}
\\

&& Pulse-PPG
& 321.497 & 14.497 & 382.435 & 16.023
& 291.316 & 13.543 & 237.512 & \underline{12.598}
& 271.118 & 13.579 & 418.904 & 16.017
& 300.412 & 14.384 & 309.956 & 14.512
& \underline{194.090} & 13.322 & \cellcolor{bestblue}\textbf{136.191} & \cellcolor{bestblue}\textbf{9.417}
\\

\cmidrule{2-23}
&\multirow{4}{*}{DBP}
& AnyPPG
& 245.139 & 11.525 & 285.319 & 12.774
& 248.110 & 11.537 & 220.622 & 12.007
& 329.623 & 14.543 & \underline{199.901} & \underline{11.371}
& 213.504 & 11.639 & 363.325 & 15.538
& 309.329 & 14.167 & \cellcolor{bestblue}\textbf{153.459} & \cellcolor{bestblue}\textbf{10.195}
\\

&& PaPaGei-P
& 336.107 & 15.038 & 356.623 & 15.129
& 334.480 & 14.987 & 216.306 & 12.184
& 211.046 & 12.168 & 311.798 & 14.176
& 228.868 & 12.446 & \underline{200.726} & \underline{12.153}
& 218.867 & 12.320 & \cellcolor{bestblue}\textbf{186.804} & \cellcolor{bestblue}\textbf{11.337}
\\

&& PaPaGei-S
& 287.337 & 13.617 & 412.088 & 15.609
& 288.868 & 13.626 & \underline{207.111} & \cellcolor{bestblue}\textbf{11.881}
& 213.396 & 12.157 & 551.298 & 18.605
& 362.254 & 14.935 & 260.305 & 13.131
& 299.418 & 13.876 & \cellcolor{bestblue}\textbf{193.974} & \underline{11.898}
\\

&& Pulse-PPG
& \underline{239.984} & \underline{12.160} & 277.450 & 13.754
& 240.546 & 12.233 & 608.723 & 18.827
& 547.154 & 16.620 & 244.191 & 12.772
& 285.420 & 13.961 & 353.672 & 15.587
& 752.999 & 22.112 & \cellcolor{bestblue}\textbf{176.193} & \cellcolor{bestblue}\textbf{11.405}
\\

\cmidrule{2-23}
& \multirow{4}{*}{SpO$_2$}
& AnyPPG
& 1.787 & 1.061 & 2.231 & 1.200
& 2.080 & 1.151 & 1.638 & 1.047
& 1.795 & 1.098 & 1.827 & 1.107
& 1.488 & 0.985 & 1.789 & 1.073
& \underline{1.288} & \underline{0.937} & \cellcolor{bestblue}\textbf{0.898} & \cellcolor{bestblue}\textbf{0.763}
\\

&& PaPaGei-P
& 2.093 & 1.182 & 2.998 & 1.453
& 2.205 & 1.208 & \underline{1.626} & 1.089
& 1.719 & 1.137 & 1.927 & 1.133
& 1.774 & 1.121 & 1.710 & 1.124
& 1.665 & \underline{1.079} & \cellcolor{bestblue}\textbf{0.793} & \cellcolor{bestblue}\textbf{0.749}
\\

&& PaPaGei-S
& 2.015 & 1.134 & 2.417 & 1.315
& 2.728 & 1.299 & 1.918 & 1.190
& 1.869 & 1.160 & 7.452 & 2.301
& 2.506 & 1.366 & 1.857 & 1.163
& \underline{1.665} & \underline{1.094} & \cellcolor{bestblue}\textbf{0.935} & \cellcolor{bestblue}\textbf{0.797}
\\

&& Pulse-PPG
& 2.211 & 1.226 & 2.443 & 1.296
& 2.239 & 1.245 & \underline{1.622} & \underline{1.076}
& 2.498 & 1.281 & 1.976 & 1.176
& 2.480 & 1.376 & 2.267 & 1.278
& 1.693 & 1.094 & \cellcolor{bestblue}\textbf{0.872} & \cellcolor{bestblue}\textbf{0.758}
\\

\midrule
\multirow{4}{*}{BUT PPG} & \multirow{4}{*}{HR}
& AnyPPG
& 202.635 & 10.675 & 204.919 & 10.632
& 203.507 & 10.615 & 212.448 & 11.282
& 187.598 & 10.222 & 198.694 & 10.550
& 337.001 & 13.130 & \underline{185.012} & 10.208
& 191.083 & \underline{10.107} & \cellcolor{bestblue}\textbf{179.911} & \cellcolor{bestblue}\textbf{9.931}
\\

&& PaPaGei-P
& 212.401 & 11.336 & 236.696 & 11.803
& 214.771 & 11.282 & 206.196 & 11.037
& 203.959 & \cellcolor{bestblue}\textbf{10.917} & 225.336 & 11.347
& 214.450 & 11.324 & \underline{203.093} & 11.007
& \cellcolor{bestblue}\textbf{201.520} & \underline{10.937} & 207.383 & 11.265
\\

&& PaPaGei-S
& 237.665 & 12.054 & 255.788 & 12.136
& 242.740 & 12.066 & \underline{214.732} & \cellcolor{bestblue}\textbf{11.140}
& 216.468 & 11.422 & 236.288 & 11.675
& 233.095 & 11.814 & 215.611 & 11.479
& \cellcolor{bestblue}\textbf{211.060} & \underline{11.257} & 215.597 & 11.318
\\

&& Pulse-PPG
& 228.978 & 11.830 & 229.720 & 12.059
& 231.559 & 11.812 & 269.051 & 12.855
& 265.401 & 12.305 & 223.648 & 11.600
& \cellcolor{bestblue}\textbf{207.564} & \cellcolor{bestblue}\textbf{11.135} & 250.855 & 12.435
& 303.459 & 13.693 & \underline{208.575} & \underline{11.205}
\\

\midrule
\multicolumn{3}{c|}{\textbf{Win Rate}}
& \multicolumn{2}{c|}{0\%}
& \multicolumn{2}{c|}{3.1\%}
& \multicolumn{2}{c|}{0\%}
& \multicolumn{2}{c|}{3.1\%}
& \multicolumn{2}{c|}{4.7\%}
& \multicolumn{2}{c|}{0\%}
& \multicolumn{2}{c|}{3.1\%}
& \multicolumn{2}{c|}{0\%}
& \multicolumn{2}{c|}{3.1\%}
& \multicolumn{2}{c}{82.8\%}\\
\bottomrule

\end{tabular}
}
\vspace{1.5mm}
\begin{minipage}{\textwidth}
\footnotesize
* ``Win Rate'' denotes the percentage of MSE and MAE comparisons in which a method achieves the best result.\\
\end{minipage}
\vspace{-6mm}
\end{table*}

\begin{table*}[t]
\centering
\caption{Model performance on classification tasks and the complete results are provided in Table \ref{tab:Complete_overall_performance_2} of Appendix \ref{appendix:overall}.}
\label{tab:overall_performance_2}
\resizebox{\textwidth}{!}{
\begin{tabular}{
c|c|c|
*{9}{cc|}
cc
}
\toprule
\multirow{2}{*}{\raisebox{-0.5ex}{\textbf{Data}}}
& \multirow{2}{*}{\raisebox{-0.5ex}{\textbf{Tasks}}}
& \multirow{2}{*}{\raisebox{-0.5ex}{\textbf{FM}}}
& \multicolumn{2}{c|}{\textbf{Linear Probe}}
& \multicolumn{2}{c|}{\textbf{Full-FT}}
& \multicolumn{2}{c|}{\textbf{LoRA}}
& \multicolumn{2}{c|}{\textbf{QLoRA}}
& \multicolumn{2}{c|}{\textbf{PFEC}}
& \multicolumn{2}{c|}{\textbf{Taylor-FO}}
& \multicolumn{2}{c|}{\textbf{FPGM}}
& \multicolumn{2}{c|}{\textbf{SRR-GR}}
& \multicolumn{2}{c|}{\textbf{Less-is-More}}
& \multicolumn{2}{c}{\textbf{Ours}}
\\

& & &
F1 $\uparrow$ & AUROC $\uparrow$
& F1 $\uparrow$ & AUROC $\uparrow$
& F1 $\uparrow$ & AUROC $\uparrow$
& F1 $\uparrow$ & AUROC $\uparrow$
& F1 $\uparrow$ & AUROC $\uparrow$
& F1 $\uparrow$ & AUROC $\uparrow$
& F1 $\uparrow$ & AUROC $\uparrow$
& F1 $\uparrow$ & AUROC $\uparrow$
& F1 $\uparrow$ & AUROC $\uparrow$
& F1 $\uparrow$ & AUROC $\uparrow$
\\
\midrule

\multirow{4}{*}{BUT PPG} & \multirow{4}{*}{SQI}
& AnyPPG
& 0.622 & 0.799 & 0.599 & 0.724
& 0.645 & 0.793 & 0.579 & 0.626
& 0.678 & \cellcolor{bestblue}\textbf{0.811} & 0.632 & 0.734
& 0.622 & 0.755 & \underline{0.697} & 0.800
& 0.695 & 0.799 & \cellcolor{bestblue}\textbf{0.707} & \underline{0.806}
\\

&& PaPaGei-P
& 0.535 & \underline{0.736} & 0.595 & 0.682
& 0.475 & 0.685 & 0.558 & 0.579
& 0.585 & 0.628 & \underline{0.599} & 0.712
& 0.544 & 0.690 & 0.581 & 0.598
& \underline{0.599} & 0.633 & \cellcolor{bestblue}\textbf{0.669} & \cellcolor{bestblue}\textbf{0.766}
\\

&& PaPaGei-S
& 0.544 & \underline{0.714} & 0.606 & 0.708
& 0.491 & 0.697 & \underline{0.629} & 0.686
& 0.614 & 0.645 & 0.612 & 0.701
& 0.594 & 0.695 & 0.600 & 0.633
& 0.616 & 0.675 & \cellcolor{bestblue}\textbf{0.686} & \cellcolor{bestblue}\textbf{0.801}
\\

&& Pulse-PPG
& 0.514 & 0.718 & 0.562 & 0.724
& 0.515 & 0.717 & 0.517 & 0.575
& 0.629 & 0.717 & 0.572 & 0.738
& 0.540 & \underline{0.764} & \underline{0.630} & 0.749
& 0.566 & 0.734 & \cellcolor{bestblue}\textbf{0.638} & \cellcolor{bestblue}\textbf{0.766}
\\

\midrule
\multirow{4}{*}{DeepBeat} & \multirow{4}{*}{AF}
& AnyPPG
& 0.544 & 0.708 & 0.651 & \underline{0.838}
& 0.468 & 0.525 & 0.609 & 0.754
& 0.642 & 0.799 & 0.646 & 0.833
& \underline{0.653} & 0.834 & 0.638 & 0.800
& 0.640 & 0.796 & \cellcolor{bestblue}\textbf{0.684} & \cellcolor{bestblue}\textbf{0.844}
\\

&& PaPaGei-P
& 0.441 & 0.608 & 0.624 & 0.785
& 0.432 & 0.545 & 0.627 & 0.754
& 0.649 & 0.781 & 0.623 & 0.769
& 0.599 & \underline{0.805} & \cellcolor{bestblue}\textbf{0.677} & 0.800
& \underline{0.662} & 0.788 & 0.649 & \cellcolor{bestblue}\textbf{0.831}
\\

&& PaPaGei-S
& 0.444 & 0.610 & 0.596 & \underline{0.789}
& 0.428 & 0.406 & 0.585 & 0.671
& 0.617 & 0.738 & 0.594 & 0.775
& 0.589 & 0.785 & \underline{0.618} & 0.735
& 0.604 & 0.727 & \cellcolor{bestblue}\textbf{0.638} & \cellcolor{bestblue}\textbf{0.809}
\\

&& Pulse-PPG
& 0.496 & 0.653 & 0.625 & 0.808
& 0.525 & 0.600 & 0.648 & 0.716
& 0.633 & 0.753 & 0.618 & 0.803
& 0.609 & 0.806 & \underline{0.657} & 0.807
& 0.643 & \underline{0.809} & \cellcolor{bestblue}\textbf{0.671} & \cellcolor{bestblue}\textbf{0.841}
\\

\midrule
\multicolumn{3}{c|}{\textbf{Win Rate}}
& \multicolumn{2}{c|}{0\%}
& \multicolumn{2}{c|}{0\%}
& \multicolumn{2}{c|}{0\%}
& \multicolumn{2}{c|}{0\%}
& \multicolumn{2}{c|}{6.3\%}
& \multicolumn{2}{c|}{0\%}
& \multicolumn{2}{c|}{0\%}
& \multicolumn{2}{c|}{6.3\%}
& \multicolumn{2}{c|}{0\%}
& \multicolumn{2}{c}{87.5\%}\\
\bottomrule

\end{tabular}
}
\vspace{-4mm}
\end{table*}
\subsection{Personalized Training}
\begin{figure*}[t]
\centering
\includegraphics[width=0.86\textwidth]{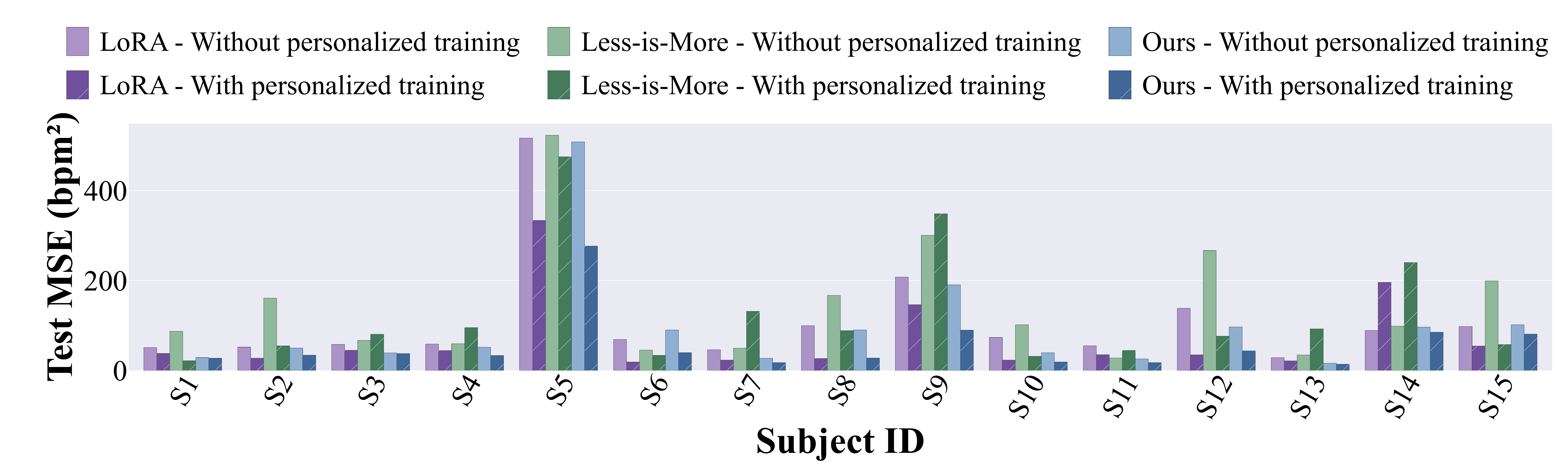}
\vspace{-2mm}
\caption{Performance of PaPaGei-S with and without personalized training.}
\label{fig:personalized_papagei-s}
\vspace{-4mm}
\end{figure*}

To further investigate the personalized modeling capability of PPG FMs,
we evaluate personalized training with SPAFT, LoRA,
and Less-is-More on the temporal PPG-DaLiA split. Each method
starts from its corresponding task-adapted checkpoint. For each
subject, the earlier half of the original test samples is divided
chronologically into training and validation sets at an 8:2
ratio, while the later half is used for evaluation.
As shown in Figure~\ref{fig:personalized_papagei-s}, SPAFT
achieves lower post-personalization test MSE than LoRA for
10 of 15 subjects and Less-is-More for 12 subjects on PaPaGei-S.
Its own MSE decreases for all 15 subjects, with particularly
visible improvements for S5 and S9. Across the same evaluation
samples, the overall MSE decreases from 96.643 to 56.648 for
SPAFT (41.38\%), compared with 110.504 to 73.173 for LoRA
(33.78\%) and 146.365 to 127.635 for Less-is-More (12.80\%)
(see Table~\ref{tab:personalized_training_improvement} in
Appendix~\ref{app:personalized_adaptation}). These results
illustrate the potential of the parameter paths selected by
SPAFT for personalized adaptation, as SPAFT consistently improves performance across all subjects and achieves the lowest overall MSE after personalized training on PaPaGei-S. Full settings and
results for all four PPG FMs are provided in
Appendix~\ref{app:personalized_adaptation}.

\subsection{Ablation Studies}
To better understand the contribution of each design in SPAFT, we conduct ablation studies on 
\begin{wraptable}{r}{0.46\textwidth}
\vspace{-3mm}
\centering
\caption{SPAFT component ablation study.}
\label{tab:ablation}
\resizebox{0.46\textwidth}{!}{
\begin{tabular}{llcccc}
    \toprule

    \multirow{2}{*}{\textbf{Model}}
    & \multirow{2}{*}{\textbf{Variant}}
    & \multicolumn{2}{c}{\textbf{UCI-BP DBP}}
    & \multicolumn{2}{c}{\textbf{BUT PPG SQI}} \\

    \cmidrule(lr){3-4}
    \cmidrule(lr){5-6}

    &
    & \textbf{MSE} $\downarrow$
    & \textbf{MAE} $\downarrow$
    & \textbf{F1} $\uparrow$
    & \textbf{AUROC} $\uparrow$ \\

    \midrule

    \multirow{5}{*}{\textbf{PAPAGEI-P}}
    & w/o IFD
    &65.089$\pm$\textit{\scriptsize 0.416}
    &5.441$\pm$\textit{\scriptsize 0.054}
    &0.595$\pm$\textit{\scriptsize 0.069}
    &0.726$\pm$\textit{\scriptsize 0.009} \\

    & w/o SRRBS
    &65.746$\pm$\textit{\scriptsize 0.147}
    &5.506$\pm$\textit{\scriptsize 0.034}
    &0.572$\pm$\textit{\scriptsize 0.100}
    &0.737$\pm$\textit{\scriptsize 0.007} \\

    & w/o SFF
    &87.165$\pm$\textit{\scriptsize 0.225}
    &6.746$\pm$\textit{\scriptsize 0.016}
    &0.512$\pm$\textit{\scriptsize 0.049}
    &0.627$\pm$\textit{\scriptsize 0.078} \\

    & w/o ABFA
    &66.072$\pm$\textit{\scriptsize 3.360}
    &5.483$\pm$\textit{\scriptsize 0.196}
    &0.607$\pm$\textit{\scriptsize 0.017}
    &0.664$\pm$\textit{\scriptsize 0.061} \\

    & \textbf{SPAFT}
    &63.760$\pm$\textit{\scriptsize 0.730}
    &5.373$\pm$\textit{\scriptsize 0.052}
    &0.640$\pm$\textit{\scriptsize 0.057}
    &0.765$\pm$\textit{\scriptsize 0.021} \\

    \midrule

    \multirow{5}{*}{\textbf{Pulse-PPG}}
    & w/o IFD
    &55.500$\pm$\textit{\scriptsize 2.761}
    &4.936$\pm$\textit{\scriptsize 0.141}
    &0.457$\pm$\textit{\scriptsize 0.000}
    &0.718$\pm$\textit{\scriptsize 0.020} \\

    & w/o SRRBS
    &55.203$\pm$\textit{\scriptsize 0.912}
    &4.908$\pm$\textit{\scriptsize 0.046}
    &0.551$\pm$\textit{\scriptsize 0.033}
    &0.716$\pm$\textit{\scriptsize 0.060} \\

    & w/o SFF
    &56.042$\pm$\textit{\scriptsize 3.107}
    &4.923$\pm$\textit{\scriptsize 0.192}
    &0.487$\pm$\textit{\scriptsize 0.023}
    &0.722$\pm$\textit{\scriptsize 0.047} \\

    & w/o ABFA
    &59.584$\pm$\textit{\scriptsize 2.840}
    &5.170$\pm$\textit{\scriptsize0.163}
    &0.554$\pm$\textit{\scriptsize 0.074}
    &0.740$\pm$\textit{\scriptsize 0.036} \\

    & \textbf{SPAFT}
    &50.747$\pm$\textit{\scriptsize 1.465}
    &4.659$\pm$\textit{\scriptsize 0.086}
    &0.589$\pm$\textit{\scriptsize 0.052}
    &0.768$\pm$\textit{\scriptsize 0.008} \\

    \bottomrule
\end{tabular}
}
\vspace{-3mm}
\end{wraptable}
UCI-BP DBP and BUT PPG SQI. We individually remove importance-guided filter deletion (\textbf{W/O IFD}), SRR-guided block selection (\textbf{W/O SRRBS}), selective filter fine-tuning (\textbf{W/O SFF}), and adaptive blockwise fine-tuning allocation (\textbf{W/O ABFA}). As shown in Table~\ref{tab:ablation}, SPAFT achieves the best mean performance across all metrics for both PPG FMs, while removing any component consistently degrades performance. For the structural pruning stage, removing IFD causes clear performance drops. 
Removing SRRBS leads to inferior results. This suggests that structural redundancy provides useful guidance for allocating pruning opportunities across blocks, complementing task importance for within-block filter selection. For the  fine-tuning stage, removing SFF consistently leads to performance degradation. Moreover, without ABFA, the MSE on Pulse-PPG increases to 59.584, while the AUROC on PaPaGei-P decreases to 0.664. 
These results demonstrate the effectiveness of reallocating the fine-tuning budget according to the task importance estimated after pruning.

\subsection{Case Study}
\begin{wrapfigure}{r}{0.41\textwidth}
\vspace{-4mm}
\centering
\includegraphics[width=0.41\textwidth]{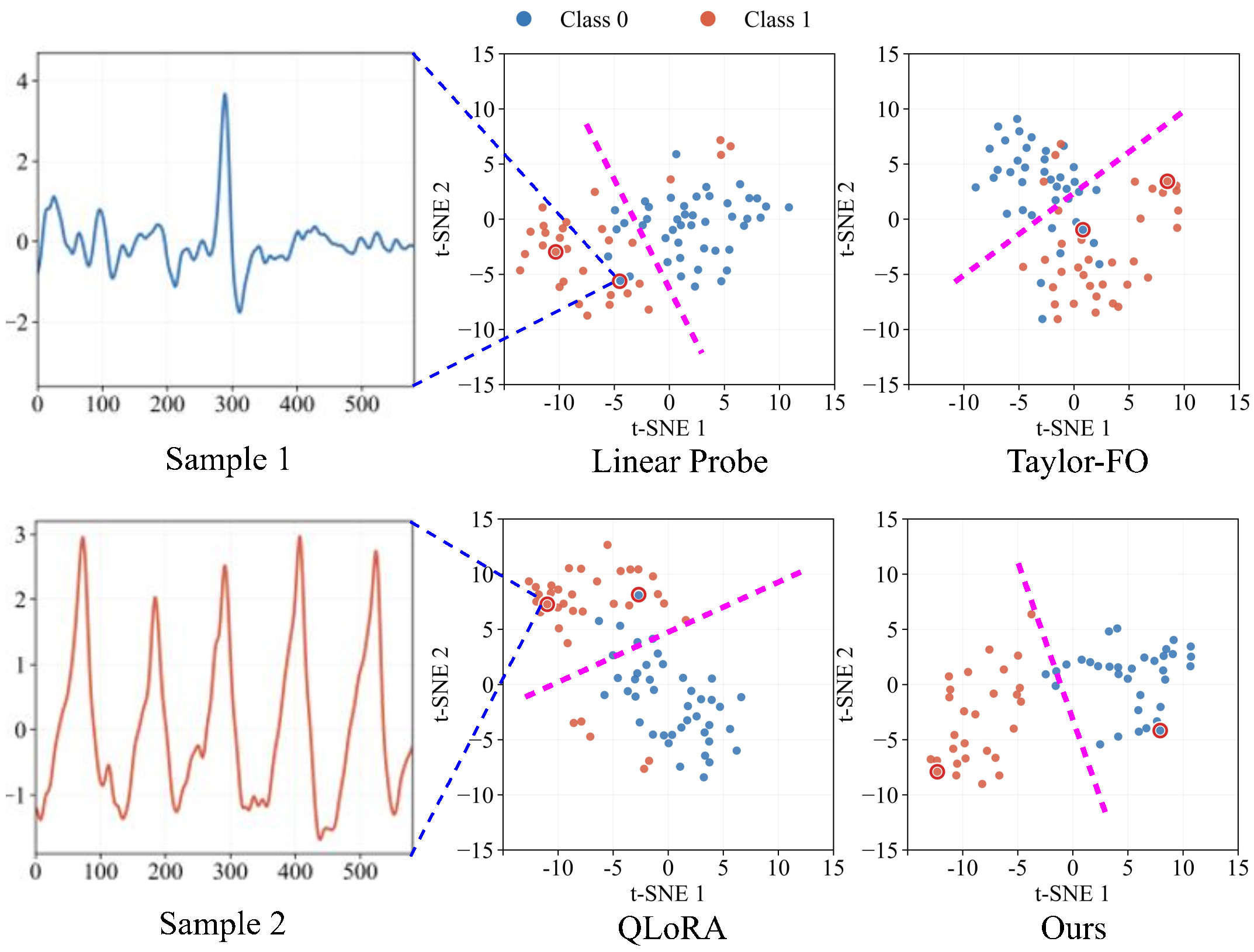}
\vspace{-6mm}
\caption{t-SNE visualization on SQI.}
\label{fig:tsne}
\vspace{-4mm}
\end{wrapfigure}
We further analyze the representations learned by different methods on the BUT PPG SQI test set. We select representative samples from the two classes for waveform visualization. Sample 1 is a low-quality waveform exhibiting pronounced morphological distortions and transient artifacts, whereas Sample 2 is a high-quality waveform showing clearer and more regular periodic pulse patterns. As shown in Figure~\ref{fig:tsne}, representative methods such as QLoRA fail to distinguish the two waveforms, with substantial inter-class overlap remaining in their feature spaces. In contrast, SPAFT produces more compact intra-class clusters and clearer inter-class separation. Samples with these distinct waveform morphologies are mapped to different class regions in our representation space, suggesting that the learned representations can effectively encode waveform differences associated with signal quality.

\section{Conclusion}
This work reveals a key challenge: how to effectively adapt pre-trained PPG FMs to achieve improved performance on downstream tasks. To address this challenge, we propose a “prune-then-finetune” framework, SPAFT, which identifies and extracts critical subnetworks through structured pruning. 
Extensive experiments demonstrate that our method can significantly improve predictive performance. 
To the best of our knowledge, this is the first work to primarily employ structured pruning to improve model performance through architectural specialization, providing a promising pathway toward unlocking the full potential of PPG FMs in real-world applications. 

\section*{Acknowledgments}
Carl Yang was not supported by any fund from China.
\section*{AI use statement}
In this work, we used generative AI tools to assist with language-related tasks, including language polishing, grammar correction, and Chinese-to-English translation. We did not use generative AI tools to generate research ideas, design experiments, analyze data, develop the core methodology, or draw scientific conclusions.

\section*{Ethics statement}
We adhere to the ICLR Code of Ethics. This study uses only de-identified data, with no direct interaction with human subjects. No personally identifiable information (PII/PHI) was accessed or disclosed, and no attempt at re-identification was made. This work is for research purposes only and is not a medical device. No conflicts of interest or sensitive sponsorships are present.



\bibliography{iclr2027_conference}
\bibliographystyle{iclr2027_conference}
\clearpage
\appendix
\clearpage
\appendix

\startcontents[appendices]

\begin{center}
    {\large\bfseries\scshape Table of Contents in Appendix}
\end{center}
\vspace{0.5em}

\printcontents[appendices]{}{1}{
    \setcounter{tocdepth}{2}
}

\vspace{1em}


\section{Implementation Details}
\subsection{Datasets and Tasks}\label{appendix:data}
Descriptions of the downstream datasets are provided below, with detailed statistics summarized in Table \ref{tab:datasets}.

\textbf{PPG-DaLiA Dataset.}  PPG-DaLiA \citep{reiss2019ppg} is a multimodal dataset designed for heart rate estimation during daily life activities. It contains synchronized physiological and motion data collected from 15 subjects performing eight different activities. These activities were specifically selected to introduce varying levels of motion artifacts under realistic conditions. The dataset includes PPG and three-axis accelerometer signals collected from wrist- and chest-worn devices, along with synchronized ECG recordings, which serve as the ground truth.

\textbf{BUT PPG Dataset.} The Brno University of Technology Smartphone PPG (BUT PPG) Dataset \citep{nemcova2021brno} is specifically designed for signal quality assessment and heart rate estimation on consumer devices. It comprises 3,888 ten-second PPG segments collected from a gender-balanced cohort of 50 subjects aged 19–76 years, covering various conditions, including rest and motion. The dataset contains smartphone-recorded PPG signals sampled at 30 Hz, synchronized with high-resolution single-lead ECG signals sampled at 1,000 Hz and tri-axial accelerometer data sampled at 100 Hz. It provides comprehensive ground-truth annotations, including binary signal quality labels and expert-validated heart rates derived from ECG QRS detection, along with auxiliary physiological metadata such as blood pressure.

\textbf{DeepBeat Dataset.} The DeepBeat dataset \citep{torres2020multi} contains PPG recordings from 132 patients with atrial fibrillation and provides publicly available validation and test splits. Consistent with the experimental protocol adopted in prior work \citep{chen2025gpt}, we used the validation split for model training and internal validation, while the official test split was reserved as an independent holdout test set.

\textbf{UCI-BP Dataset.} UCI-BP \citep{kachuee2015cuff} is a benchmark dataset for cuffless blood pressure estimation, consisting of preprocessed and quality-validated recordings derived from the MIMIC-II Waveform Database. The dataset contains synchronized fingertip PPG, Lead II ECG, and invasive arterial blood pressure (ABP, in mmHg) signals sampled at 125 Hz. The invasive ABP measurements serve as reference targets for systolic and diastolic blood pressure regression.

 \textbf{Ring-PPG Dataset.} The Ring-PPG Dataset is a newly collected PPG-centric, multi-task benchmark for physiological signal analysis, collected using safe smart rings in the stroke department of a large hospital in Asia. The dataset includes 71 patients and 19,072 PPG segments, with 635.73 hours of recordings, as well as 917 vital-sign measurements, 62 laboratory panels, 1,329 medication records, and 82 dietary records. These clinical measurements and records provide valuable reference information for multiple downstream PPG-based prediction tasks. \textit{Please note that we are continuously collecting and processing more data to build a larger dataset}.
\begin{figure*}[t]
\centering
\includegraphics[width=0.98\textwidth]{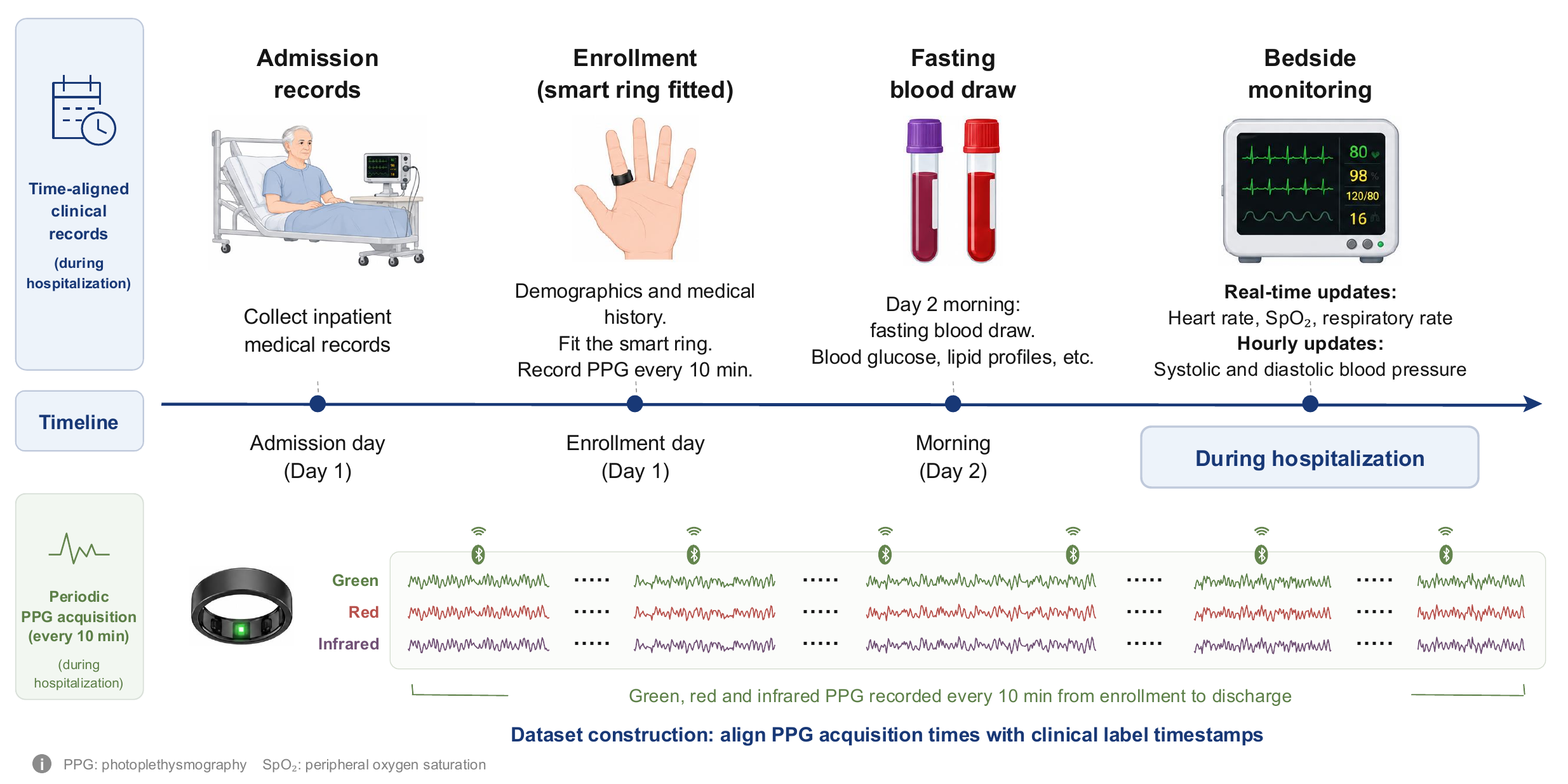}\par
\vspace{-6pt}
\caption{Overview of Ring-PPG collection and temporal alignment during hospitalization. Patients wore a smart ring that recorded green, red, and infrared PPG signals every 10 minutes from enrollment until discharge, with ring removal documented during procedures. Clinical measurements included real-time heart rate, $\mathrm{SpO_{2}}$, and respiratory rate, hourly blood pressure readings, and laboratory measurements from fasting blood samples collected on hospital day 2. PPG recordings and clinical measurements were aligned by acquisition time to construct the dataset.}
\label{appendix:ring_Data}
\vspace{-4mm}
\end{figure*}

 Figure \ref{appendix:ring_Data} illustrates the Ring-PPG collection and temporal alignment workflow during hospitalization. Upon enrollment, each patient wore a smart ring that acquired photoplethysmography (PPG) signals using green, red, and infrared light at 10-minute intervals. Concurrently, a bedside patient monitor provided real-time readings of heart rate, peripheral oxygen saturation ($\mathrm{SpO_{2}}$), and respiratory rate, while systolic and diastolic blood pressure readings were updated hourly. On the morning of hospital day 2, fasting blood samples were collected for laboratory measurements, including blood glucose and lipid profiles. 

Overall, these five datasets comprise 10 physiological analysis tasks.

\begin{table*}[htbp]
\centering
\caption{Datasets for downstream evaluation. Task type: R (regression), B (binary).}
\label{tab:datasets}

\renewcommand{\arraystretch}{1.15}
\setlength{\tabcolsep}{6pt}

\resizebox{0.82\textwidth}{!}{
\begin{tabular}{llcccc}
\toprule
\textbf{Dataset} &
\textbf{Task} &
\textbf{Task Type} &
\textbf{\#Subj.} &
\textbf{Train / Val / Test} &
\textbf{Total Samples} \\
\midrule

\multirow{2}{*}{BUT PPG}
& HR estimation
& R
& 50
& 2486 / 622 / 732
& 3840 \\

& Signal quality assessment
& B
& 50
& 2486 / 622 / 732
& 3840 \\
\midrule

PPG-DaLiA
& HR estimation
& R
& 15
& 8368 / 1829 / 2746
& 12943 \\
\midrule

DeepBeat
& Atrial fibrillation detection
& B
& N/A
& 415,025 / 103,757 / 17,617
& 536399 \\
\midrule

\multirow{2}{*}{UCI-BP}
& SBP estimation
& R
& N/A
& 167399 / 41,850 / 52,312
& 261561 \\

& DBP estimation
& R
& N/A
& 167,399 / 41,850 / 52,312
& 261561 \\
\midrule

\multirow{4}{*}{Ring-PPG (Ours)}
& SBP estimation
& R
& 15
& 172 / 23 / 44
& 239 \\

& DBP estimation
& R
& 23
& 84 / 5 / 22
& 111 \\

& HR estimation
& R
& 15
& 173 / 22 / 44
& 239 \\

& SpO$_2$ estimation
& R
& 15
& 171 / 29 / 38
& 238 \\

\bottomrule
\end{tabular}
}

\vspace{1mm}

\begin{minipage}{0.82\textwidth}
\raggedright
\footnotesize
\textit{Note:} In the Ring-PPG dataset, we place greater emphasis on SBP for stroke patients, resulting in relatively fewer DBP records.
\end{minipage}

\end{table*}
\subsection{Experimental Details}\label{appendix:exper_detail}

\textbf{Data preprocessing.}
For the public datasets, we follow the preprocessing protocol of AnyPPG~\citep{nie2026anyppg}, based on PaPaGei~\citep{pillai2025papagei}, applying 0.5--12 Hz Chebyshev band-pass filtering and segment-wise z-score normalization.
Signals are processed into 10-second windows at 125 Hz, except for DeepBeat, whose original 25-second windows are directly resampled to 1,250 points. Consistent with the AnyPPG and PaPaGei settings, Ring-PPG uses 10-second windows at 125 Hz.
All datasets thus share an input length of 1,250 samples.

\textbf{Data splits and evaluation.}
We preserve predefined train/test partitions where available.
Otherwise, we construct an 8:2 train/test split at the subject level when subject identifiers are available and at the sample level otherwise.
We further partition the training split into fitting and validation sets at an 8:2 ratio, using subject-based or target-stratified splitting as appropriate.
All methods use identical partitions, with sample counts reported in Table~\ref{tab:datasets}.
Hyperparameters and checkpoints are selected exclusively on the validation set; the selected model is then evaluated once on the held-out test set.

\textbf{Initialization and task-head fitting.}
Each encoder is initialized from its pretrained checkpoint.
Following the task-head choices and search spaces of AnyPPG~\citep{nie2026anyppg}, we use ridge regression for regression and logistic regression with the LBFGS solver for binary classification.
Heads are fitted on $\mathcal{D}_f$, with $\alpha \in \{0.1,1,10,100\}$ selected by validation MAE and $C \in \{0.01,0.1,1,10,100\}$ by validation AUROC.
We then convert the fitted head into an equivalent differentiable PyTorch module.
Feature-standardization statistics are estimated on $\mathcal{D}_f$ and remain fixed throughout pruning and fine-tuning.

\textbf{Pruning and importance estimation.} The nodes in Section~\ref{sec:model_pruning} correspond to individual Conv1 filters for PaPaGei-P/S~\citep{pillai2025papagei} and Pulse-PPG~\citep{saha2025pulse}, and complete 16-filter groups for AnyPPG~\citep{nie2026anyppg}. Physical compression removes each node together with its dependent convolutional slices and normalization entries. Consequently, $\rho$ counts original eligible nodes, whereas
$\eta$ counts retained nodes; neither directly specifies a
scalar-parameter fraction.
Following \citep{Convolutional2021}, we set $w_1=0.35$ and
$w_2=0.65$ when computing structural redundancy on the
intra-layer graphs in Eq.~\ref{equation:r}.
Both importance-estimation passes use the complete $\mathcal D_f$
in micro-batches of 256 with fixed model parameters and
BatchNorm states. Regression scoring uses MSE in the original target scale. For classification, sample averages in importance estimation use inverse-square-root class-frequency weights, normalized over $\mathcal D_f$. These weights are estimated only from $\mathcal D_f$ and are not applied to the fine-tuning classification loss.

\textbf{Adaptive partial fine-tuning.} The integer allocation in Eq. \ref{eq:quota} is implemented using capped largest-remainder allocation after reserving one node per layer. Each learning-rate candidate starts independently from the corresponding compact model and fitted task head. For a fixed compact model and fixed $(\eta,\gamma)$, candidates share the same post-pruning scores and selected nodes. To enforce the update constraint, unselected parameter entries are restored after each optimizer step and their optimizer states are masked. Regression freezes BatchNorm affine parameters and running statistics. Classification additionally updates all BatchNorm affine parameters and running statistics; accordingly, the allowed parameter set  in Eq. \ref{eq:fine-object} includes these affine parameters as well as the selected convolutional paths.

\textbf{Optimization.}
Fine-tuning uses AdamW for at most 10 epochs with early-stopping patience of three epochs.
The head and selected backbone parameters share a learning rate, with weight decay of zero and $10^{-4}$, respectively.
We accumulate micro-batches of 256 toward an effective batch size of 1,024.
Regression uses the standardized Smooth L1 loss in Eq.~\ref{eq:l1}, gradient clipping at 1.0, and 5\% linear warm-up followed by cosine decay.
Classification uses unweighted binary cross-entropy with logits and a constant learning rate.
Fine-tuning checkpoints are selected by validation RMSE, breaking ties by MAE, for regression, and by validation AUROC for classification. The pseudocode for structural pruning and adaptive fine-tuning is provided in Algorithms \ref{alg:srr_pruning} and \ref{alg:adaptive_finetuning}, respectively.

\textbf{Hyperparameter search.}
Table~\ref{tab:spaft_impl_1} summarizes the optimization settings and common hyperparameter search spaces, while Table~\ref{tab:spaft_impl_2} specifies the task-dependent fine-tuning ratio domains and BatchNorm policies.
We generally use grid search for hyperparameter tuning.
For DeepBeat AF, we use Optuna's TPE sampler over the same
discrete search domains, with 30 completed trials per backbone,
to limit the computational cost associated with its large
dataset size.
Additional fine-tuning ratio sweeps and the 1\% fine-tuning setting are reported in Appendices \ref{appendix:parameter} and \ref{appendix:ada}.

We conduct experiments using 4 NVIDIA A100 GPUs and 4 NVIDIA 5090 GPUs. The corresponding pseudocodes for pruning and fine-tuning are presented in Algorithms \ref{alg:srr_pruning} and \ref{alg:adaptive_finetuning}, respectively.

\begin{table*}[t]
\centering
\caption{SPAFT optimization settings and common search domains.}
\label{tab:spaft_impl_1}
{\small
\setlength{\tabcolsep}{5pt}
\renewcommand{\arraystretch}{1.15}
\begin{tabular}{@{}p{0.32\linewidth}p{0.58\linewidth}@{}}
\hline
\multicolumn{1}{c}{\textbf{Setting}} & \multicolumn{1}{c}{\textbf{Value or search domain}} \\
\hline
Cosine threshold $\tau$ & $\{0.1,0.2,0.3,0.4,0.5\}$ \\
Global pruning ratio $\rho$ & $\{0.05,0.10,0.15,0.20,0.30\}$ \\
Fine-tuning learning rate & $\{10^{-2},10^{-3},10^{-4},10^{-5}\}$ \\
Weight $w_1$ & $0.35$ \\
Weight $w_2$ & $0.65$ \\
Effective / micro-batch size & 1,024 / 256 \\
Maximum  fine-tuning epochs / patience & 10 / 3 \\
Backbone / head weight decay & $10^{-4}$ / 0 \\
\hline
\end{tabular}}
\end{table*}

\begin{table*}[t]
\centering
\caption{Task-specific fine-tuning ratio domains and BatchNorm policies for SPAFT.}
\label{tab:spaft_impl_2}
{\small
\setlength{\tabcolsep}{5pt}
\renewcommand{\arraystretch}{1.18}
\resizebox{0.96\textwidth}{!}{
\begin{tabular}{@{}p{0.57\linewidth}p{0.18\linewidth}p{0.16\linewidth}@{}}
\hline
\multicolumn{1}{c}{\textbf{Dataset--task pairs}} & \multicolumn{1}{c}{\textbf{fine-tuning ratio domain $\eta$}} & \multicolumn{1}{c}{\textbf{BatchNorm}} \\
\hline
UCI-BP (SBP, DBP); PPG-DaLiA (HR); BUT PPG (HR) & \multicolumn{1}{c}{$\{0.10\}$} & \multicolumn{1}{c}{Frozen} \\
Ring-PPG (HR, SBP, SpO$_2$) & \multicolumn{1}{c}{$\{0.10\}$} & \multicolumn{1}{c}{Frozen} \\
Ring-PPG (DBP) & \multicolumn{1}{c}{$\{0.10,0.20,0.30\}$} & \multicolumn{1}{c}{Frozen} \\
BUT PPG (SQI); DeepBeat (AF) & \multicolumn{1}{c}{$\{0.10,0.20,0.30\}$} & \multicolumn{1}{c}{Updated} \\
\hline
\end{tabular}}}
\end{table*}

\begin{algorithm}[t]
\caption{Structure-Redundancy-Guided Pruning}
\label{alg:srr_pruning}
\begin{algorithmic}[1]

\Require Unpruned model $F_{\theta_0}$ with a fitted task head,
         fitting set $\mathcal{D}_f$, global pruning ratio $\rho$,
         cosine threshold $\tau$
\Ensure Compact model $F_{\theta_c}$ and retained unit sets $\{V_\ell\}$

\State $V_\ell \gets \{\text{all original unit indices in layer } \ell\}$,
       $\forall \ell \in L$

\State $K \gets
\left\lceil
\rho \sum_{\ell \in L} |V_\ell|
\right\rceil$

\State $s \gets
\mathrm{Importance}(F_{\theta_0},\mathcal{D}_f)$
\Comment{Eq.~\ref{equation:s}}

\State $R_\ell \gets \mathrm{SRR}(V_\ell;\tau)$,
       $\forall \ell \in L$
\Comment{Eq.~\ref{equation:r}}

\For{$t=1$ \textbf{to} $K$}

    \State $\ell^{*} \gets
    \arg\max_{\ell:\,|V_\ell|>1} R_\ell$

    \State $j^{*} \gets
    \arg\min_{j\in V_{\ell^{*}}} s_{\ell^{*}j}$

    \State $V_{\ell^{*}} \gets
    V_{\ell^{*}}\setminus\{j^{*}\}$

    \State $R_{\ell^{*}} \gets
    \mathrm{SRR}(V_{\ell^{*}};\tau)$

\EndFor

\State $F_{\theta_c} \gets
\mathrm{Compact}
\left(
F_{\theta_0},
\{V_\ell\}_{\ell\in L}
\right)$

\State \Return $F_{\theta_c},\{V_\ell\}_{\ell\in L}$

\end{algorithmic}
\end{algorithm}
\begin{algorithm}[t]
\caption{Task-Adaptive Fine-Tuning of the Compact Model}
\label{alg:adaptive_finetuning}
\begin{algorithmic}[1]

\Require Compact model $F_{\theta_c}$, retained units $\{V_\ell\}$,
fitting set $\mathcal{D}_f$, validation set $\mathcal{D}_v$,
fine-tuning ratio $\gamma$,
training configuration $\xi$
\Ensure Validation-selected model $F^{*}$

\State Recompute importance scores $\tilde{s}$ on $F_{\theta_c}$
using $\mathcal{D}_f$ and Eq.~\ref{eq:restimation}

\For{each layer $\ell \in L$}
    \State $n_\ell \gets |V_\ell|$
\EndFor
\State $N_c \gets \sum_{\ell \in L} n_\ell$
\State $\mathrm{M} \gets \lceil \gamma N_c \rceil$

\State $\mathcal{I} \gets
\operatorname{GlobalTopM}
\left(
    \{(\ell,j,\tilde{s}_{\ell j}) :
    \ell \in L,\, j \in V_\ell\},
    \mathrm{M}
\right)$

\For{each layer $\ell \in L$}
    \State $d_\ell \gets
    \dfrac{
        |\{j : (\ell,j) \in \mathcal{I}\}|
    }{n_\ell}$
\EndFor

\State $\omega_\ell \gets
\dfrac{d_\ell}{
    \sum_{k \in L} d_k
}$
\quad $\forall\,\ell \in L$

\State $\mathbf{q} \gets
\mathbf{1} +
\operatorname{Allocate}
\left(
    \mathrm{M}-L,
    \boldsymbol{\omega},
    \mathbf{n}-\mathbf{1}
\right)$

\For{each layer $\ell \in L$}
    \State $U_\ell \gets
    \operatorname{Top}_{q_\ell}
    \{j \in V_\ell \mid \tilde{s}_{\ell j}\}$
\EndFor

\State Fine-tune the paths associated with
$\{U_\ell\}_{\ell \in L}$
and the task head on $\mathcal{D}_f$ according to $\xi$

\State $F^{*} \gets$ checkpoint with the best validation
performance on $\mathcal{D}_v$

\State \Return $F^{*}$

\end{algorithmic}
\end{algorithm}
\subsection{Baseline Details}
\label{app:baseline_details}

We provide detailed descriptions of all baseline methods used in our experiments.
The baselines are organized into five categories:
conventional adaptation methods, pruning-based adaptation methods, distillation-based adaptation method, time series foundation models and transformer-based PPG FMs.

\subsubsection{Conventional Adaptation Methods}

\begin{itemize}

    \item \textbf{Linear Probe:} freezes the pretrained PPG foundation model and optimizes only the downstream prediction head. It evaluates the direct transferability of representations learned during pretraining without modifying the pretrained backbone.

    \item \textbf{Full-FT:} follows the standard downstream adaptation procedure by jointly updating all parameters of the pretrained backbone together with the task-specific prediction head. It represents the conventional strategy of adapting the entire pretrained parameter space to each downstream PPG task.

    \item \textbf{LoRA~\citep{hu2022lora}:}
    freezes the pretrained backbone and introduces trainable low-rank adaptation modules.
    During downstream adaptation, only the injected low-rank parameters and the task-specific prediction head are optimized, reducing the number of trainable parameters compared with Full-FT.

    \item \textbf{QLoRA~\citep{dettmers2023qlora}:}
    further combines low-rank adaptation with quantization of the pretrained backbone.
    The quantized pretrained parameters remain frozen, while the low-rank adaptation parameters and the task-specific prediction head are optimized for each downstream task, further reducing the memory overhead of fine-tuning.

\end{itemize}

\subsubsection{Pruning-based Adaptation Methods}

\begin{itemize}

    \item \textbf{PFEC~\citep{li2017pruning}:} is a magnitude-based structured pruning method that measures the importance of each convolutional filter using its $\ell_1$-norm. Filters with smaller $\ell_1$-norms are regarded as less important and are removed to construct a compact network. The retained backbone is subsequently fine-tuned to recover downstream predictive performance.

    \item \textbf{Taylor-FO~\citep{molchanov2019importance}:} estimates the importance of a filter according to the change in task loss induced by its removal. Specifically, we adopt the first-order Taylor approximation, which uses gradient information to efficiently estimate the loss sensitivity of individual pruning units. Filters with lower estimated importance are removed, after which all retained backbone parameters are fine-tuned following the original fine-tuning strategy.

    \item \textbf{FPGM~\citep{he2019filter}:} identifies redundant filters according to their geometric relationships rather than relying solely on parameter magnitude. It computes the geometric median of filters within a layer and removes filters that are close to the geometric median, as these filters are considered more replaceable by the remaining ones. The resulting compact model is then fine-tuned for performance recovery.

    \item \textbf{SRR-GR~\citep{Convolutional2021}:} performs layer-adaptive pruning by explicitly measuring structural redundancy. It constructs a graph for each convolutional layer and estimates layer redundancy using graph structural properties, thereby prioritizing pruning in layers with higher redundancy. Within the selected layers, less important filters are removed, and the pruned network is subsequently fine-tuned.

    \item \textbf{Less-is-More~\citep{zhao2026less}:} is a task-aware structured pruning framework proposed for specializing time-series foundation models. It estimates the importance of pruning units according to their influence on the downstream task loss, using a loss-guided criterion with a second-order Taylor approximation and Fisher-information-based estimation. The task-irrelevant units are progressively removed to obtain a compact task-specific subnetwork, after which all retained parameters are fine-tuned on the downstream data.

\end{itemize}

\subsubsection{Distillation-based Adaptation Method}

\begin{itemize}

    \item \textbf{PPG-DISTILL~\citep{ni2025ppgdistill}:} is a knowledge distillation framework specifically designed for PPG signals. It transfers knowledge from a pretrained PPG foundation model to a lightweight student through prediction- and feature-level distillation, together with patch-level morphology and rhythm distillation that preserve local waveform characteristics and inter-patch temporal structures. During downstream adaptation, the teacher model remains frozen while the student model is optimized using the supervised and distillation objectives.

\end{itemize}

\subsubsection{Time Series Foundation Models}

\begin{itemize}

    \item \textbf{Chronos~\citep{ansari2024chronos}:} is a pretrained probabilistic time-series modeling framework that tokenizes time-series values using scaling and quantization into a fixed vocabulary and trains Transformer-based language model architectures on the resulting tokens via the cross-entropy loss. It supports zero-shot forecasting on unseen datasets without task-specific fine-tuning.

    \item \textbf{MOMENT~\citep{goswami2024moment}:} is a family of open-source foundation models for general-purpose time-series analysis, pretrained using a masked time-series prediction task on large amounts of time-series data drawn from diverse domains. It breaks time series into disjoint patches and learns representations by reconstructing randomly masked patches with a Transformer encoder and a lightweight reconstruction head. For downstream tasks, MOMENT can be used through end-to-end fine-tuning, linear probing, or zero-shot inference depending on the task.

\end{itemize}

\subsubsection{Transformer-based PPG Foundation Models}

\begin{itemize}

    \item \textbf{GPT-PPG~\citep{chen2025gpt}:} is a generative pre-trained Transformer foundation model specifically tailored to continuous PPG signals. It adapts the standard GPT architecture to patch-wise PPG modeling and employs a logit-Laplace distribution loss for generative signal modeling during pre-training. For downstream adaptation, GPT-PPG adopts a mixed-objective fine-tuning framework that jointly optimizes the task-specific objective loss and the signal modeling loss, allowing the model to adapt to the PPG distribution of each downstream dataset.

    \item \textbf{SIGMA-PPG~\citep{guo2026sigmappg}:} is a generative PPG foundation model based on a Statistical-prior Informed Generative Masking Architecture. It first employs a Spectrum-Aware Semantic Tokenizer to map continuous PPG signals into discrete semantic tokens, using spectral reconstruction and a semantic consistency constraint to preserve physiological characteristics while reducing noise-induced redundancy. It then performs Prior-Guided Masked Generative Pre-training. The pretrained backbone can subsequently be adapted to downstream tasks through full fine-tuning or evaluated with a frozen encoder under linear probing.

\end{itemize}

For all pruning-based methods, we follow their respective pruning and fine-tuning procedures.
For each combination of a PPG foundation model and downstream task, we independently tune the global pruning ratio and the method-specific fine-tuning hyperparameters on the validation set.
To ensure a controlled comparison, all methods use the same common training configurations, including the optimizer, batch size, and number of fine-tuning epochs.
\section{Additional Empirical Study}\label{appendix:empirical}
\subsection{Insignificant weight and Outputs of filters}
\begin{figure*}[t]
\centering
\includegraphics[width=0.98\textwidth]{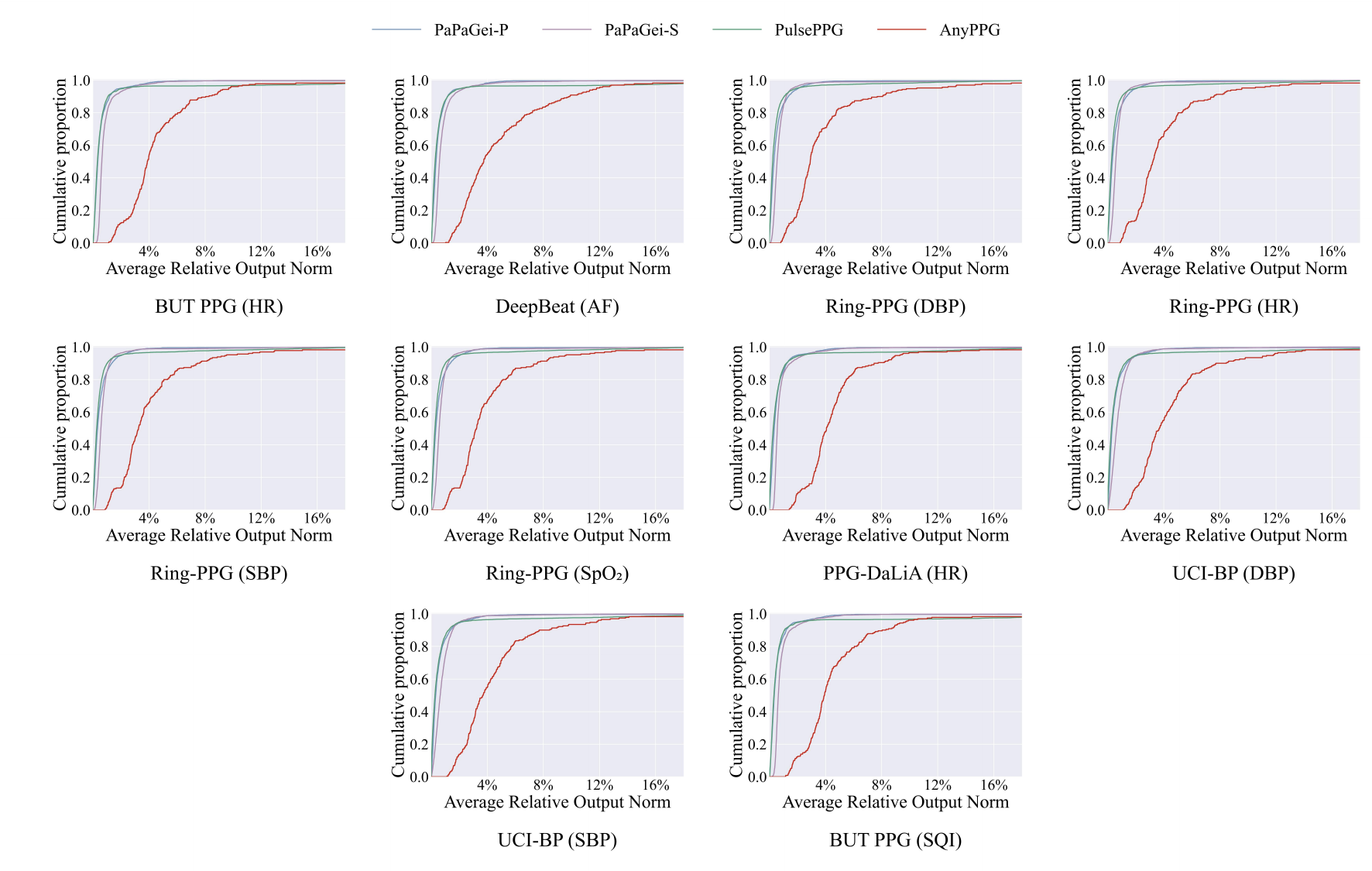}\par
\vspace{-6pt}
\caption{Cumulative distribution of the average relative output norm of filters.}
\label{appendix:cumu}
\vspace{-4mm}
\end{figure*}

We analyze the average relative output norm of each filter in four PPG FMs. As shown in Figure \ref{appendix:cumu}, Pulse-PPG and PaPaGei do not actually utilize all filters when performing specific tasks; instead, a large number of low-contribution structures exist, indicating that these models exhibit structural redundancy. The curve of AnyPPG is shifted overall to the right, mainly due to its inherent architectural characteristics, where the statistical unit is a group consisting of 16 filters.

\begin{figure*}[t]
\centering
\includegraphics[width=0.98\textwidth]{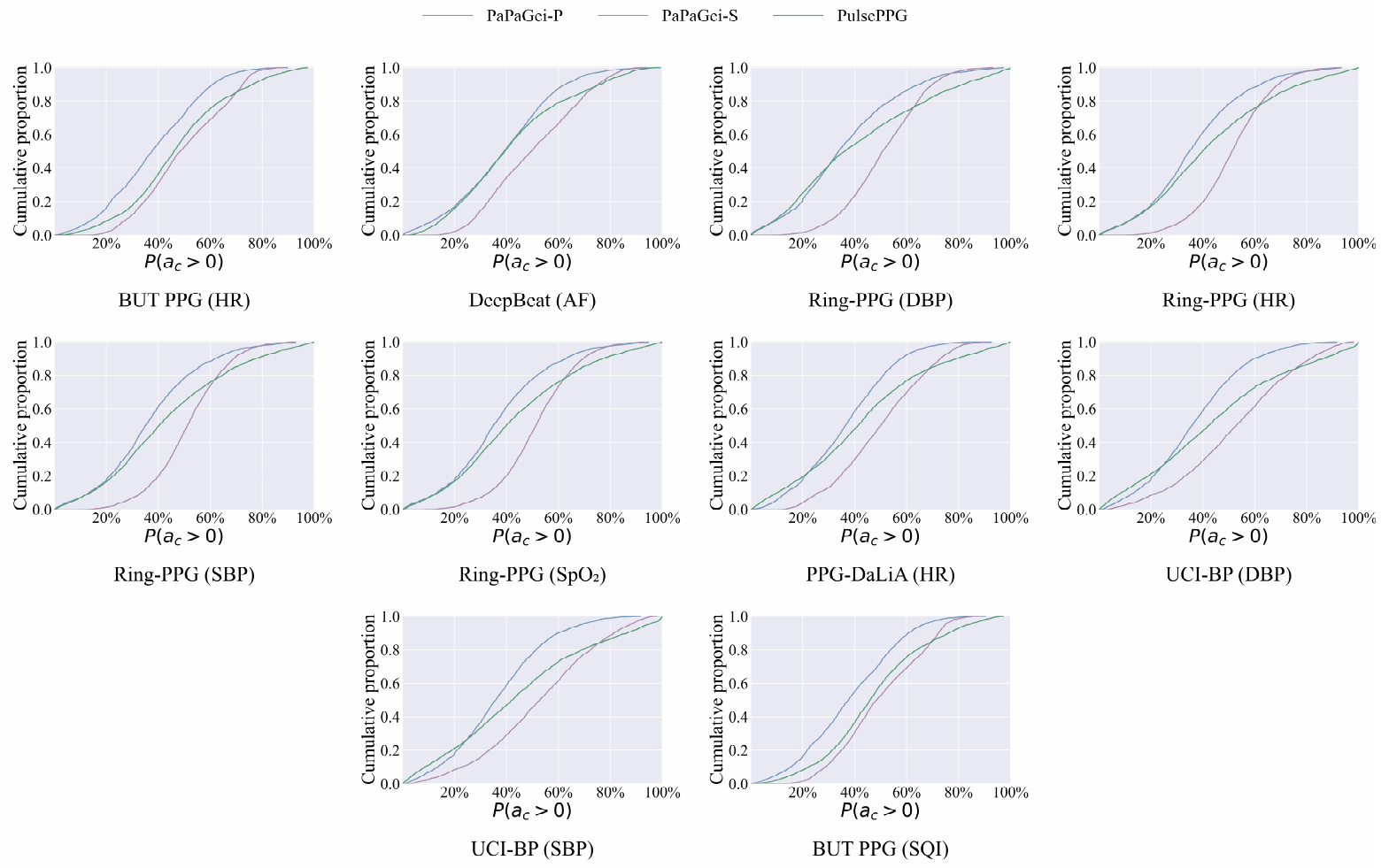}\par
\vspace{-6pt}
\caption{Cumulative distribution of the activation probability of a single filter.}
\label{appendix:activation}
\vspace{-4mm}
\end{figure*}

\begin{figure*}[t]
\centering
\includegraphics[width=0.98\textwidth]{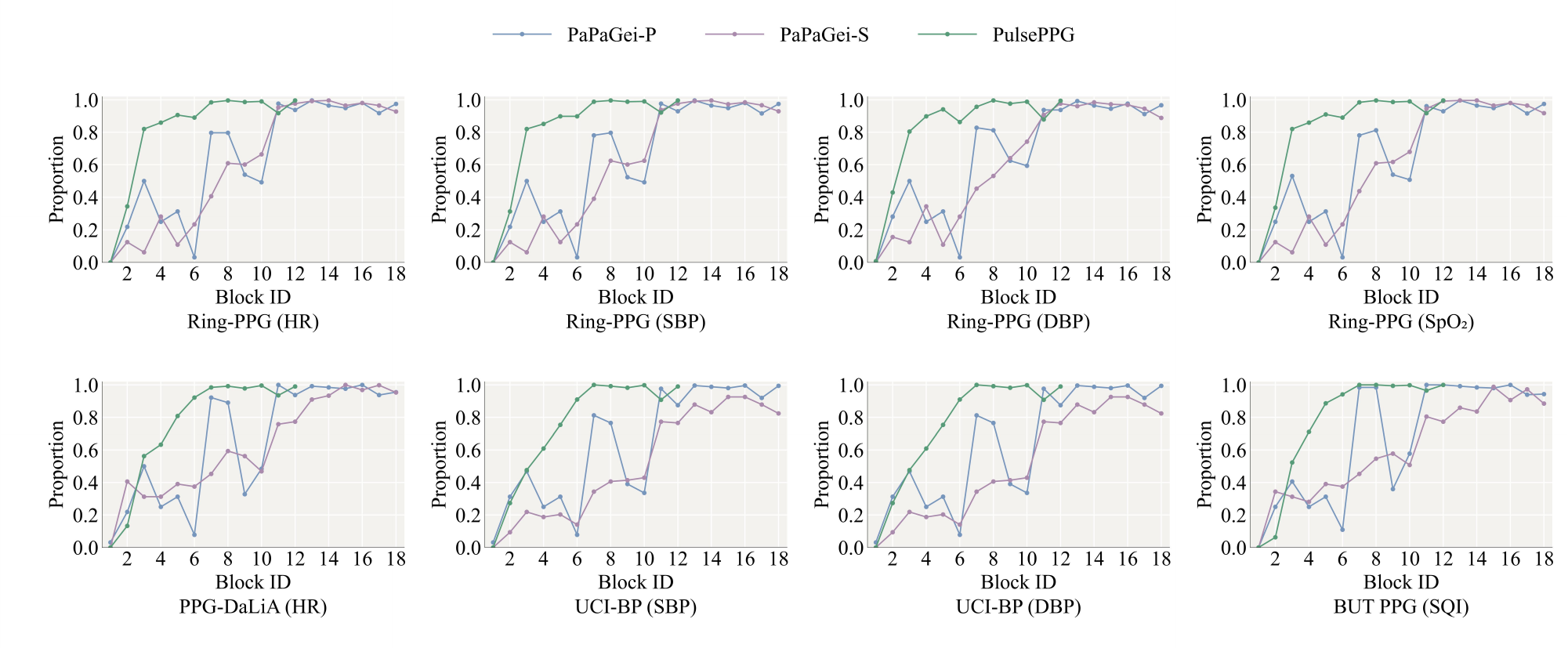}\par
\vspace{-6pt}
\caption{Cumulative distribution of the average relative output norm of a single block.}
\label{appendix:block_activation}
\vspace{-4mm}
\end{figure*}

\begin{figure*}[t]
\centering
\includegraphics[width=0.98\textwidth]{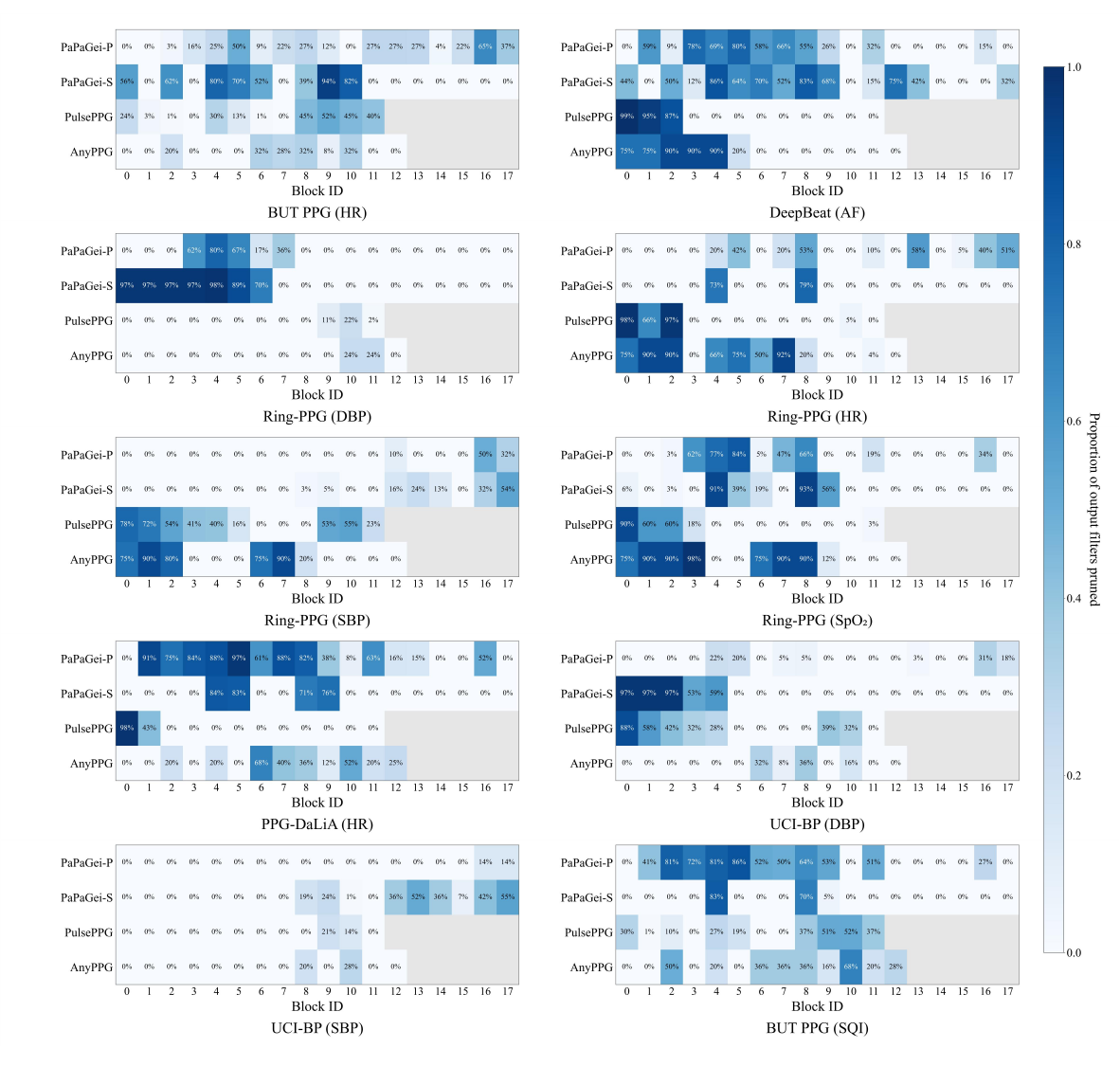}\par
\vspace{-6pt}
\caption{Output filter pruning ratio.}
\label{appendix:pruning}
\vspace{-4mm}
\end{figure*}
\subsection{Sparse Activations in filters}
As shown in Figure \ref{appendix:activation}, the three pretrained PPG FMs exhibit stable yet distinctly different filter activation patterns across downstream tasks. The activation probability distribution of PaPaGei-P is overall shifted toward the low-probability region, indicating that it contains more infrequently activated filters. In contrast, the distribution of PaPaGei-S is shifted to the right, exhibiting a denser filter utilization pattern, while PulsePPG generally lies between the two. This trend is relatively consistent across multiple datasets and different tasks. Meanwhile, unlike the large number of zero-activation filters reported in time-series foundation models, these three PPG FMs contain almost no substantial proportion of completely inactive filters, suggesting that their redundancy is more likely manifested as infrequent activation rather than simply as dead filters. Since AnyPPG performs computations in groups of 16 filters, we do not include its activation statistics.

\subsection{Task-Specific Subnetwork}
As shown in Figure \ref{appendix:block_activation} each model exhibits highly consistent patterns across datasets. Specifically, PaPaGei-P shows the most pronounced inter-layer differences, with generally low contributions in deeper layers. PaPaGei-S exhibits the smoothest trend with increasing depth. PulsePPG shows the strongest depth dependence, with the most pronounced redundancy in the deeper layers.

As shown in Figure \ref{appendix:pruning}, we analyze the model sparsification results by comparing the pruning ratios of output filters. Overall, in most cases, the models exhibit more aggressive pruning in the shallower layers, while non-negligible pruning is also observed in the deeper layers. In contrast, for PaPaGei on Ring-PPG (SBP) and UCI-BP, pruning is more concentrated after the 12th layer. Furthermore, we observe that pruning patterns vary across different tasks for the same model, as well as across different models for the same task, further supporting the rationale of our structured pruning strategy.
\section{Additional Experimental Results}\label{appendix:add_experimental}
\subsection{Overall Performance}\label{appendix:overall}
Tables \ref{tab:Complete_overall_performance_1} and \ref{tab:Complete_overall_performance_2} report the complete model performance, including the mean and standard deviation. 

We then compare two time-series foundation models (Chronos and MOMENT) and two Transformer-based PPG FMs (GPT-PPG and SIGMA-PPG). As shown in Tables \ref{appendix:other_baselines_performance_1} and \ref{appendix:other_baselines_performance_2}, our method achieves the best overall performance. In addition, general-purpose time-series models demonstrate strong performance on PPG data, whereas Transformer-based PPG FMs perform relatively poorly on downstream tasks when using conventional adaptation methods. For example, on the SBP task, MOMENT with QLoRA reduces the MSE by 19.4\% and the MAE by 11.5\% compared with SIGMA-PPG with QLoRA. This further validates the effectiveness of our proposed structured pruning and adaptive fine-tuning approach. Furthermore, we compare our method with the state-of-the-art distillation method for PPG FMs, PPG-Distill. As shown in Tables \ref{tab:compared_with_PPG-Distill_performance_regression} and \ref{tab:compared_with_PPG-Distill_performance_classification}, our method consistently outperforms PPG-Distill on both classification and regression tasks. These extensive experimental results demonstrate the effectiveness of our method.
\begin{table*}[t]
\centering
\caption{Comparison of our SOTA model with time-series foundation models and Transformer-based PPG foundation models on regression tasks.}
\label{appendix:other_baselines_performance_1}
\resizebox{0.78\textwidth}{!}{
\begin{tabular}{
c|c|c|
*{3}{cc|}
cc
}
\toprule
\multirow{2}{*}{\textbf{Data}}
& \multirow{2}{*}{\textbf{Tasks}}
& \multirow{2}{*}{\textbf{FM}}
& \multicolumn{2}{c|}{\textbf{Linear Probe}}
& \multicolumn{2}{c|}{\textbf{Full-FT}}
& \multicolumn{2}{c|}{\textbf{LoRA}}
& \multicolumn{2}{c}{\textbf{QLoRA}}
\\

& & &
MSE $\downarrow$ & MAE $\downarrow$
& MSE $\downarrow$ & MAE $\downarrow$
& MSE $\downarrow$ & MAE $\downarrow$
& MSE $\downarrow$ & MAE $\downarrow$
\\
\midrule

\multirow{10}{*}{UCI-BP}&\multirow{5}{*}{SBP}
& MOMENT
& 365.007 & 14.971
& 301.557 & 13.470
& 255.651 & 12.765
& \underline{247.612} & \underline{12.534}
\\

&& Chronos-T5-base
& 421.585 & 16.597
& 339.169 & 14.015
& 301.742 & 13.564
& 294.418 & 13.296
\\

&& GPT-PPG
& 453.942 & 17.175
& 395.618 & 15.525
& 362.828 & 14.711
& 387.024 & 15.410
\\

&& SIGMA-PPG
& 442.144 & 16.835
& 342.883 & 14.247
& 365.191 & 14.743
& 353.973 & 14.367
\\

\cmidrule{4-11}
&& \textbf{Ours}
& \multicolumn{8}{c}{
MSE: $\mathbf{143.514}$
\qquad
MAE: $\mathbf{8.388}$
}
\\

\cmidrule{2-11}
& \multirow{5}{*}{DBP} & MOMENT
& 84.407 & 6.707
& 79.831 & 6.357
& 76.043 & 6.099
& 77.658 & 6.174
\\

&& Chronos-T5-base
& 101.588 & 7.569
& 99.874 & 7.256
& 88.832 & 6.859
& \underline{66.531} & \underline{5.894}
\\

&& GPT-PPG
& 102.172 & 7.598
& 88.307 & 6.742
& 82.755 & 6.595
& 86.502 & 6.659
\\

&& SIGMA-PPG
& 100.374 & 7.527
& 81.490 & 6.435
& 77.967 & 6.197
& 81.022 & 6.485
\\

\cmidrule{4-11}
&& \textbf{Ours}
& \multicolumn{8}{c}{
MSE: $\mathbf{40.842}$
\qquad
MAE: $\mathbf{4.096}$
}
\\

\midrule
\multirow{5}{*}{PPG-DaLiA}&\multirow{5}{*}{HR}
& MOMENT
& 113.288 & 7.213
& 101.678 & 6.812
& 89.981 & \underline{5.437}
& 82.522 & 5.613
\\

&& Chronos-T5-base
& 123.294 & 8.057
& 105.355 & 6.997
& 78.145 & 5.787
& \underline{77.361} & 5.665
\\

&& GPT-PPG
& 198.964 & 11.143
& 259.287 & 12.678
& 229.536 & 11.516
& 224.433 & 11.148
\\

&& SIGMA-PPG
& 158.977 & 9.304
& 156.160 & 9.402
& 188.671 & 10.152
& 186.254 & 10.133
\\

\cmidrule{4-11}
&& \textbf{Ours}
& \multicolumn{8}{c}{
MSE: $\mathbf{71.771}$
\qquad
MAE: $\mathbf{5.059}$
}
\\

\midrule

\multirow{20}{*}{Ring-PPG}&\multirow{5}{*}{HR} 
& MOMENT
& 48.160 & 5.086
& 66.584 & 6.781
& 71.630 & 6.996
& 71.617 & 6.978
\\

&& Chronos-T5-base
& 60.981 & 6.255
& 89.017 & 7.915
& 106.499 & 8.511
& 106.513 & 8.574
\\

&& GPT-PPG
& 74.648 & 6.884
& 132.293 & 9.790
& 160.015 & 10.851
& 126.699 & 9.515
\\

&& SIGMA-PPG
& $\underline{43.622}$ & $\mathbf{4.916}$
& 51.376 & $\underline{5.030}$
& 79.381 & 6.985
& 107.964 & 8.931
\\

\cmidrule{4-11}
&& \textbf{Ours}
& \multicolumn{8}{c}{
MSE: $\mathbf{43.530}$
\qquad
MAE: $5.309$
}
\\

\cmidrule{2-11}

&\multirow{5}{*}{SBP}
& MOMENT
& 303.569 & 14.533
& 244.184 & 13.509
& 174.462 & 11.335
& $\underline{174.382}$ & $11.330$
\\

&& Chronos-T5-base
& 259.487 & 13.588
& 201.496 & 11.995
& 176.510 & 11.394
& 176.546 & 11.396
\\

&& GPT-PPG
& 187.959 & 11.599
& 183.676 & 11.349
& 179.338 & 11.151
& 177.636 & \underline{11.121}
\\

&& SIGMA-PPG
& 424.379 & 15.405
& 389.565 & 14.356
& 377.425 & 14.106
& 379.584 & 14.197
\\

\cmidrule{4-11}
&& \textbf{Ours}
& \multicolumn{8}{c}{
MSE: $\mathbf{163.024}$
\qquad
MAE: $\mathbf{10.641}$
}
\\

\cmidrule{2-11}

&\multirow{5}{*}{DBP}
& MOMENT
& 246.735 & 11.868
& 221.155 & 11.589
& 207.385 & 11.747
& 233.322 & 13.075
\\

&& Chronos-T5-base
& 303.392 & 14.994
& 284.128 & 14.658
& 206.827 & 11.710
& 206.322 & 11.710
\\

&& GPT-PPG
& 154.259 & 10.411
& 228.191 & 14.635
& 157.335 & 11.475
& 197.613 & 13.663
\\

&& SIGMA-PPG
& 151.368 & 10.385
& $\underline{147.584}$ & $\underline{10.366}$
& 167.062 & 10.569
& 148.321 & 10.370
\\

\cmidrule{4-11}
&& \textbf{Ours}
& \multicolumn{8}{c}{
MSE: $\mathbf{135.673}$
\qquad
MAE: $\mathbf{9.568}$
}
\\

\cmidrule{2-11}

& \multirow{5}{*}{SpO$_2$}
& MOMENT
& 1.608 & 1.031
& 1.535 & $\underline{1.002}$
& 1.700 & 1.123
& 1.692 & 1.118
\\

&& Chronos-T5-base
& 2.900 & 1.404
& 2.785 & 1.368
& 1.652 & 1.097
& 1.656 & 1.098
\\

&& GPT-PPG
& 1.888 & 1.120
& 1.799 & 1.110
& 1.585 & 1.014
& 1.624 & 1.036
\\

&& SIGMA-PPG
& 1.740 & 1.055
& 1.977 & 1.160
& $\underline{1.526}$ & \underline{1.002}
& 1.624 & 1.027
\\

\cmidrule{4-11}
&& \textbf{Ours}
& \multicolumn{8}{c}{
MSE: $\mathbf{0.839}$
\qquad
MAE: $\mathbf{0.776}$
}
\\

\midrule

\multirow{5}{*}{BUT PPG} & \multirow{5}{*}{HR}
& MOMENT
& 195.396 & 10.595
& 189.560 & 10.386
& $\mathbf{172.417}$ & $\mathbf{9.760}$
& $\underline{172.970}$ & $\underline{9.882}$
\\

&& Chronos-T5-base
& 202.510 & 11.009
& 197.598 & 11.136
& 190.324 & 10.996
& 187.736 & 10.857
\\

&& GPT-PPG
& 212.962 & 11.279
& 204.208 & 11.144
& 213.417 & 10.939
& 204.995 & 11.103
\\

&& SIGMA-PPG
& 205.074 & 11.111
& 197.215 & 11.036
& 202.356 & 11.101
& 203.652 & 11.114
\\

\cmidrule{4-11}
&& \textbf{Ours}
& \multicolumn{8}{c}{
MSE: $177.866$
\qquad
MAE: $9.958$
}
\\
\midrule
\end{tabular}
}
\end{table*}
\begin{table*}[t]
\centering
\caption{Comparison of our SOTA model with time-series foundation models and Transformer-based PPG FMs on classification tasks.}
\label{appendix:other_baselines_performance_2}
\resizebox{0.78\textwidth}{!}{
\begin{tabular}{
c|c|c|
*{3}{cc|}
cc
}
\toprule
\multirow{2}{*}{\textbf{Data}}
& \multirow{2}{*}{\textbf{Tasks}}
& \multirow{2}{*}{\textbf{FM}}
& \multicolumn{2}{c|}{\textbf{Linear Probe}}
& \multicolumn{2}{c|}{\textbf{Full-FT}}
& \multicolumn{2}{c|}{\textbf{LoRA}}
& \multicolumn{2}{c}{\textbf{QLoRA}}
\\

& & 
& F1 $\uparrow$ & AUROC $\uparrow$
& F1 $\uparrow$ & AUROC $\uparrow$
& F1 $\uparrow$ & AUROC $\uparrow$
& F1 $\uparrow$ & AUROC $\uparrow$
\\
\midrule

\multirow{5}{*}{BUT PPG} & \multirow{5}{*}{SQI}
& MOMENT
& 0.634 & 0.813
& 0.650 & 0.815
& $\underline{0.675}$ & $\underline{0.825}$
& 0.669 & $\mathbf{0.826}$
\\

&& Chronos-T5-base
& 0.580 & 0.765
& 0.573 & 0.760
& 0.564 & 0.757
& 0.621 & 0.764
\\

&& GPT-PPG
& 0.523 & 0.660
& 0.505 & 0.657
& 0.464 & 0.657
& 0.498 & 0.711
\\

&& SIGMA-PPG
& 0.585 & 0.737
& 0.631 & 0.761
& 0.604 & 0.742
& 0.671 & 0.758
\\

\cmidrule{4-11}
&& \textbf{Ours}
& \multicolumn{8}{c}{
F1: $\mathbf{0.716}$
\qquad
AUROC: $0.813$
}
\\

\midrule

\multirow{5}{*}{DeepBeat} & \multirow{5}{*}{AF}
& MOMENT
& 0.547 & 0.717
& 0.580 & 0.765
& 0.587 & 0.740
& 0.602 & 0.740
\\

&& Chronos-T5-base
& 0.510 & 0.664
& 0.537 & 0.674
& 0.560 & 0.752
& 0.562 & 0.754
\\

&& GPT-PPG
& 0.431 & 0.570
& 0.467 & 0.669
& 0.450 & 0.650
& 0.457 & 0.621
\\

&& SIGMA-PPG
& 0.467 & 0.592
& 0.581 & $\underline{0.784}$
& 0.583 & 0.745
& $\underline{0.615}$ & 0.765
\\

\cmidrule{4-11}
&& \textbf{Ours}
& \multicolumn{8}{c}{
F1: $\mathbf{0.680}$
\qquad
AUROC: $\mathbf{0.834}$
}
\\

\midrule

\end{tabular}
}
\end{table*}
\begin{table*}[t]
\centering
\caption{Comparison with PPG-Distill on regression tasks.}
\label{tab:compared_with_PPG-Distill_performance_regression}

\resizebox{\textwidth}{!}{
\begin{tabular}{
c|c|c|
*{6}{cc|}
cc
}
\toprule
\multirow{2}{*}{\textbf{Data}}
& \multirow{2}{*}{\textbf{Tasks}}
& \multirow{2}{*}{\textbf{FM}}
& \multicolumn{2}{c|}{\textbf{Linear Probe}}
& \multicolumn{2}{c|}{\textbf{Full-FT}}
& \multicolumn{2}{c|}{\textbf{QLoRA}}
& \multicolumn{2}{c|}{\textbf{SRR-GR}}
& \multicolumn{2}{c|}{\textbf{Less-is-More}}
& \multicolumn{2}{c|}{\textbf{PPG-DISTILL}}
& \multicolumn{2}{c}{\textbf{Ours}}
\\

& & &
MSE $\downarrow$ & MAE $\downarrow$
& MSE $\downarrow$ & MAE $\downarrow$
& MSE $\downarrow$ & MAE $\downarrow$
& MSE $\downarrow$ & MAE $\downarrow$
& MSE $\downarrow$ & MAE $\downarrow$
& MSE $\downarrow$ & MAE $\downarrow$
& MSE $\downarrow$ & MAE $\downarrow$
\\
\midrule

\multirow{2}{*}{UCI-BP}
& SBP
& PaPaGei-S
& 461.959 & 17.296
& \underline{396.019} & \underline{15.678}
& 479.878 & 17.635
& 449.427 & 16.861
& 442.235 & 17.005
& 426.989 & 16.473
& \textbf{296.102} & \textbf{12.935}
\\

\cmidrule{2-17}

& DBP
& PaPaGei-S
& 104.351 & 7.697
& \underline{83.100} & \underline{6.504}
& 111.985 & 7.770
& 102.661 & 7.599
& 105.346 & 7.766
& 92.758 & 7.054
& \textbf{65.138} & \textbf{5.414}
\\

\midrule

PPG-DaLiA & HR & PaPaGei-S
& 244.569 & 12.075
& 274.732 & 12.418
& 296.118 & 13.759
& 220.111 & 11.389
& \underline{182.748} & \underline{10.174}
& 215.872 & 10.979
& \textbf{133.177} & \textbf{8.304}
\\

\midrule

\multirow{4}{*}{Ring-PPG}
& HR
& PaPaGei-S
& 82.417 & 7.474
& 90.252 & 7.721
& 71.257 & 7.156
& 91.127 & 8.159
& \underline{66.563} & \textbf{6.491}
& 89.961 & 7.220
& \textbf{64.737} & \underline{6.631}
\\

\cmidrule{2-17}

& SBP
& PaPaGei-S
& 213.565 & 12.105
& 258.791 & 13.101
& 201.459 & 12.211
& 240.100 & 12.786
& \underline{191.251} & \underline{12.000}
& 252.562 & 13.344
& \textbf{176.200} & \textbf{11.440}
\\

\cmidrule{2-17}

& DBP
& PaPaGei-S
& 237.432 & 12.517
& 459.309 & 16.044
& 207.794 & 11.798
& 362.523 & 15.010
& 265.109 & 12.947
& \textbf{186.740} & \underline{11.707}
& \underline{194.784} & \textbf{11.700}
\\

\cmidrule{2-17}

& SpO$_2$
& PaPaGei-S
& 2.339 & 1.292
& 2.374 & 1.306
& 2.325 & 1.305
& 2.260 & 1.296
& \underline{1.736} & \underline{1.113}
& 2.609 & 1.303
& \textbf{1.367} & \textbf{0.991}
\\

\midrule

BUT PPG & HR & PaPaGei-S
& 234.902 & 11.951
& 245.010 & 11.977
& 215.458 & \textbf{11.197}
& 218.412 & 11.662
& \textbf{212.663} & 11.287
& 239.410 & 12.164
& \underline{212.989} & \underline{11.223}
\\

\midrule

\end{tabular}
}
\end{table*}
\begin{table*}[t]
\centering
\caption{Comparison with PPG-Distill on classification tasks.}
\label{tab:compared_with_PPG-Distill_performance_classification}

\resizebox{\textwidth}{!}{
\begin{tabular}{
c|c|c|
*{6}{cc|}
cc
}
\toprule
\multirow{2}{*}{\textbf{Data}}
& \multirow{2}{*}{\textbf{Tasks}}
& \multirow{2}{*}{\textbf{FM}}
& \multicolumn{2}{c|}{\textbf{Linear Probe}}
& \multicolumn{2}{c|}{\textbf{Full-FT}}
& \multicolumn{2}{c|}{\textbf{QLoRA}}
& \multicolumn{2}{c|}{\textbf{SRR-GR}}
& \multicolumn{2}{c|}{\textbf{Less-is-More}}
& \multicolumn{2}{c|}{\textbf{PPG-DISTILL}}
& \multicolumn{2}{c}{\textbf{Ours}}
\\

& & &
F1 $\uparrow$ & AUROC $\uparrow$
& F1 $\uparrow$ & AUROC $\uparrow$
& F1 $\uparrow$ & AUROC $\uparrow$
& F1 $\uparrow$ & AUROC $\uparrow$
& F1 $\uparrow$ & AUROC $\uparrow$
& F1 $\uparrow$ & AUROC $\uparrow$
& F1 $\uparrow$ & AUROC $\uparrow$
\\
\midrule

BUT PPG & SQI & PaPaGei-S
& 0.584 & 0.707
& 0.639 & 0.727
& \underline{0.646} & \underline{0.732}
& 0.570 & 0.671
& 0.590 & 0.706
& 0.580 & 0.686
& \textbf{0.666} & \textbf{0.791}
\\

\midrule

Deepbeat & AF & PaPaGei-S
& 0.444 & 0.610
& 0.616 & \underline{0.786}
& 0.584 & 0.660
& \underline{0.636} & 0.749
& 0.597 & 0.723
& 0.493 & 0.754
& \textbf{0.659} & \textbf{0.815}
\\

\midrule
\bottomrule

\end{tabular}
}
\end{table*}

\subsection{Personalized Training on PPG-DaLiA}\label{appendix:personal}
\begin{figure*}[t]
    \centering
    \includegraphics[width=0.9\textwidth]{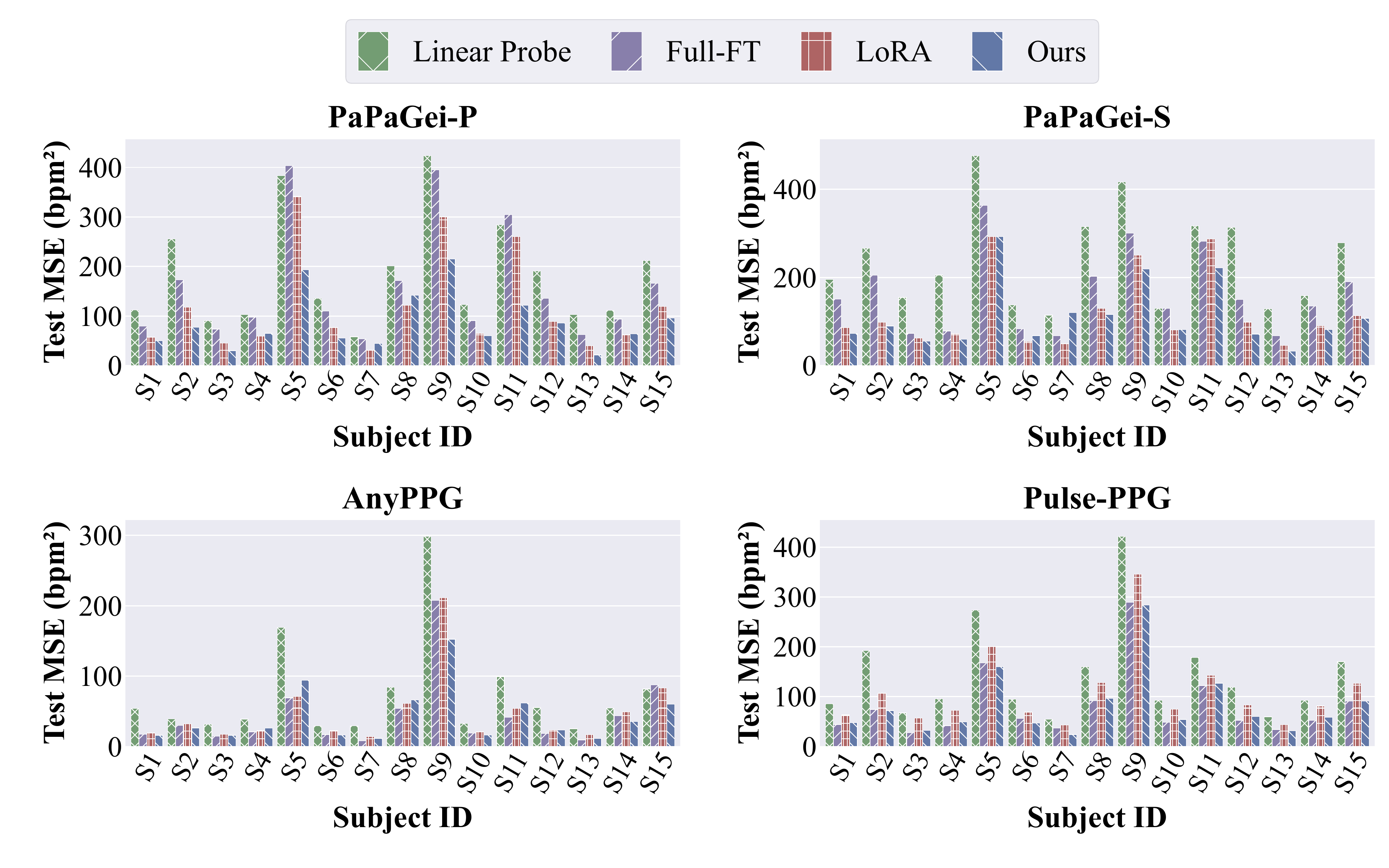}
    \vspace{-1mm}
    \caption{subject-level performance comparison in terms of MSE.}
    \label{fig:patient_mse}
\end{figure*}

\begin{figure*}[t]
    \centering
    \includegraphics[width=0.92\textwidth]{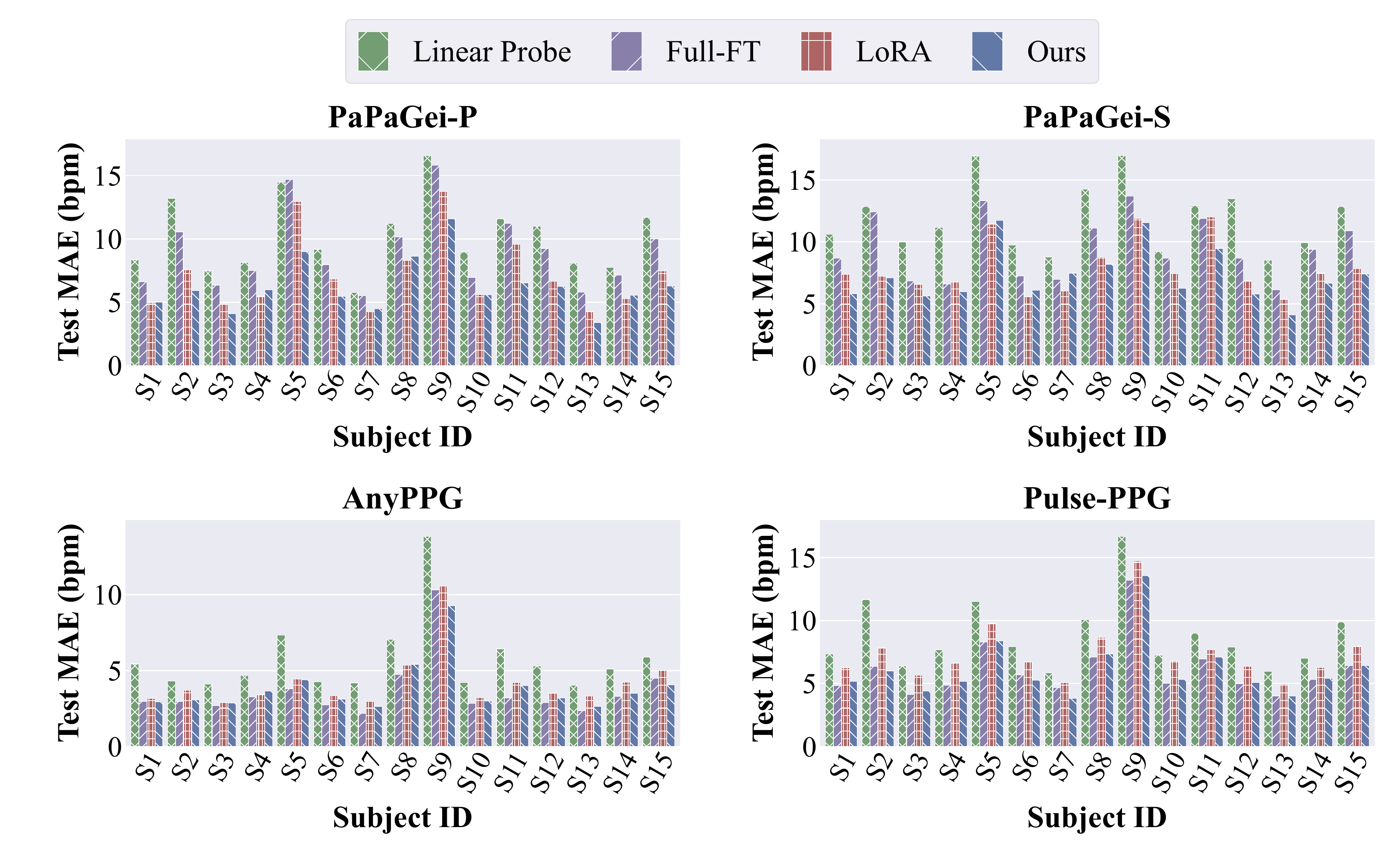}
    \caption{subject-level performance comparison in terms of MAE.}
    \label{fig:patient_mae}
\end{figure*}

\begin{wraptable}{r}{0.4\textwidth}
\vspace{-2mm}
\centering
\caption{Performance comparison.}
\label{tab:personalized_adaptation_result}
\resizebox{0.4\textwidth}{!}{
\begin{tabular}{clrr}
\toprule
\textbf{Model} &
\textbf{Method} &
\textbf{Test MSE $\downarrow$} &
\textbf{Test MAE $\downarrow$} \\

\midrule

\multirow{5}{*}{PaPaGei-P}
& Linear Probe & 184.968 & 10.172 \\
& Full-FT      & 160.699 & 8.988 \\
& LoRA         & 119.482 & 7.146 \\
& Less-is-More         & \underline{102.964} & \underline{6.637} \\
& Ours         & \textbf{88.019} & \textbf{6.226} \\

\midrule

\multirow{5}{*}{PaPaGei-S}
& Linear Probe & 240.210 & 11.845 \\
& Full-FT      & 166.317 & 9.504 \\
& LoRA         & \underline{122.317} & \underline{7.963} \\
& Less-is-More         & 159.275 & 9.254 \\
& Ours         & \textbf{114.125} & \textbf{7.292} \\

\midrule

\multirow{5}{*}{AnyPPG}
& Linear Probe & 75.138 & 5.744 \\
& Full-FT      & \underline{43.454} & \textbf{3.634} \\
& LoRA         & 47.533 & 4.209 \\
& Less-is-More         & 66.720 & 4.444 \\
& Ours         & \textbf{42.134} & \underline{3.837} \\

\midrule

\multirow{5}{*}{Pulse-PPG}
& Linear Probe & 143.267 & 8.751 \\
& Full-FT      & \textbf{81.855} & \textbf{6.089} \\
& LoRA         & 109.107 & 7.395 \\
& Less-is-More         & 118.016 & 7.886 \\
& Ours         & \underline{82.352} & \underline{6.144} \\

\bottomrule
\end{tabular}
}
\vspace{-3mm}
\end{wraptable}
To evaluate the model’s personalized adaptation capability and cross-temporal generalization ability on temporally evolving individual data, we use the first 60\% of each subject’s PPG data for training, 20\% of the data for validation, and the subsequent 20\% of the data for testing, thereby simulating a real-world scenario in which historical user data are used for model personalization and subsequent physiological states are predicted. As shown in Table \ref{tab:personalized_adaptation_result}, our method achieves the best performance on PaPaGei-P, PaPaGei-S, and AnyPPG, and the second-best performance on Pulse-PPG. In addition, we calculate the MSE and MAE for each subject. As shown in Figures \ref{fig:patient_mse} and \ref{fig:patient_mae}, our method achieves the best performance for the majority of subjects. These results demonstrate that our method can learn stable individual-specific representations from limited historical data and that these personalized features can generalize to future data. Meanwhile, the results also demonstrate the robustness of our method to within-subject temporal distribution shifts, representing an important step toward personalized modeling of PPG FMs.

\label{app:personalized_adaptation}
\begin{figure}[t]
    \centering

    \begin{subfigure}[t]{0.48\linewidth}
        \centering
        \includegraphics[width=\linewidth]
        {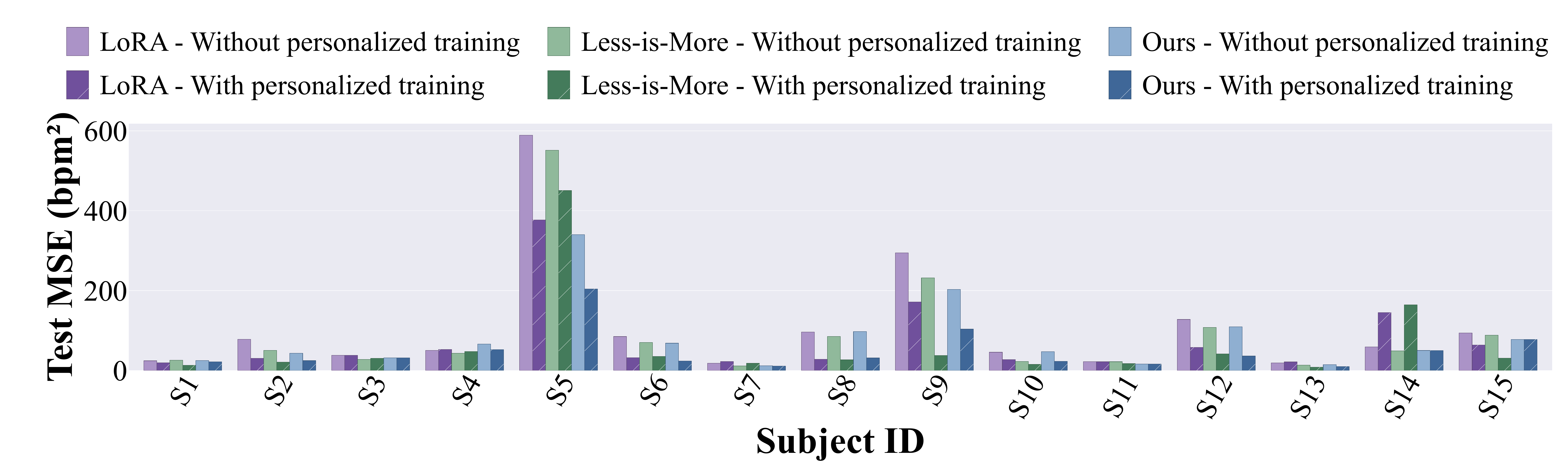}
        \caption{Performance of PaPaGei-P with and without personalized training.}
        \label{fig:personalized_papagei-p}
    \end{subfigure}
    \hfill
    \begin{subfigure}[t]{0.48\linewidth}
        \centering
        \includegraphics[width=\linewidth]
        {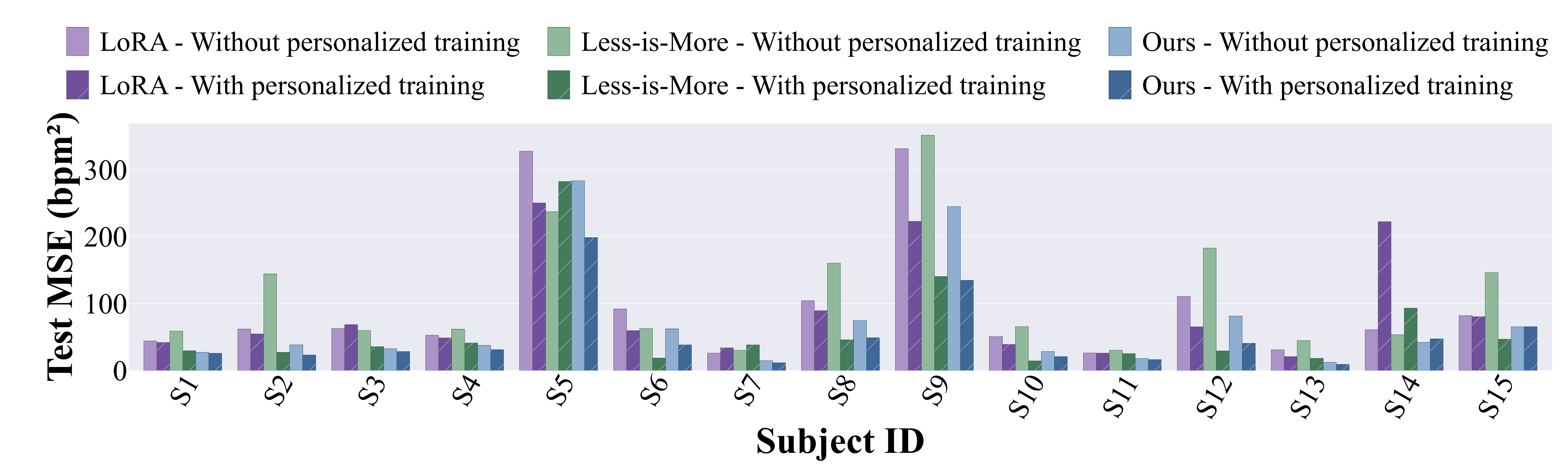}
        \caption{Performance of PulsePPG with and without personalized training.}
        \label{fig:personalized_pulseppg}
    \end{subfigure}

    \vspace{6pt}

    \makebox[\linewidth][c]{%
        \begin{subfigure}[t]{0.48\linewidth}
            \centering
            \includegraphics[width=\linewidth]
            {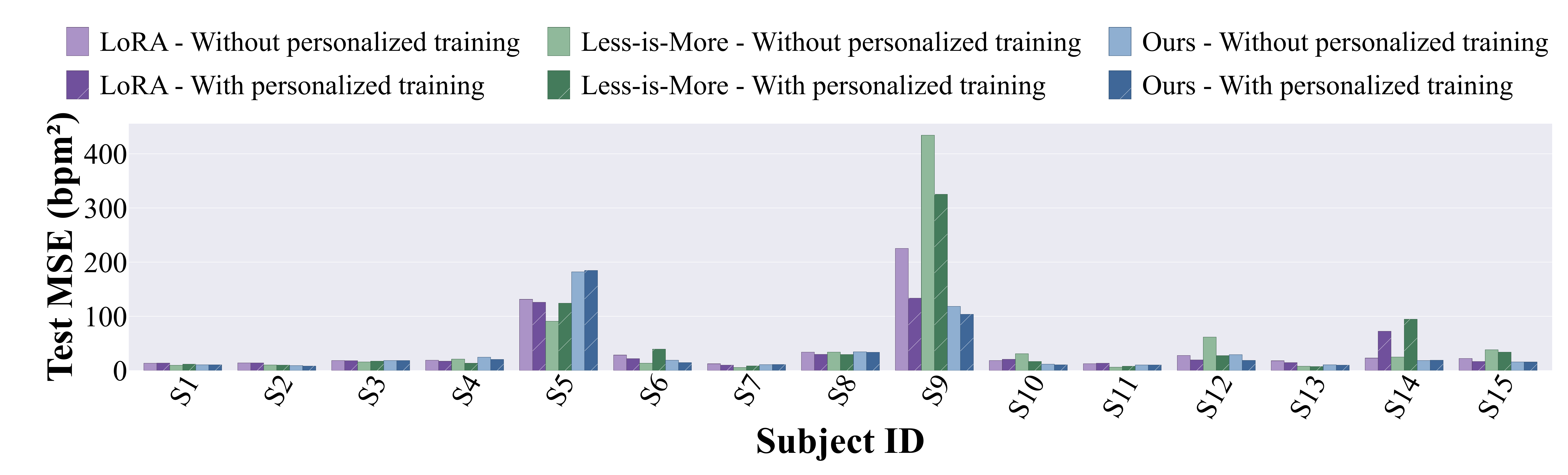}
            \caption{Performance of AnyPPG with and without personalized training.}
            \label{fig:personalized_anyppg}
        \end{subfigure}%
    }

    \caption{Comparison of test MSE with and without personalized training across different models.}
    \label{fig:personalized_training_mse}
    \vspace{-1mm}
\end{figure}

\begin{table*}[t]
\centering
\caption{Impact of personalized training on different PPG FMs.}
\label{tab:personalized_training_improvement}
\renewcommand{\arraystretch}{1.15}
\setlength{\tabcolsep}{6pt}
\resizebox{0.7\textwidth}{!}{
\begin{tabular}{llcc|cc|cc}
\toprule
\multirow{2}{*}{\textbf{Model}} &
\multirow{2}{*}{\textbf{Method}} &
\multicolumn{2}{c|}{\textbf{w/o Personalized Training}} &
\multicolumn{2}{c|}{\textbf{w/ Personalized Training}} &
\multicolumn{2}{c}{\textbf{Relative Improvement}} \\

\cmidrule(lr){3-4}
\cmidrule(lr){5-6}
\cmidrule(lr){7-8}

&
& \textbf{MSE}$\downarrow$
& \textbf{MAE}$\downarrow$
& \textbf{MSE}$\downarrow$
& \textbf{MAE}$\downarrow$
& \textbf{MSE Gain}
& \textbf{MAE Gain}
\\
\midrule

\multirow{3}{*}{PaPaGei-P}

& LoRA 
& 109.455 
& 6.546 
& 75.293 
& 5.562 
& 31.21\% 
& 15.03\% 
\\

& Less-is-More
& 93.445
& 6.167
& 65.593
& 4.922
& 29.81\%
& 20.19\%
\\

& Ours
& 79.733 
& 5.750 
& 48.270
& 4.607 
& 39.46\% 
& 19.88\% 
\\

\midrule

\multirow{3}{*}{PaPaGei-S}

& LoRA 
& 110.504 
& 7.594 
& 73.173 
& 5.742 
& 33.78\%
& 24.39\%
\\

& Less-is-More
& 146.365
& 8.604
& 127.635
& 8.204
& 12.80\%
& 4.65\%
\\

& Ours
& 96.643 
& 6.528
& 56.648 
& 5.014
& 41.38\%
& 23.19\%
\\

\midrule

\multirow{3}{*}{AnyPPG}

& LoRA 
& 41.180
& 3.945
& 36.494
& 3.750 
& 11.38\%
& 4.94\%
\\

& Less-is-More
& 53.471
& 3.942
& 50.680
& 4.027
& 5.22\%
& $-2.16\%$
\\

& Ours 
& 35.272
& 3.531 
& 33.177
& 3.467
& 5.94\%
& 1.81\%
\\

\midrule

\multirow{3}{*}{PulsePPG}

& LoRA 
& 96.678 
& 6.776 
& 88.338
& 6.680 
& 8.63\%
& 1.42\%
\\

& Less-is-More
& 111.117
& 8.019
& 60.375
& 5.306
& 45.67\%
& 33.83\%
\\

& Ours 
& 70.216 
& 5.588 
& 49.231 
& 4.768 
& 29.89\%
& 14.67\%
\\

\bottomrule
\end{tabular}
}
\end{table*}

Furthermore, we compare subject-specific personalization with
SPAFT, LoRA and Less-is-More, starting from each method's corresponding
task-adapted checkpoint. Please note that since the UCI-BP, BUT PPG, and Ring-PPG datasets do not contain timestamp information, we did not perform personalized training on these three datasets. For each subject, we chronologically divide the data from the earlier half of the original test samples into training and validation sets at an 8:2 ratio, while reserving the data from the later half for evaluation. All methods start from their corresponding task-adapted
checkpoints. SPAFT and Less-is-More use previously pruned
models: SPAFT updates only the previously selected convolutional
parameter paths and task head, whereas Less-is-More fully
fine-tunes the remaining backbone and task head. LoRA starts
from its unpruned, task-adapted model and updates only its
existing adapters and task head. Models with and without
personalized training are evaluated on the same test set,
comprising 1,303 samples from 15 subjects.

As shown in Figures~\ref{fig:personalized_papagei-s}
and~\ref{fig:personalized_training_mse}, personalized training
benefits a broad range of subjects, although its effectiveness
varies across methods and models. Compared with its counterpart
without personalized training, SPAFT achieves lower MSE for 14, 15, 12, and 13 subjects on PaPaGei-P,
PaPaGei-S, AnyPPG, and PulsePPG, respectively. 

Table~\ref{tab:personalized_training_improvement} summarizes
the overall results with and without personalized training.
All evaluated model--method combinations exhibit reductions
in pooled MSE. SPAFT achieves the lowest post-personalization
MSE and MAE across all four PPG FMs, reducing its MSE by
39.46\%, 41.38\%, 5.94\%, and 29.89\% on PaPaGei-P,
PaPaGei-S, AnyPPG, and PulsePPG, respectively. Although
Less-is-More achieves a larger relative MSE reduction on
PulsePPG (45.67\%), SPAFT attains a lower final MSE.
On AnyPPG, the improvement is more modest, but SPAFT
achieves a final MSE of 33.177, compared with 36.494 for
LoRA and 50.680 for Less-is-More. These results demonstrate that the parameter paths selected by SPAFT retain the capacity for further individual adaptation, and that personalized training can improve model performance at the patient level.

\subsection{Task Transferability}\label{appendix:transfer}
We further investigate whether the specialized subnetworks learned by SPAFT can transfer across related downstream tasks. Specifically, we first specialize each PPG foundation model on either SBP or DBP estimation, transfer the resulting backbone to the other task,
\begin{wraptable}{r}{0.44\textwidth}
\vspace{-1mm}
\centering
\caption{Transferability Study on UCI-BP.}
\label{tab:transferability}
\vspace{-1mm}
\resizebox{0.44\textwidth}{!}{
\begin{tabular}{llcc|cc}
\toprule
\multirow{2}{*}{\textbf{Source$\rightarrow$Target}} &
\multirow{2}{*}{\textbf{PPG FM}} &
\multicolumn{2}{c|}{\textbf{Linear Probe}} &
\multicolumn{2}{c}{\textbf{SPAFT Transfer}} \\
& & \textbf{MSE$\downarrow$} & \textbf{MAE$\downarrow$}
  & \textbf{MSE$\downarrow$} & \textbf{MAE$\downarrow$} \\
\midrule
\multirow{4}{*}{DBP$\rightarrow$SBP}
& AnyPPG    & 379.386 & 15.384 & \textbf{306.571} & \textbf{13.562} \\
& PaPaGei-P & 430.122 & 16.539 & \textbf{348.212} & \textbf{14.595} \\
& PaPaGei-S & 461.959 & 17.296 & \textbf{377.095} & \textbf{15.327} \\
& Pulse-PPG  & 417.640 & 16.264 & \textbf{333.899} & \textbf{14.267} \\
\midrule
\multirow{4}{*}{SBP$\rightarrow$DBP}
& AnyPPG    & 87.652  & 6.880 & \textbf{71.777} & \textbf{6.068} \\
& PaPaGei-P & 98.560  & 7.437 & \textbf{80.074} & \textbf{6.494} \\
& PaPaGei-S & 104.351 & 7.697 & \textbf{84.164} & \textbf{6.706} \\
& Pulse-PPG  & 95.379  & 7.284 & \textbf{78.220} & \textbf{6.412} \\
\bottomrule
\end{tabular}
}
\vspace{-4mm}
\end{wraptable}
and train only a new task head on the target-task training set. Linear Probe, which trains the same target-task head on the original pretrained backbone, is used as the reference. In Table~\ref{tab:transferability}, SPAFT transfer consistently outperforms Linear Probe. Compared with Linear Probe, transferring from DBP to SBP reduces MSE by 18.37\%-20.05\% and MAE by 11.38\%-12.28\%, while transferring from SBP to DBP yields reductions of 17.99\%-19.35\% and 11.80\%-12.88\%, respectively. These results demonstrate the cross-task transferability of the specialized structures learned by SPAFT.
\subsection{Inference Efficiency}\label{appendix:infer}
To evaluate the inference efficiency of SPAFT, we compare the computational and memory footprints of the original PPG FMs with those of the models after SPAFT-based pruning and fine-tuning. As shown in Table~\ref{tab:inference_efficiency}, we report the test memory usage (TMU), FLOPs, and number of parameters across three downstream tasks and four PPG FMs.
\begin{wraptable}{r}{0.55\textwidth}
\vspace{-1mm}
\centering
\caption{Comparison of computational efficiency.}
\label{tab:inference_efficiency}
\vspace{-1mm}
\resizebox{0.55\textwidth}{!}{
\begin{tabular}{llccc|ccc}
\toprule
\multirow{2}{*}{\textbf{Dataset}} &
\multirow{2}{*}{\textbf{PPG FMs}} &
\multicolumn{3}{c|}{\textbf{Original}} &
\multicolumn{3}{c}{\textbf{SPAFT}} \\
\cmidrule(lr){3-5}
\cmidrule(lr){6-8}
&
&
\textbf{TMU (GB)} &
\textbf{FLOPs (G)} &
\textbf{Params (M)} &
\textbf{TMU (GB)} &
\textbf{FLOPs (G)} &
\textbf{Params (M)} \\
\midrule

\multirow{4}{*}{\textbf{Ring-PPG HR}}
& AnyPPG
& 0.0479 & 0.2173 & 4.0448
& \textbf{0.0442} & \textbf{0.1781} & \textbf{3.4372} \\

& PaPaGei-S
& 0.0565 & 0.1502 & 5.7861
& \textbf{0.0533} & \textbf{0.0803} & \textbf{5.6323} \\

& PaPaGei-P
& 0.0534 & 0.1502 & 4.9935
& \textbf{0.0480} & \textbf{0.1135} & \textbf{3.3352} \\

& Pulse-PPG
& 0.2359 & 5.6784 & 28.4984
& \textbf{0.2240} & \textbf{5.5162} & \textbf{26.4704} \\

\midrule

\multirow{4}{*}{\textbf{PPG-DaLiA HR}}
& AnyPPG
& 0.0479 & 0.2173 & 4.0448
& \textbf{0.0440} & \textbf{0.1583} & \textbf{3.1867} \\

& PaPaGei-S
& 0.0566 & 0.1502 & 5.7861
& \textbf{0.0556} & \textbf{0.1198} & \textbf{5.6224} \\

& PaPaGei-P
& 0.0529 & 0.1502 & 4.9935
& \textbf{0.0471} & \textbf{0.0837} & \textbf{4.0084} \\

& Pulse-PPG
& 0.2359 & 5.6784 & 28.4984
& \textbf{0.2116} & \textbf{4.6048} & \textbf{27.9910} \\

\midrule

\multirow{4}{*}{\textbf{UCI-BP SBP}}
& AnyPPG
& 0.0479 & 0.2173 & 4.0448
& \textbf{0.0472} & \textbf{0.2080} & \textbf{3.8808} \\

& PaPaGei-S
& 0.0560 & 0.1502 & 5.7861
& \textbf{0.0490} & \textbf{0.1245} & \textbf{3.9142} \\

& PaPaGei-P
& 0.0529 & 0.1502 & 4.9935
& \textbf{0.0526} & \textbf{0.1471} & \textbf{4.5953} \\

& Pulse-PPG
& 0.2359 & 5.6784 & 28.4984
& \textbf{0.2242} & \textbf{5.5162} & \textbf{26.4704} \\

\bottomrule
\end{tabular}
}
\vspace{-3.5mm}
\end{wraptable}
 Overall,  SPAFT consistently reduces all three efficiency metrics across the evaluated backbone--task combinations. Averaged over the 12 settings, SPAFT reduces test memory usage, FLOPs, and parameter count by 6.60\%, 19.06\%, and 12.92\%, respectively, with maximum reductions of 12.50\%, 46.54\%, and 33.21\%. The reduction magnitude varies across different backbone--task combinations, which is expected because SPAFT adaptively determines the compact structure according to task-specific redundancy rather than enforcing a fixed pruning configuration. Together with the performance improvements reported in Tables~\ref{tab:overall_performance_1} and~\ref{tab:overall_performance_2}, these results show that SPAFT enables more compact and computationally efficient specialization of PPG foundation models without trading predictive performance for model compression.
\subsection{Adaptivity Analysis}\label{appendix:adaptive}

\begin{wrapfigure}{r}{0.50\textwidth}
\vspace{-1mm}
\centering
\includegraphics[width=0.49\textwidth]{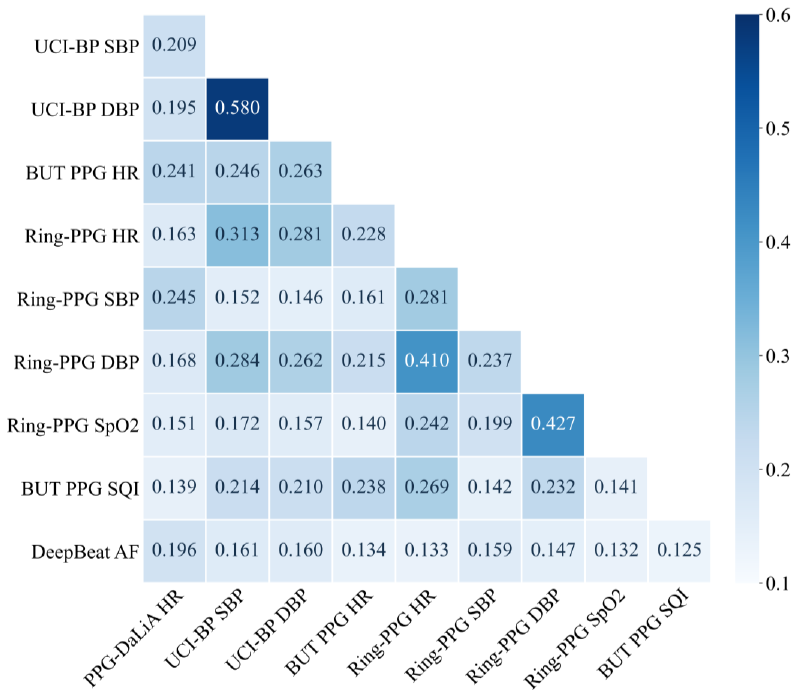}
\vspace{-3mm}
\caption{Visualization of the Overlap Rate.}
\label{fig:case}
\vspace{-4mm}
\end{wrapfigure}
To evaluate the impact of our task-relevant parameter selection strategy,
we use PAPAGEI-P to analyze the layer-wise distribution of fine-tuning
filters and the overlap between task-specific fine-tuning filter sets.

\begin{figure*}[t]
\centering
\includegraphics[width=\textwidth]{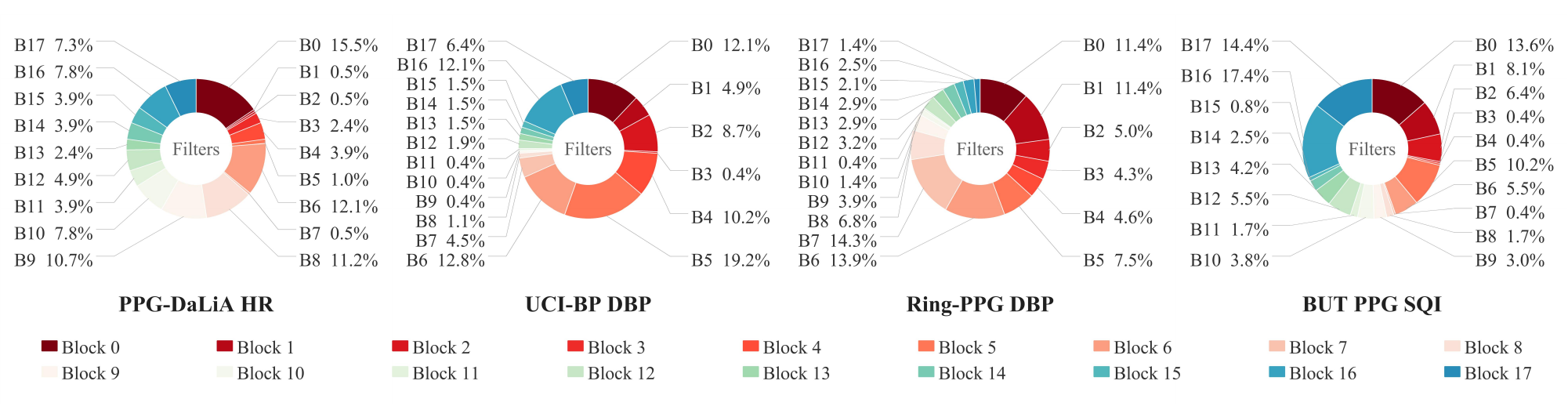}\par
\vspace{-6pt}
\caption{Distribution of the Top 10\% Filters in the  PAPAGEI-P.}
\label{fig:Filters Distribution}
\vspace{-1mm}
\end{figure*}

\textbf{Layer-Wise Distribution of Fine-Tuning Filter Proportions:}
The distribution of fine-tuning filters across different datasets reflects
task relevance. As shown in Figure~\ref{fig:Filters Distribution}, the
top 10\% filters of UCI-BP DBP are mainly concentrated in Blocks 5 and 6,
indicating that this task relies more heavily on low-level feature extraction.
In contrast, the fine-tuning filters of BUT PPG SQI are primarily
concentrated in Blocks 16 and 17, suggesting that this task relies more
heavily on high-level features. These results demonstrate the importance
of task-aware parameter selection.

\textbf{Overlap Between Task-Specific Filter Sets:}
As shown in Figure~\ref{fig:case}, different tasks exhibit relatively
low filter overlap. For example, the overlap between UCI-BP SBP and
UCI-BP DBP is 0.58. In contrast, the overlap between DeepBeat AF and
BUT PPG SQI is 0.125. Overall, the relatively low filter overlap further
validates the effectiveness of our adaptive filter selection strategy.

\begin{figure*}[t]
\centering
\includegraphics[width=0.96\textwidth]{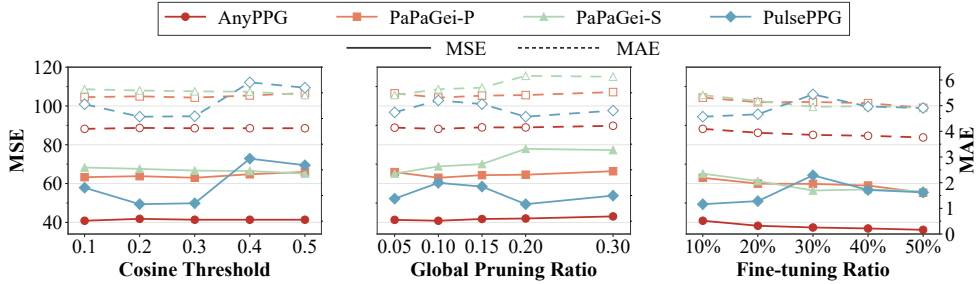}\par
\vspace{-4pt}
\caption{Hyperparameter sensitivity of SPAFT on the UCI-BP DBP task across different PPG FMs, including the cosine threshold, global pruning ratio, and fine-tuning ratio.}
\label{fig:hyperparameters_experiment_uci_bp_dbp}
\end{figure*}

\begin{figure*}[t]
\centering
\includegraphics[width=0.96\textwidth]{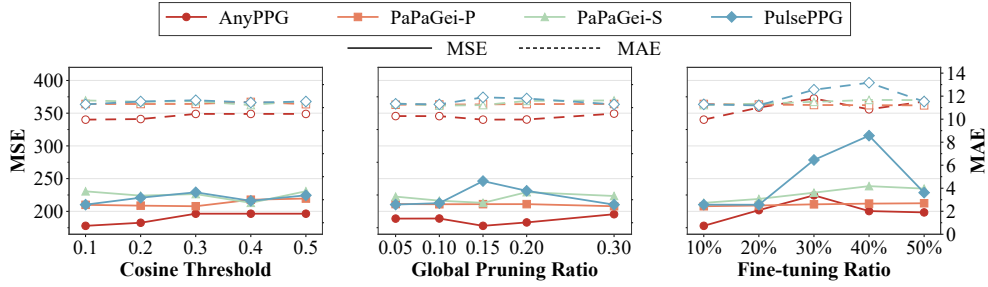}\par
\vspace{-4pt}
\caption{Hyperparameter sensitivity of SPAFT on the BUT PPG HR task across different PPG FMs, including the cosine threshold, global pruning ratio, and fine-tuning ratio.}
\label{fig:hyperparameters_experiment_but_ppg_hr}
\end{figure*}
\subsection{Parameter Analysis}\label{appendix:parameter}
We investigate the sensitivity of SPAFT to three key hyperparameters, including the cosine threshold, global pruning ratio, and fine-tuning ratio during fine-tuning.

\textbf{Varying cosine threshold.}
As shown in the left panels of Figures~\ref{fig:hyperparameters_experiment_uci_bp_dbp} and~\ref{fig:hyperparameters_experiment_but_ppg_hr}, the optimal threshold varies across FMs and downstream tasks. For example, on UCI-BP DBP, AnyPPG achieves its lowest MSE at a threshold of 0.1, whereas PaPaGei-S performs best at 0.5. Since the cosine threshold controls edge formation in the intra-layer structural graph, a small value may connect weakly similar filters and overestimate structural redundancy, whereas an overly large value may overlook potentially substitutable filters. These results suggest that different pretrained PPG FMs exhibit distinct structural similarity patterns and therefore require different thresholds for redundancy estimation.

\textbf{Varying global pruning ratio.}
As shown in the middle panels of Figures~\ref{fig:hyperparameters_experiment_uci_bp_dbp} and~\ref{fig:hyperparameters_experiment_but_ppg_hr}, On UCI-BP DBP, the MSE-optimal global pruning ratios range from 0.05 to 0.20 across the four foundation models, whereas on BUT PPG HR they range from 0.15 to 0.30. This variation indicates that the amount of removable redundancy differs across backbone--task combinations. A small pruning ratio may retain excessive redundant structures, while overly aggressive pruning may remove filters that remain important for downstream prediction. Therefore, an appropriate compression strength is required to balance redundancy removal and preservation of task-relevant capacity.

\textbf{Varying fine-tuning ratio.}
As shown in the right panels of Figures~\ref{fig:hyperparameters_experiment_uci_bp_dbp} and~\ref{fig:hyperparameters_experiment_but_ppg_hr}, on UCI-BP DBP, the performance of AnyPPG, PaPaGei-P, and PaPaGei-S generally improves as the fine-tuning ratio increases. However, Pulse-PPG favors a smaller fine-tuning ratio, indicating that the required adaptation capacity also varies across backbone models. In contrast, on BUT PPG HR, most models achieve their best or near-best performance with only 10\%-20\% of the selected filters. These contrasting trends suggest that too few trainable paths may limit task fine-tuning, whereas unnecessarily broad parameter updates may weaken the benefits of selective adaptation.
\newpage
\subsection{Effectiveness of Sparse Parameter Adaptation.}\label{appendix:ada}
\begin{wraptable}{r}{0.40\textwidth}
\vspace{-3mm}
\centering
\caption{Performance of adapting only the top 1\% task-sensitive filters.}
\label{tab:top1_adaptation_ucibp_dbp}
\resizebox{0.40\textwidth}{!}{
\begin{tabular}{lccc}
\toprule
\textbf{FM} & \textbf{Method} & \textbf{MSE} $\downarrow$ & \textbf{MAE} $\downarrow$ \\
\midrule
\multirow{3}{*}{AnyPPG}
& LoRA & 78.374 & 6.224 \\
& Taylor-FO & \textbf{53.811} & \textbf{4.811} \\
& Top 1\% Adaptation & 59.508 & 5.175 \\
\midrule
\multirow{3}{*}{PaPaGei-P}
& LoRA & 88.482 & 6.768 \\
& Taylor-FO & 89.920 & 6.882 \\
& Top 1\% Adaptation & \textbf{74.135} & \textbf{5.942} \\
\midrule
\multirow{3}{*}{PaPaGei-S}
& LoRA & 95.854 & 7.112 \\
& Taylor-FO & 84.882 & 6.631 \\
& Top 1\% Adaptation & \textbf{77.486} & \textbf{6.232} \\
\midrule
\multirow{3}{*}{Pulse-PPG}
& LoRA & 82.934 & 6.449 \\
& Taylor-FO & 65.958 & 5.518 \\
& Top 1\% Adaptation & \textbf{62.617} & \textbf{5.330} \\
\bottomrule
\end{tabular}
}
\vspace{-2mm}
\end{wraptable}
To further investigate whether downstream specialization can be achieved by updating only a small subset of task-relevant parameters, we adapt only the top 1\% task-sensitive filters on the UCI-BP DBP task and compare them with the representative baseline methods LoRA and Taylor-FO. As shown in Table~\ref{tab:top1_adaptation_ucibp_dbp}, the proposed sparse adaptation strategy achieves competitive performance with extremely limited parameter updates.  Compared with LoRA, top 1\% adaptation consistently improves performance across all pretrained models, demonstrating the advantage of explicitly identifying task-relevant parameters rather than relying on generic low-rank updates. Compared with Taylor-FO, it achieves better results on three out of four foundation models while updating only a small fraction of parameters. These results further support our motivation that PPG foundation models contain task-specific substructures, and selectively adapting important parameter paths can provide effective downstream specialization.

\begin{sidewaystable*}[p]
\centering
\caption{The performance of the methods on regression tasks.
We mark the winner in \textbf{bold}, and the runner-up is indicated by an \underline{underline}.}
\label{tab:Complete_overall_performance_1}

\scriptsize
\setlength{\tabcolsep}{3pt}
\renewcommand{\arraystretch}{0.88}
\resizebox{\textwidth}{!}{
\begin{tabular}{
c|c|c|
*{9}{cc|}
cc
}
\toprule
{} & {} & {}
& \multicolumn{2}{c|}{\textbf{Linear Probe}}
& \multicolumn{2}{c|}{\textbf{Full-FT}}
& \multicolumn{2}{c|}{\textbf{LoRA}}
& \multicolumn{2}{c|}{\textbf{QLoRA}}
& \multicolumn{2}{c|}{\textbf{PFEC}}
& \multicolumn{2}{c|}{\textbf{Taylor-FO}}
& \multicolumn{2}{c|}{\textbf{FPGM}}
& \multicolumn{2}{c|}{\textbf{SRR-GR}}
& \multicolumn{2}{c|}{\textbf{Less-is-More}}
& \multicolumn{2}{c}{\textbf{Ours}}
\\

\textbf{Data} & \textbf{Tasks} & \textbf{FM}
& MSE $\downarrow$ & MAE $\downarrow$
& MSE $\downarrow$ & MAE $\downarrow$
& MSE $\downarrow$ & MAE $\downarrow$
& MSE $\downarrow$ & MAE $\downarrow$
& MSE $\downarrow$ & MAE $\downarrow$
& MSE $\downarrow$ & MAE $\downarrow$
& MSE $\downarrow$ & MAE $\downarrow$
& MSE $\downarrow$ & MAE $\downarrow$
& MSE $\downarrow$ & MAE $\downarrow$
& MSE $\downarrow$ & MAE $\downarrow$
\\
\midrule

\multirow{8}{*}{UCI-BP} & \multirow{4}{*}{SBP} & AnyPPG
& $379.531_{\pm 0.127}$ & $15.388_{\pm 0.003}$
& $221.626_{\pm 1.736}$ & $10.804_{\pm 0.047}$
& $327.334_{\pm 0.798}$ & $13.870_{\pm 0.020}$
& $392.230_{\pm 31.812}$ & $15.488_{\pm 0.702}$
& $460.250_{\pm 92.183}$ & $16.821_{\pm 1.944}$
& $\underline{191.174_{\pm 1.302}}$ & $\underline{9.871_{\pm 0.039}}$
& $223.532_{\pm 1.259}$ & $10.885_{\pm 0.037}$
& $338.435_{\pm 25.794}$ & $14.133_{\pm 0.605}$
& $379.268_{\pm 28.346}$ & $15.086_{\pm 0.664}$
& $\cellcolor{bestblue}\mathbf{143.083_{\pm 1.651}}$ & $\cellcolor{bestblue}\mathbf{8.345_{\pm 0.057}}$ \\

&& PaPaGei-P
& $430.231_{\pm 0.117}$ & $16.542_{\pm 0.005}$
& $371.494_{\pm 2.088}$ & $14.974_{\pm 0.064}$
& $379.325_{\pm 1.389}$ & $15.183_{\pm 0.038}$
& $471.292_{\pm 29.559}$ & $17.481_{\pm 0.769}$
& $482.716_{\pm 71.832}$ & $17.735_{\pm 1.762}$
& $371.448_{\pm 11.121}$ & $14.924_{\pm 0.336}$
& $\underline{365.637_{\pm 5.323}}$ & $\underline{14.757_{\pm 0.119}}$
& $482.458_{\pm 59.984}$ & $17.775_{\pm 1.506}$
& $434.871_{\pm 34.974}$ & $16.591_{\pm 0.868}$
& $\cellcolor{bestblue}\mathbf{245.788_{\pm 2.398}}$ & $\cellcolor{bestblue}\mathbf{11.438_{\pm 0.080}}$ \\

&& PaPaGei-S
& $461.856_{\pm 0.141}$ & $17.295_{\pm 0.003}$
& $383.028_{\pm 11.254}$ & $15.299_{\pm 0.328}$
& $422.433_{\pm 1.216}$ & $16.301_{\pm 0.034}$
& $438.100_{\pm 36.182}$ & $16.600_{\pm 0.896}$
& $492.882_{\pm 28.091}$ & $17.987_{\pm 0.684}$
& $393.232_{\pm 1.726}$ & $15.628_{\pm 0.076}$
& $\underline{380.783_{\pm 1.033}}$ & $\underline{15.240_{\pm 0.055}}$
& $445.977_{\pm 13.978}$ & $16.893_{\pm 0.262}$
& $387.265_{\pm 47.610}$ & $15.478_{\pm 1.323}$
& $\cellcolor{bestblue}\mathbf{259.401_{\pm 31.810}}$ & $\cellcolor{bestblue}\mathbf{11.858_{\pm 0.937}}$ \\

&& Pulse-PPG
& $417.647_{\pm 0.058}$ & $16.263_{\pm 0.001}$
& $256.021_{\pm 1.548}$ & $11.825_{\pm 0.038}$
& $356.951_{\pm 0.886}$ & $14.652_{\pm 0.034}$
& $354.624_{\pm 56.666}$ & $14.660_{\pm 1.457}$
& $335.480_{\pm 11.302}$ & $14.175_{\pm 0.256}$
& $257.260_{\pm 0.977}$ & $11.845_{\pm 0.026}$
& $\underline{254.844_{\pm 0.405}}$ & $\underline{11.788_{\pm 0.016}}$
& $312.325_{\pm 33.778}$ & $13.553_{\pm 0.898}$
& $277.007_{\pm 48.312}$ & $12.581_{\pm 1.314}$
& $\cellcolor{bestblue}\mathbf{183.581_{\pm 27.051}}$ & $\cellcolor{bestblue}\mathbf{9.661_{\pm 0.861}}$ \\

\cmidrule{2-23}

&\multirow{4}{*}{DBP}
& AnyPPG
& $87.673_{\pm 0.019}$ & $6.881_{\pm 0.001}$
& $59.591_{\pm 0.091}$ & $5.156_{\pm 0.007}$
& $78.304_{\pm 0.072}$ & $6.224_{\pm 0.012}$
& $92.546_{\pm 13.155}$ & $7.243_{\pm 0.857}$
& $82.621_{\pm 3.812}$ & $6.564_{\pm 0.251}$
& $\underline{53.857_{\pm 0.414}}$ & $\underline{4.816_{\pm 0.016}}$
& $78.587_{\pm 29.176}$ & $6.081_{\pm 1.416}$
& $76.814_{\pm 7.203}$ & $6.256_{\pm 0.348}$
& $80.273_{\pm 7.243}$ & $6.475_{\pm 0.365}$
& $\cellcolor{bestblue}\mathbf{41.276_{\pm 0.674}}$ & $\cellcolor{bestblue}\mathbf{4.111_{\pm 0.030}}$ \\

&& PaPaGei-P
& $98.548_{\pm 0.011}$ & $7.436_{\pm 0.001}$
& $87.802_{\pm 1.013}$ & $6.735_{\pm 0.034}$
& $88.378_{\pm 0.173}$ & $6.752_{\pm 0.016}$
& $92.338_{\pm 6.301}$ & $7.192_{\pm 0.343}$
& $88.084_{\pm 5.032}$ & $6.814_{\pm 0.310}$
& $90.350_{\pm 0.930}$ & $6.846_{\pm 0.055}$
& $\underline{87.061_{\pm 0.710}}$ & $\underline{6.659_{\pm 0.112}}$
& $99.238_{\pm 5.593}$ & $7.458_{\pm 0.151}$
& $89.353_{\pm 7.836}$ & $6.992_{\pm 0.459}$
& $\cellcolor{bestblue}\mathbf{62.908_{\pm 0.796}}$ & $\cellcolor{bestblue}\mathbf{5.317_{\pm 0.036}}$ \\

&& PaPaGei-S
& $104.376_{\pm 0.024}$ & $7.697_{\pm 0.001}$
& $\underline{82.521_{\pm 0.509}}$ & $\underline{6.446_{\pm 0.054}}$
& $95.727_{\pm 0.110}$ & $7.113_{\pm 0.018}$
& $99.174_{\pm 11.130}$ & $7.196_{\pm 0.498}$
& $88.765_{\pm 6.796}$ & $6.918_{\pm 0.351}$
& $85.060_{\pm 0.251}$ & $6.608_{\pm 0.026}$
& $84.573_{\pm 1.127}$ & $6.579_{\pm 0.085}$
& $95.835_{\pm 6.522}$ & $7.156_{\pm 0.394}$
& $90.817_{\pm 12.584}$ & $6.973_{\pm 0.687}$
& $\cellcolor{bestblue}\mathbf{64.756_{\pm 0.411}}$ & $\cellcolor{bestblue}\mathbf{5.420_{\pm 0.032}}$ \\

&& Pulse-PPG
& $95.357_{\pm 0.025}$ & $7.283_{\pm 0.001}$
& $66.772_{\pm 0.356}$ & $5.575_{\pm 0.070}$
& $82.859_{\pm 0.119}$ & $6.460_{\pm 0.025}$
& $82.715_{\pm 8.425}$ & $6.559_{\pm 0.496}$
& $96.364_{\pm 15.146}$ & $7.074_{\pm 0.563}$
& $\underline{66.458_{\pm 0.773}}$ & $\underline{5.550_{\pm 0.052}}$
& $67.699_{\pm 0.457}$ & $5.617_{\pm 0.040}$
& $78.050_{\pm 5.533}$ & $6.202_{\pm 0.345}$
& $91.655_{\pm 12.926}$ & $6.907_{\pm 0.533}$
& $\cellcolor{bestblue}\mathbf{49.865_{\pm 0.777}}$ & $\cellcolor{bestblue}\mathbf{4.609_{\pm 0.052}}$ \\

\midrule

\multirow{4}{*}{PPG-DaLiA}&\multirow{4}{*}{HR}
& AnyPPG
& $92.093_{\pm 11.260}$ & $6.505_{\pm 0.461}$
& $82.302_{\pm 5.740}$ & $5.597_{\pm 0.248}$
& $87.603_{\pm 9.595}$ & $6.304_{\pm 0.461}$
& $242.786_{\pm 136.950}$ & $11.836_{\pm 4.099}$
& $151.705_{\pm 80.864}$ & $7.873_{\pm 3.175}$
& $\underline{75.095_{\pm 2.726}}$ & $\underline{5.385_{\pm 0.308}}$
& $164.685_{\pm 120.136}$ & $7.387_{\pm 2.637}$
& $142.293_{\pm 42.226}$ & $7.181_{\pm 1.269}$
& $145.387_{\pm 68.814}$ & $7.923_{\pm 3.345}$
& $\cellcolor{bestblue}\mathbf{71.870_{\pm 2.496}}$ & $\cellcolor{bestblue}\mathbf{5.066_{\pm 0.183}}$ \\

&& PaPaGei-P
& $196.916_{\pm 17.537}$ & $10.282_{\pm 0.584}$
& $222.989_{\pm 47.020}$ & $11.029_{\pm 1.466}$
& $190.257_{\pm 19.109}$ & $10.043_{\pm 0.643}$
& $263.867_{\pm 20.264}$ & $12.824_{\pm 0.643}$
& $212.701_{\pm 29.589}$ & $11.105_{\pm 1.108}$
& $212.615_{\pm 7.322}$ & $11.002_{\pm 0.326}$
& $259.243_{\pm 78.467}$ & $12.630_{\pm 2.657}$
& $190.598_{\pm 9.246}$ & $10.054_{\pm 0.479}$
& $\underline{174.190_{\pm 12.932}}$ & $\underline{9.386_{\pm 0.483}}$
& $\cellcolor{bestblue}\mathbf{108.284_{\pm 0.922}}$ & $\cellcolor{bestblue}\mathbf{6.818_{\pm 0.280}}$ \\

&& PaPaGei-S
& $228.658_{\pm 14.162}$ & $11.563_{\pm 0.458}$
& $258.853_{\pm 85.998}$ & $12.017_{\pm 2.480}$
& $217.661_{\pm 11.192}$ & $11.195_{\pm 0.418}$
& $277.062_{\pm 16.512}$ & $13.108_{\pm 0.564}$
& $229.755_{\pm 36.121}$ & $11.384_{\pm 1.323}$
& $\underline{174.363_{\pm 6.933}}$ & $\underline{9.736_{\pm 0.407}}$
& $311.547_{\pm 227.648}$ & $13.493_{\pm 6.197}$
& $215.235_{\pm 13.761}$ & $11.246_{\pm 0.344}$
& $285.047_{\pm 90.690}$ & $13.502_{\pm 2.934}$
& $\cellcolor{bestblue}\mathbf{130.235_{\pm 3.036}}$ & $\cellcolor{bestblue}\mathbf{7.977_{\pm 0.286}}$ \\

&& Pulse-PPG
& $149.143_{\pm 11.429}$ & $8.996_{\pm 0.392}$
& $144.507_{\pm 21.643}$ & $8.707_{\pm 0.722}$
& $145.238_{\pm 9.353}$ & $8.869_{\pm 0.314}$
& $199.532_{\pm 16.657}$ & $10.704_{\pm 0.421}$
& $221.702_{\pm 39.349}$ & $11.086_{\pm 1.103}$
& $130.593_{\pm 4.935}$ & $8.291_{\pm 0.212}$
& $\underline{121.318_{\pm 13.596}}$ & $\underline{7.922_{\pm 0.569}}$
& $242.504_{\pm 40.288}$ & $11.332_{\pm 1.808}$
& $253.971_{\pm 60.775}$ & $12.197_{\pm 1.449}$
& $\cellcolor{bestblue}\mathbf{106.274_{\pm 6.373}}$ & $\cellcolor{bestblue}\mathbf{7.191_{\pm 0.274}}$ \\

\midrule

\multirow{16}{*}{Ring-PPG}&\multirow{4}{*}{HR} 
& AnyPPG
& $74.642_{\pm 25.915}$ & $\underline{6.167_{\pm 0.660}}$
& $80.774_{\pm 30.232}$ & $7.376_{\pm 1.903}$
& $77.026_{\pm 20.236}$ & $6.446_{\pm 0.872}$
& $62.443_{\pm 3.770}$ & $6.479_{\pm 0.254}$
& $75.636_{\pm 15.809}$ & $7.004_{\pm 0.874}$
& $76.961_{\pm 37.754}$ & $6.427_{\pm 1.206}$
& $92.972_{\pm 26.540}$ & $7.778_{\pm 1.113}$
& $64.864_{\pm 15.587}$ & $6.469_{\pm 0.932}$
& $\underline{59.285_{\pm 6.331}}$ & $6.371_{\pm 0.394}$
& $\cellcolor{bestblue}\mathbf{58.447_{\pm 27.911}}$ & $\cellcolor{bestblue}\mathbf{5.717_{\pm 1.194}}$ \\

&& PaPaGei-P
& $78.552_{\pm 43.761}$ & $6.645_{\pm 1.585}$
& $91.423_{\pm 49.723}$ & $7.289_{\pm 2.019}$
& $80.245_{\pm 48.526}$ & $6.703_{\pm 1.898}$
& $74.180_{\pm 32.358}$ & $7.021_{\pm 1.606}$
& $\cellcolor{bestblue}\mathbf{58.648_{\pm 7.707}}$ & $\cellcolor{bestblue}\mathbf{6.147_{\pm 0.771}}$
& $76.612_{\pm 42.938}$ & $6.611_{\pm 1.602}$
& $154.472_{\pm 70.143}$ & $10.378_{\pm 2.727}$
& $60.984_{\pm 12.892}$ & $6.424_{\pm 0.824}$
& $\underline{59.671_{\pm 13.663}}$ & $\underline{6.283_{\pm 0.904}}$
& $70.638_{\pm 44.415}$ & $6.296_{\pm 1.713}$ \\

&& PaPaGei-S
& $\underline{92.354_{\pm 24.679}}$ & $7.999_{\pm 1.215}$
& $129.305_{\pm 69.725}$ & $9.622_{\pm 3.049}$
& $153.571_{\pm 54.652}$ & $10.458_{\pm 2.651}$
& $94.347_{\pm 19.997}$ & $8.019_{\pm 0.747}$
& $99.860_{\pm 10.662}$ & $8.328_{\pm 0.375}$
& $209.814_{\pm 106.681}$ & $12.192_{\pm 3.781}$
& $109.040_{\pm 44.631}$ & $8.859_{\pm 2.621}$
& $102.507_{\pm 10.060}$ & $8.464_{\pm 0.284}$
& $94.254_{\pm 24.013}$ & $\underline{7.893_{\pm 1.215}}$
& $\cellcolor{bestblue}\mathbf{77.574_{\pm 26.008}}$ & $\cellcolor{bestblue}\mathbf{7.278_{\pm 1.275}}$ \\

&& Pulse-PPG
& $76.184_{\pm 46.625}$ & $6.715_{\pm 1.974}$
& $\cellcolor{bestblue}\mathbf{69.563_{\pm 33.674}}$ & $\cellcolor{bestblue}\mathbf{6.531_{\pm 1.676}}$
& $81.410_{\pm 36.366}$ & $7.009_{\pm 1.825}$
& $80.308_{\pm 26.854}$ & $7.259_{\pm 1.171}$
& $93.583_{\pm 40.504}$ & $7.684_{\pm 1.759}$
& $77.627_{\pm 50.300}$ & $\underline{6.544_{\pm 1.782}}$
& $164.395_{\pm 98.554}$ & $10.448_{\pm 3.768}$
& $97.916_{\pm 52.945}$ & $7.799_{\pm 2.078}$
& $86.114_{\pm 49.536}$ & $7.228_{\pm 2.212}$
& $\underline{75.062_{\pm 42.857}}$ & $6.697_{\pm 2.061}$ \\

\cmidrule{2-23}

&\multirow{4}{*}{SBP}
& AnyPPG
& $335.459_{\pm 58.707}$ & $14.787_{\pm 1.533}$
& $555.882_{\pm 188.921}$ & $19.105_{\pm 3.516}$
& $323.997_{\pm 10.794}$ & $14.672_{\pm 0.313}$
& $214.162_{\pm 62.672}$ & $12.225_{\pm 1.534}$
& $234.743_{\pm 45.512}$ & $12.668_{\pm 0.945}$
& $321.300_{\pm 119.961}$ & $13.957_{\pm 3.361}$
& $333.044_{\pm 50.720}$ & $14.664_{\pm 0.569}$
& $224.976_{\pm 63.280}$ & $12.301_{\pm 1.563}$
& $\underline{199.295_{\pm 19.970}}$ & $\underline{11.839_{\pm 0.288}}$
& $\cellcolor{bestblue}\mathbf{155.748_{\pm 24.544}}$ & $\cellcolor{bestblue}\mathbf{10.751_{\pm 0.956}}$ \\

&& PaPaGei-P
& $389.351_{\pm 76.498}$ & $16.079_{\pm 1.725}$
& $391.395_{\pm 72.262}$ & $16.431_{\pm 1.746}$
& $358.480_{\pm 56.990}$ & $15.566_{\pm 1.500}$
& $212.676_{\pm 51.067}$ & $12.178_{\pm 0.933}$
& $219.713_{\pm 71.550}$ & $12.468_{\pm 1.738}$
& $388.049_{\pm 55.387}$ & $16.048_{\pm 1.249}$
& $239.318_{\pm 64.047}$ & $13.021_{\pm 1.716}$
& $223.084_{\pm 77.738}$ & $12.487_{\pm 1.806}$
& $\underline{188.714_{\pm 10.058}}$ & $\underline{11.984_{\pm 0.609}}$
& $\cellcolor{bestblue}\mathbf{162.895_{\pm 9.696}}$ & $\cellcolor{bestblue}\mathbf{10.843_{\pm 0.613}}$ \\

&& PaPaGei-S
& $391.653_{\pm 176.428}$ & $15.372_{\pm 3.132}$
& $427.919_{\pm 148.141}$ & $16.231_{\pm 2.757}$
& $339.616_{\pm 98.663}$ & $14.218_{\pm 1.724}$
& $\underline{189.743_{\pm 10.320}}$ & $11.814_{\pm 0.348}$
& $204.748_{\pm 28.279}$ & $12.191_{\pm 0.752}$
& $580.246_{\pm 64.269}$ & $19.623_{\pm 1.393}$
& $230.095_{\pm 68.058}$ & $12.836_{\pm 1.744}$
& $197.943_{\pm 36.535}$ & $11.873_{\pm 0.792}$
& $191.154_{\pm 15.809}$ & $\underline{11.752_{\pm 0.327}}$
& $\cellcolor{bestblue}\mathbf{152.555_{\pm 22.081}}$ & $\cellcolor{bestblue}\mathbf{10.454_{\pm 0.894}}$ \\

&& Pulse-PPG
& $321.497_{\pm 109.443}$ & $14.497_{\pm 2.524}$
& $382.435_{\pm 71.882}$ & $16.023_{\pm 1.940}$
& $291.316_{\pm 80.279}$ & $13.543_{\pm 1.243}$
& $237.512_{\pm 30.698}$ & $\underline{12.598_{\pm 0.720}}$
& $271.118_{\pm 34.147}$ & $13.579_{\pm 0.388}$
& $418.904_{\pm 247.312}$ & $16.017_{\pm 5.019}$
& $300.412_{\pm 95.612}$ & $14.384_{\pm 2.824}$
& $309.956_{\pm 127.727}$ & $14.512_{\pm 3.145}$
& $\underline{194.090_{\pm 14.710}}$ & $13.322_{\pm 2.886}$
& $\cellcolor{bestblue}\mathbf{136.191_{\pm 38.240}}$ & $\cellcolor{bestblue}\mathbf{9.417_{\pm 1.683}}$ \\

\cmidrule{2-23}

&\multirow{4}{*}{DBP}
& AnyPPG
& $245.139_{\pm 106.675}$ & $11.525_{\pm 1.715}$
& $285.319_{\pm 93.920}$ & $12.774_{\pm 0.700}$
& $248.110_{\pm 109.902}$ & $11.537_{\pm 1.724}$
& $220.622_{\pm 30.646}$ & $12.007_{\pm 0.625}$
& $329.623_{\pm 83.452}$ & $14.543_{\pm 2.303}$
& $\underline{199.901_{\pm 44.736}}$ & $\underline{11.371_{\pm 1.284}}$
& $213.504_{\pm 30.217}$ & $11.639_{\pm 0.250}$
& $363.325_{\pm 81.829}$ & $15.538_{\pm 2.215}$
& $309.329_{\pm 60.555}$ & $14.167_{\pm 1.116}$
& $\cellcolor{bestblue}\mathbf{153.459_{\pm 15.796}}$ & $\cellcolor{bestblue}\mathbf{10.195_{\pm 0.580}}$ \\

&& PaPaGei-P
& $336.107_{\pm 152.513}$ & $15.038_{\pm 3.283}$
& $356.623_{\pm 180.406}$ & $15.129_{\pm 3.217}$
& $334.480_{\pm 149.860}$ & $14.987_{\pm 3.197}$
& $216.306_{\pm 7.073}$ & $12.184_{\pm 0.496}$
& $211.046_{\pm 4.889}$ & $12.168_{\pm 0.284}$
& $311.798_{\pm 100.772}$ & $14.176_{\pm 1.751}$
& $228.868_{\pm 21.851}$ & $12.446_{\pm 0.781}$
& $\underline{200.726_{\pm 2.816}}$ & $\underline{12.153_{\pm 0.094}}$
& $218.867_{\pm 8.853}$ & $12.320_{\pm 0.081}$
& $\cellcolor{bestblue}\mathbf{186.804_{\pm 15.548}}$ & $\cellcolor{bestblue}\mathbf{11.337_{\pm 0.791}}$ \\

&& PaPaGei-S
& $287.337_{\pm 79.549}$ & $13.617_{\pm 1.556}$
& $412.088_{\pm 126.399}$ & $15.609_{\pm 2.259}$
& $288.868_{\pm 79.755}$ & $13.626_{\pm 1.521}$
& $\underline{207.111_{\pm 12.269}}$ & $\cellcolor{bestblue}\mathbf{11.881_{\pm 0.259}}$
& $213.396_{\pm 12.900}$ & $12.157_{\pm 0.470}$
& $551.298_{\pm 387.857}$ & $18.605_{\pm 7.410}$
& $362.254_{\pm 133.959}$ & $14.935_{\pm 2.733}$
& $260.305_{\pm 88.574}$ & $13.131_{\pm 1.654}$
& $299.418_{\pm 106.219}$ & $13.876_{\pm 2.386}$
& $\cellcolor{bestblue}\mathbf{193.974_{\pm 0.704}}$ & $\underline{11.898_{\pm 0.207}}$ \\

&& Pulse-PPG
& $\underline{239.984_{\pm 5.556}}$ & $\underline{12.160_{\pm 0.141}}$
& $277.450_{\pm 35.596}$ & $13.754_{\pm 1.296}$
& $240.546_{\pm 4.151}$ & $12.233_{\pm 0.132}$
& $608.723_{\pm 316.731}$ & $18.827_{\pm 3.123}$
& $547.154_{\pm 220.928}$ & $16.620_{\pm 2.982}$
& $244.191_{\pm 12.084}$ & $12.772_{\pm 0.758}$
& $285.420_{\pm 51.069}$ & $13.961_{\pm 1.893}$
& $353.672_{\pm 63.593}$ & $15.587_{\pm 1.596}$
& $752.999_{\pm 96.218}$ & $22.112_{\pm 2.459}$
& $\cellcolor{bestblue}\mathbf{176.193_{\pm 4.934}}$ & $\cellcolor{bestblue}\mathbf{11.405_{\pm 0.097}}$ \\

\cmidrule{2-23}

& \multirow{4}{*}{SpO$_2$}
& AnyPPG
& $1.787_{\pm 0.062}$ & $1.061_{\pm 0.094}$
& $2.231_{\pm 0.507}$ & $1.200_{\pm 0.153}$
& $2.080_{\pm 0.119}$ & $1.151_{\pm 0.079}$
& $1.638_{\pm 0.117}$ & $1.047_{\pm 0.072}$
& $1.795_{\pm 0.363}$ & $1.098_{\pm 0.176}$
& $1.827_{\pm 0.272}$ & $1.107_{\pm 0.069}$
& $1.488_{\pm 0.721}$ & $0.985_{\pm 0.244}$
& $1.789_{\pm 0.269}$ & $1.073_{\pm 0.147}$
& $\underline{1.288_{\pm 0.250}}$ & $\underline{0.937_{\pm 0.075}}$
& $\cellcolor{bestblue}\mathbf{0.898_{\pm 0.171}}$ & $\cellcolor{bestblue}\mathbf{0.763_{\pm 0.078}}$ \\

&& PaPaGei-P
& $2.093_{\pm 0.589}$ & $1.182_{\pm 0.181}$
& $2.998_{\pm 1.143}$ & $1.453_{\pm 0.235}$
& $2.205_{\pm 0.397}$ & $1.208_{\pm 0.117}$
& $\underline{1.626_{\pm 0.079}}$ & $1.089_{\pm 0.030}$
& $1.719_{\pm 0.246}$ & $1.137_{\pm 0.156}$
& $1.927_{\pm 0.474}$ & $1.133_{\pm 0.157}$
& $1.774_{\pm 0.483}$ & $1.121_{\pm 0.183}$
& $1.710_{\pm 0.060}$ & $1.124_{\pm 0.025}$
& $1.665_{\pm 0.342}$ & $\underline{1.079_{\pm 0.164}}$
& $\cellcolor{bestblue}\mathbf{0.793_{\pm 0.220}}$ & $\cellcolor{bestblue}\mathbf{0.749_{\pm 0.106}}$ \\

&& PaPaGei-S
& $2.015_{\pm 0.317}$ & $1.134_{\pm 0.146}$
& $2.417_{\pm 1.042}$ & $1.315_{\pm 0.334}$
& $2.728_{\pm 2.017}$ & $1.299_{\pm 0.566}$
& $1.918_{\pm 0.353}$ & $1.190_{\pm 0.100}$
& $1.869_{\pm 0.335}$ & $1.160_{\pm 0.089}$
& $7.452_{\pm 1.584}$ & $2.301_{\pm 0.361}$
& $2.506_{\pm 0.772}$ & $1.366_{\pm 0.271}$
& $1.857_{\pm 0.350}$ & $1.163_{\pm 0.116}$
& $\underline{1.665_{\pm 0.082}}$ & $\underline{1.094_{\pm 0.034}}$
& $\cellcolor{bestblue}\mathbf{0.935_{\pm 0.385}}$ & $\cellcolor{bestblue}\mathbf{0.797_{\pm 0.171}}$ \\

&& Pulse-PPG
& $2.211_{\pm 1.058}$ & $1.226_{\pm 0.405}$
& $2.443_{\pm 1.153}$ & $1.296_{\pm 0.407}$
& $2.239_{\pm 1.001}$ & $1.245_{\pm 0.373}$
& $\underline{1.622_{\pm 0.203}}$ & $\underline{1.076_{\pm 0.063}}$
& $2.498_{\pm 0.837}$ & $1.281_{\pm 0.190}$
& $1.976_{\pm 0.775}$ & $1.176_{\pm 0.300}$
& $2.480_{\pm 0.865}$ & $1.376_{\pm 0.289}$
& $2.267_{\pm 1.064}$ & $1.278_{\pm 0.360}$
& $1.693_{\pm 0.275}$ & $1.094_{\pm 0.069}$
& $\cellcolor{bestblue}\mathbf{0.872_{\pm 0.417}}$ & $\cellcolor{bestblue}\mathbf{0.758_{\pm 0.180}}$ \\

\midrule

\multirow{4}{*}{BUT PPG} & \multirow{4}{*}{HR}
& AnyPPG
& $202.635_{\pm 2.369}$ & $10.675_{\pm 0.053}$
& $204.919_{\pm 9.240}$ & $10.632_{\pm 0.173}$
& $203.507_{\pm 2.719}$ & $10.615_{\pm 0.050}$
& $212.448_{\pm 2.224}$ & $11.282_{\pm 0.528}$
& $187.598_{\pm 3.073}$ & $10.222_{\pm 0.120}$
& $198.694_{\pm 2.927}$ & $10.550_{\pm 0.041}$
& $337.001_{\pm 50.060}$ & $13.130_{\pm 1.096}$
& $\underline{185.012_{\pm 5.443}}$ & $10.208_{\pm 0.067}$
& $191.083_{\pm 10.642}$ & $\underline{10.107_{\pm 0.101}}$
& $\cellcolor{bestblue}\mathbf{179.911_{\pm 2.311}}$ & $\cellcolor{bestblue}\mathbf{9.931_{\pm 0.035}}$ \\

&& PaPaGei-P
& $212.401_{\pm 3.264}$ & $11.336_{\pm 0.046}$
& $236.696_{\pm 27.146}$ & $11.803_{\pm 0.074}$
& $214.771_{\pm 2.135}$ & $11.282_{\pm 0.071}$
& $206.196_{\pm 2.755}$ & $11.037_{\pm 0.067}$
& $203.959_{\pm 1.785}$ & $\cellcolor{bestblue}\mathbf{10.917_{\pm 0.043}}$
& $225.336_{\pm 11.010}$ & $11.347_{\pm 0.180}$
& $214.450_{\pm 4.245}$ & $11.324_{\pm 0.094}$
& $\underline{203.093_{\pm 0.818}}$ & $11.007_{\pm 0.006}$
& $\cellcolor{bestblue}\mathbf{201.520_{\pm 1.483}}$ & $\underline{10.937_{\pm 0.032}}$
& $207.383_{\pm 0.504}$ & $11.265_{\pm 0.099}$ \\

&& PaPaGei-S
& $237.665_{\pm 6.593}$ & $12.054_{\pm 0.160}$
& $255.788_{\pm 20.345}$ & $12.136_{\pm 0.347}$
& $242.740_{\pm 6.394}$ & $12.066_{\pm 0.159}$
& $\underline{214.732_{\pm 1.609}}$ & $\cellcolor{bestblue}\mathbf{11.140_{\pm 0.080}}$
& $216.468_{\pm 1.334}$ & $11.422_{\pm 0.254}$
& $236.288_{\pm 9.659}$ & $11.675_{\pm 0.183}$
& $233.095_{\pm 7.323}$ & $11.814_{\pm 0.134}$
& $215.611_{\pm 2.499}$ & $11.479_{\pm 0.227}$
& $\cellcolor{bestblue}\mathbf{211.060_{\pm 1.398}}$ & $\underline{11.257_{\pm 0.026}}$
& $215.597_{\pm 2.858}$ & $11.318_{\pm 0.095}$ \\

&& Pulse-PPG
& $228.978_{\pm 4.254}$ & $11.830_{\pm 0.140}$
& $229.720_{\pm 15.857}$ & $12.059_{\pm 0.654}$
& $231.559_{\pm 4.841}$ & $11.812_{\pm 0.128}$
& $269.051_{\pm 38.955}$ & $12.855_{\pm 1.081}$
& $265.401_{\pm 27.272}$ & $12.305_{\pm 0.359}$
& $223.648_{\pm 6.046}$ & $11.600_{\pm 0.107}$
& $\cellcolor{bestblue}\mathbf{207.564_{\pm 1.828}}$ & $\cellcolor{bestblue}\mathbf{11.135_{\pm 0.062}}$
& $250.855_{\pm 25.734}$ & $12.435_{\pm 0.277}$
& $303.459_{\pm 20.781}$ & $13.693_{\pm 0.339}$
& $\underline{208.575_{\pm 2.300}}$ & $\underline{11.205_{\pm 0.193}}$
\\

\midrule

\multicolumn{3}{c|}{\textbf{Win Rate}}
& \multicolumn{2}{c|}{\textbf{0\%}}
& \multicolumn{2}{c|}{\textbf{3.1\%}}
& \multicolumn{2}{c|}{\textbf{0\%}}
& \multicolumn{2}{c|}{\textbf{3.1\%}}
& \multicolumn{2}{c|}{\textbf{4.7\%}}
& \multicolumn{2}{c|}{\textbf{0\%}}
& \multicolumn{2}{c|}{\textbf{3.1\%}}
& \multicolumn{2}{c|}{\textbf{0\%}}
& \multicolumn{2}{c|}{\textbf{3.1\%}}
& \multicolumn{2}{c}{\textbf{82.8\%}}
\\

\bottomrule
\end{tabular}
}

\vspace{1mm}

\begin{minipage}{\textwidth}
\footnotesize
* ``Win Rate'' denotes the percentage of MSE and MAE comparisons in which a method achieves the best result.
\end{minipage}

\vspace{3mm}

\vspace{3mm}

\caption{The performance of the methods on classification tasks.}
\label{tab:Complete_overall_performance_2}

\resizebox{\textwidth}{!}{
\begin{tabular}{
c|c|c|
*{9}{cc|}
cc
}
\toprule
{} & {} & {}
& \multicolumn{2}{c|}{\textbf{Linear Probe}}
& \multicolumn{2}{c|}{\textbf{Full-FT}}
& \multicolumn{2}{c|}{\textbf{LoRA}}
& \multicolumn{2}{c|}{\textbf{QLoRA}}
& \multicolumn{2}{c|}{\textbf{PFEC}}
& \multicolumn{2}{c|}{\textbf{Taylor-FO}}
& \multicolumn{2}{c|}{\textbf{FPGM}}
& \multicolumn{2}{c|}{\textbf{SRR-GR}}
& \multicolumn{2}{c|}{\textbf{Less-is-More}}
& \multicolumn{2}{c}{\textbf{Ours}}
\\

\textbf{Data} & \textbf{Tasks} & \textbf{FM}
& F1 $\uparrow$ & AUROC $\uparrow$
& F1 $\uparrow$ & AUROC $\uparrow$
& F1 $\uparrow$ & AUROC $\uparrow$
& F1 $\uparrow$ & AUROC $\uparrow$
& F1 $\uparrow$ & AUROC $\uparrow$
& F1 $\uparrow$ & AUROC $\uparrow$
& F1 $\uparrow$ & AUROC $\uparrow$
& F1 $\uparrow$ & AUROC $\uparrow$
& F1 $\uparrow$ & AUROC $\uparrow$
& F1 $\uparrow$ & AUROC $\uparrow$
\\
\midrule

\multirow{4}{*}{BUT PPG} & \multirow{4}{*}{SQI}
& AnyPPG
& $0.622_{\pm 0.012}$ & $0.799_{\pm 0.017}$
& $0.599_{\pm 0.021}$ & $0.724_{\pm 0.039}$
& $0.645_{\pm 0.031}$ & $0.793_{\pm 0.013}$
& $0.579_{\pm 0.059}$ & $0.626_{\pm 0.150}$
& $0.678_{\pm 0.035}$ & $\cellcolor{bestblue}\mathbf{0.811_{\pm 0.031}}$
& $0.632_{\pm 0.010}$ & $0.734_{\pm 0.005}$
& $0.622_{\pm 0.013}$ & $0.755_{\pm 0.020}$
& $\underline{0.697_{\pm 0.029}}$ & $0.800_{\pm 0.016}$
& $0.695_{\pm 0.059}$ & $0.799_{\pm 0.019}$
& $\cellcolor{bestblue}\mathbf{0.707_{\pm 0.008}}$ & $\underline{0.806_{\pm 0.009}}$
\\

&& PaPaGei-P
& $0.535_{\pm 0.042}$ & $\underline{0.736_{\pm 0.008}}$
& $0.595_{\pm 0.024}$ & $0.682_{\pm 0.043}$
& $0.475_{\pm 0.032}$ & $0.685_{\pm 0.048}$
& $0.558_{\pm 0.043}$ & $0.579_{\pm 0.036}$
& $0.585_{\pm 0.041}$ & $0.628_{\pm 0.031}$
& $\underline{0.599_{\pm 0.014}}$ & $0.712_{\pm 0.016}$
& $0.544_{\pm 0.076}$ & $0.690_{\pm 0.009}$
& $0.581_{\pm 0.029}$ & $0.598_{\pm 0.017}$
& $\underline{0.599_{\pm 0.031}}$ & $0.633_{\pm 0.025}$
& $\cellcolor{bestblue}\mathbf{0.669_{\pm 0.006}}$ & $\cellcolor{bestblue}\mathbf{0.766_{\pm 0.028}}$
\\

&& PaPaGei-S
& $0.544_{\pm 0.037}$ & $\underline{0.714_{\pm 0.006}}$
& $0.606_{\pm 0.029}$ & $0.708_{\pm 0.016}$
& $0.491_{\pm 0.031}$ & $0.697_{\pm 0.014}$
& $\underline{0.629_{\pm 0.019}}$ & $0.686_{\pm 0.040}$
& $0.614_{\pm 0.024}$ & $0.645_{\pm 0.082}$
& $0.612_{\pm 0.015}$ & $0.701_{\pm 0.020}$
& $0.594_{\pm 0.037}$ & $0.695_{\pm 0.013}$
& $0.600_{\pm 0.026}$ & $0.633_{\pm 0.033}$
& $0.616_{\pm 0.024}$ & $0.675_{\pm 0.029}$
& $\cellcolor{bestblue}\mathbf{0.686_{\pm 0.019}}$ & $\cellcolor{bestblue}\mathbf{0.801_{\pm 0.014}}$
\\

&& Pulse-PPG
& $0.514_{\pm 0.013}$ & $0.718_{\pm 0.004}$
& $0.562_{\pm 0.091}$ & $0.724_{\pm 0.011}$
& $0.515_{\pm 0.013}$ & $0.717_{\pm 0.005}$
& $0.517_{\pm 0.019}$ & $0.575_{\pm 0.130}$
& $0.629_{\pm 0.045}$ & $0.717_{\pm 0.036}$
& $0.572_{\pm 0.036}$ & $0.738_{\pm 0.004}$
& $0.540_{\pm 0.045}$ & $\underline{0.764_{\pm 0.047}}$
& $\underline{0.630_{\pm 0.060}}$ & $0.749_{\pm 0.066}$
& $0.566_{\pm 0.043}$ & $0.734_{\pm 0.015}$
& $\cellcolor{bestblue}\mathbf{0.638_{\pm 0.027}}$ & $\cellcolor{bestblue}\mathbf{0.766_{\pm 0.015}}$
\\

\midrule

\multirow{4}{*}{DeepBeat} & \multirow{4}{*}{AF}
& AnyPPG
& $0.544_{\pm 0.001}$ & $0.708_{\pm 0.001}$
& $0.651_{\pm 0.009}$ & $\underline{0.838_{\pm 0.006}}$
& $0.468_{\pm 0.032}$ & $0.525_{\pm 0.043}$
& $0.609_{\pm 0.076}$ & $0.754_{\pm 0.053}$
& $0.642_{\pm 0.065}$ & $0.799_{\pm 0.043}$
& $0.646_{\pm 0.005}$ & $0.833_{\pm 0.003}$
& $\underline{0.653_{\pm 0.008}}$ & $0.834_{\pm 0.007}$
& $0.638_{\pm 0.074}$ & $0.800_{\pm 0.049}$
& $0.640_{\pm 0.084}$ & $0.796_{\pm 0.063}$
& $\cellcolor{bestblue}\mathbf{0.684_{\pm 0.007}}$ & $\cellcolor{bestblue}\mathbf{0.844_{\pm 0.015}}$
\\

&& PaPaGei-P
& $0.441_{\pm 0.001}$ & $0.608_{\pm 0.001}$
& $0.624_{\pm 0.009}$ & $0.785_{\pm 0.005}$
& $0.432_{\pm 0.000}$ & $0.545_{\pm 0.044}$
& $0.627_{\pm 0.011}$ & $0.754_{\pm 0.008}$
& $0.649_{\pm 0.011}$ & $0.781_{\pm 0.012}$
& $0.623_{\pm 0.005}$ & $0.769_{\pm 0.004}$
& $0.599_{\pm 0.027}$ & $\underline{0.805_{\pm 0.008}}$
& $\cellcolor{bestblue}\mathbf{0.677_{\pm 0.007}}$ & $0.800_{\pm 0.007}$
& $\underline{0.662_{\pm 0.004}}$ & $0.788_{\pm 0.002}$
& $0.649_{\pm 0.006}$ & $\cellcolor{bestblue}\mathbf{0.831_{\pm 0.006}}$
\\

&& PaPaGei-S
& $0.444_{\pm 0.000}$ & $0.610_{\pm 0.001}$
& $0.596_{\pm 0.018}$ & $\underline{0.789_{\pm 0.003}}$
& $0.428_{\pm 0.009}$ & $0.406_{\pm 0.026}$
& $0.585_{\pm 0.003}$ & $0.671_{\pm 0.027}$
& $0.617_{\pm 0.022}$ & $0.738_{\pm 0.029}$
& $0.594_{\pm 0.003}$ & $0.775_{\pm 0.002}$
& $0.589_{\pm 0.017}$ & $0.785_{\pm 0.022}$
& $\underline{0.618_{\pm 0.015}}$ & $0.735_{\pm 0.013}$
& $0.604_{\pm 0.007}$ & $0.727_{\pm 0.005}$
& $\cellcolor{bestblue}\mathbf{0.638_{\pm 0.026}}$ & $\cellcolor{bestblue}\mathbf{0.809_{\pm 0.006}}$
\\

&& Pulse-PPG
& $0.496_{\pm 0.042}$ & $0.653_{\pm 0.048}$
& $0.625_{\pm 0.035}$ & $0.808_{\pm 0.022}$
& $0.525_{\pm 0.020}$ & $0.600_{\pm 0.032}$
& $0.648_{\pm 0.001}$ & $0.716_{\pm 0.089}$
& $0.633_{\pm 0.026}$ & $0.753_{\pm 0.051}$
& $0.618_{\pm 0.005}$ & $0.803_{\pm 0.005}$
& $0.609_{\pm 0.011}$ & $0.806_{\pm 0.005}$
& $\underline{0.657_{\pm 0.008}}$ & $0.807_{\pm 0.003}$
& $0.643_{\pm 0.026}$ & $\underline{0.809_{\pm 0.004}}$
& $\cellcolor{bestblue}\mathbf{0.671_{\pm 0.009}}$ & $\cellcolor{bestblue}\mathbf{0.841_{\pm 0.005}}$
\\

\midrule

\multicolumn{3}{c|}{\textbf{Win Rate}}
& \multicolumn{2}{c|}{\textbf{0\%}}
& \multicolumn{2}{c|}{\textbf{0\%}}
& \multicolumn{2}{c|}{\textbf{0\%}}
& \multicolumn{2}{c|}{\textbf{0\%}}
& \multicolumn{2}{c|}{\textbf{6.3\%}}
& \multicolumn{2}{c|}{\textbf{0\%}}
& \multicolumn{2}{c|}{\textbf{0\%}}
& \multicolumn{2}{c|}{\textbf{6.3\%}}
& \multicolumn{2}{c|}{\textbf{0\%}}
& \multicolumn{2}{c}{\textbf{87.5\%}}
\\
\bottomrule
\end{tabular}
}
\end{sidewaystable*}

\end{document}